\documentclass[a4paper,fleqn, twocolumn]{cas-dc}

\usepackage{color,soul}
\usepackage{pgfplots}
\pgfplotsset{compat=1.14}
\usepackage{subcaption}
\usepackage{mathtools, siunitx}
\usepackage{graphicx}
\usepackage{float}
\usepackage{lipsum}

\begin{document}

\shorttitle{Hardware Spiking Neurons for Embedded Artificial Intelligence}
\shortauthors{Abderrahmane et al.}

\title [mode = title]{Design Space Exploration of Hardware Spiking Neurons for Embedded Artificial Intelligence}                      

\author[1]{Nassim Abderrahmane}
\ead{nassim.abderrahmane@univ-cotedazur.fr}
\author[1,2]{Edgar Lemaire}
\ead{edgar.lemaire@thalesgroup.com}
\author[1]{Beno\^it Miramond}
\ead{benoit.miramond@univ-cotedazur.fr}

\address[1]{Universit\'e C\^ote d'Azur, CNRS, LEAT, France}
\address[2]{Thales Research Technology France / STI Group / LCHP, Palaiseau, France}

\begin{abstract}[S U M M A R Y]
Machine learning is yielding unprecedented interest in research and industry, due to recent success in many applied contexts such as image classification and object recognition. However, the deployment of these systems requires huge computing capabilities, thus making them unsuitable for embedded systems. To deal with this limitation, many researchers are investigating brain-inspired computing, which would be a perfect alternative to the conventional Von Neumann architecture based computers (CPU/GPU) that meet the requirements for computing performance, but not for energy-efficiency. Therefore, neuromorphic hardware circuits that are adaptable for both parallel and distributed computations need to be designed. In this paper, we focus on Spiking Neural Networks (SNNs) with a comprehensive study of information coding methods and hardware exploration. In this context, we propose a framework for neuromorphic hardware design space exploration, which allows to define a suitable architecture based on application-specific constraints and starting from a wide variety of possible architectural choices. For this framework, we have developed a behavioral level simulator for neuromorphic hardware architectural exploration named NAXT. Moreover, we propose modified versions of the standard Rate Coding technique to make trade-offs with the Time Coding paradigm, which is characterized by the low number of spikes propagating in the network. Thus, we are able to reduce the number of spikes while keeping the same neuron's model, which results in an SNN with fewer events to process. By doing so, we seek to reduce the amount of power consumed by the hardware. %EDGAR : STYLE DEBUT 
Furthermore, we present three neuromorphic hardware architectures in order to quantitatively study the implementation of SNNs. One of these architectures integrates a novel hybrid structure: a highly-parallel computation core for most solicited layers, and time-multiplexed computation units for deeper layers. These architectures are derived from a novel funnel-like Design Space Exploration framework for neuromorphic hardware.
%EDGAR : STYLE FIN

\end{abstract}


\maketitle

\section{Introduction}
\label{introduction}

Over the past decade, Artificial Intelligence (AI) has been increasingly attracting the interest of industry and research organizations. Artificial Neural Networks (ANNs) are derived and inspired from the biological brain, and have become the most well-known and frequently used form of AI. Even though ANNs have garnered a lot of interest in recent years, they stem from the 1940s with the apparition of the first computer. Subsequent work and advancements have lead to the development of a wide variety of ANN models. However, many of these models settled for theory and were not implemented for industrial purposes back then. %EDGAR : STYLE DEBUT
Recently, those algorithms became competitive because of two factors:
%BM: Ces 2 facteurs ont aussi des inconvénients qui ne sont pas mentionnés: cout énergétique pour le premier, annotation des données pour le 2e. Il faudrait tourner la phrase pour le mettre en parallèle. Ou bien en faire mention plus tard.
%NA: Okey
first, modern computers have reached sufficient computing performance to process ANN training and inference; 
%BM: evident :
%which was not the case earlier in computing history; 
%NA: Okey
second, the amount of data available is growing exponentially, satisfying the extensive training data requirements for ANNs.
However, the energy and hardware-resources intensiveness imposed by computation in complex form of ANNs are not matching with another current emerging technology: IoT (Internet of Things) and Edge Computing. To allow for ANNs to be executed in such embedded context, one must deploy dedicated hardware architectures for ANN acceleration.\\
%EDGAR : STYLE FIN
%BM: cette référence ne justifie par la phrase qui la précéde. Il faudrait une réf qui fasse un survey de l'implémentation Hw des réseaux de neurones.
%NA: Je l'ai changé par le papier de survey de ZSE
In this case, the design of neuromorphic architectures is particularly interesting when combined with the study of spiking neural networks. 
Spiking Neural Networks for Deep Learning and Knowledge Representation is a current issue \cite{kasabov2018time} that is particularly relevant for a community of researchers interested in both neurosciences and machine learning.
Our work is part of this approach and attempts to contribute by studying more precisely the question of the hardware design of these models. These networks are all the more advantageous as we plan to execute them in dedicated accelerators. They then take full advantage of the event-driven nature of data flows, the simplicity of its elementary operators and its local and distributed computing and learning properties. 
Several specific hardware solutions have already been proposed in the literature, but they are only solutions isolated from the overall design space where network topologies are often constrained by the characteristics of the circuit architecture. We recommend the opposite approach, which consists in generating the architecture that best supports the network topology. 
Through this study, we therefore propose an exploration framework that makes it possible to evaluate the impact of different spiking models on the effectiveness of their hardware implementation.

\subsection{Spiking neurons for inference} %[in image classification]

The recent achievements of Deep Neural Networks (DNNs) on image classification have given them the leading role in machine learning algorithms and AI research. After the first phase of offline experiments, these methods began to proliferate in our daily lives through autonomous applications close to the user. Thus, more and more applications such as smart devices, IoT or autonomous vehicles require embedded and efficient implementation.
%EDGAR STYLE DEBUT
However, their initial implementation on CPU were too resource-intensive for such constrained systems. Indeed, generic sequential processors are not adapted to intrinsically parallel neural algorithms. Therefore, it became essential to deploy them onto dedicated neuromorphic systems. These architectures are designed to fit the parallel and distributed computation paradigm of ANNs, permitting their implementation in embedded systems.
% BM: la conso a t elle été estimée et réduite de bout en bout dans le papier ?
%NA-EL: OK

%EDGAR STYLE FIN

%--SNN vs. FNNs scientific comparison
ANNs could be separated into three different generations, distinguished by neural computation and information coding. The first generation is characterized by the traditional McCulloch and Pitts neuron model, which outputs discrete binary values ('$0$' or '$1$') \cite{schuman_survey_2017}.
%EDGAR : STYLE DEBUT
The second generation is characterized by the use of continuous activation functions in neural networks forming more complex architectures, such as Boltzmann Machines \cite{ackley1985learning}, Hopfield Networks \cite{hopfield1982neural}, Perceptrons,  Multi-Layer Perceptrons (MLP) \cite{rumelhart1986parallel} and Convolutional Neural Networks (CNN) \cite{alexnet}.
%EDGAR : STYLE FIN
Finally, the third generation of neural algorithms, on which this paper is focused, is Spiking Neural Networks (SNNs). In this model, information is encoded into spikes, inspiring from neuroscience. Indeed, this neuron model mimic biological neurons and synaptic communication mechanisms based on action potentials. The information is thus represented as a flow of spikes, with a wide variety of information coding techniques (see section \ref{information_coding}).

%EDGAR STYLE DEBUT : Cette phrase ne sert à rien je trouve
%Alternatively, ANNs can be sorted in two main families: Formal Neural Networks (FNNs), in which information is coded continuously, and SNNs, in which information is coded in spikes.
%EDGAR STYLE FIN

%--LIF description + avantages pour l'implem hardware
According to this information coding paradigm, SNN processing is performed in an event-based fashion: computation is operated by a spiking neuron when and only when it receives an input spike. Without any stimulation, the neuron remains idle. Hence, computation is strictly performed for relevant information propagation, in contrast to Formal Neural Networks (FNNs), where the states of every neuron are updated periodically. Moreover, the computation is usually much simpler in spiking neurons than in formal neurons. Indeed, even though several models have been identified in neuroscience studies, in a machine learning context, spiking neurons are most often based on a simple (Leaky) Integrate and Fire (IF) model \cite{abbott1999lapicque}. Let us compare IF computation rule with Formal computation rule. The computation rule for Formal Neurons is presented in equation \ref{eqformel}, and the computation rule for Spiking Neurons (IF model) is shown in equation \ref{eqspike}:
\begin{equation}
    \label{eqformel}
    y_{j}^{l}(t)= f(s_{j}^{l}(t)), ~~~s_{j}^{l}(t)=\sum^{N_{l-1}-1}_{i=0}w_{ij}*y_{i}^{l-1}(t)
\end{equation}
With $ y_{j}^{l}(t)$ being the output of the $j^{th}$ neuron of layer $l$, $f()$ a non-linear activation function, $s_{j}^{l}(t)$ the membrane potential of the $j^{th}$ neuron of layer $l$ and $w_{ij}$ the synaptic weight between $i^{th}$ neuron of layer $l-1$ and $j^{th}$ neuron of layer $l$.
%BM: Pourquoi éliminer directement le Leaky sans même l'évoquer ? Cela sera sûrement reproché. => Ajouter une phrase en fin de sous-section.
%NA-EL: OK
\begin{equation}
    \label{eqspike}
    \begin{split}
    \gamma_{j}^{l}(t) = \begin{cases}
 & 1 \text{ if } s_{j}^{l}(t)\geq \theta \\ 
 & 0 \text{ otherwise }
\end{cases}, \\
p_{j}^{l}(t)= \begin{cases}
 &  s_{j}^{l}(t) \text{ if } s_{j}^{l}(t)\leq \theta  \\ 
 &  s_{j}^{l}(t) - \theta \text{ otherwise }
\end{cases}, \\
s_{j}^{l}(t)=p_{j}^{l}(t-1)+\sum^{N_{l-1}-1}_{i=0}(w_{ij}*  \gamma_{i}^{l-1}(t))
\end{split}
\end{equation}
%NA: it was \gamma_{j}^{l}(t)
%BM: dans l'équation x ne sert à rien
%EL:OK
With $\gamma_{j}^{l}(t)$ being the binary output of the $j^{th}$ neuron of layer $l$, $p_{j}^{l}(t)$ the membrane potential of the $j^{th}$ neuron of layer $l$, and $\theta$ the activation threshold of the $j^{th}$ neuron of layer $l$.

The multiplicative operation and the non-linear function $f()$ in eq. \ref{eqformel} are very resource-intensive when considering hardware implementation, whereas the simple accumulation, comparison and reset operations found in eq. \ref{eqspike} are much more competitive. Hence, SNNs are much more promising for low-power embedded hardware implementations than FNNs, considering the advantages in terms of event-driven computation and resource consumption brought by the Integrate and Fire model. Other spiking models exist, such as the slightly more complex Leaky Integrate and Fire (LIF) \cite{liu2001spike}, which implies a continuously decreasing membrane potential; or the Izhikevich neuron model \cite{izhikevich2003simple}, which reproduces more realistic biological neuron behaviors. Other neuron models are described in \cite{kasabov2018time}, which introduces details about spiking neuron models found in literature, alongside a wide variety of learning methods in spiking domain. However, we have chosen to use the simpler IF neuron model in our work, due to increased computational cost with more complex neuron models. Moreover, the IF neuron model is already known to be sufficient for spike-based classification applications \cite{DSNN2019, cao_spiking_2015,cassidy_2013,if_stdp_2012,if_crossbar_2011}.
%{Concerning SNN training, we have chosen the transcoding method, which consists to train a conventional ANN with the same topology as the SNN we want to deploy, and transfer synaptic weights from the ANN model to the SNN model. }

% BM: C'est ici qu'il faut mettre une phrase et une citation sur le fait que IF est suffisant pour ce qu'on veut faire.
%EL-NA: OK

\subsection{Neuromorphic hardware} \label{SOTA}
%AC - is this ok? State-of-the-Art is a bit clunky in this context, and the simple headline seems sufficient?
%BM OK on laisse comme ça

In this subsection, we introduce some of the most recent SNNs hardware implementations found in the literature. Those systems consist of ASIC\footnote{Application Specific Integrated Circuits} or FPGA\footnote{Field Programmable Gate Arrays} chips, designed to simulate large numbers of spiking neurons. We give a brief description of their features, alongside energy consumption information. Those information are summed up in table \ref{BiblioHardArchi}.

\subsubsection*{SpiNNaker}

SpiNNaker \cite{furber2014spinnaker} is a fully digital system aiming to simulate very large spiking networks in real-time, and in an event-driven processing fashion.
%BM Inutile: It has been developed at Manchester University.
%EL OK
A SpiNNaker board is composed of 864 ARM9 cores, divided into 48 chips containing 18 cores each. The memory is highly distributed, as there is no global memory unit, but one small local memory unit for each core and a shared memory for each chip. The main feature of SpiNNaker is its efficient communication system: all the nodes are interconnected through high-throughput connections designed for small packet routing, which contain Address Event Representation (AER) spikes, i.e., the address of the transmitter neuron, the date of the emission, and the destination neuron. This communication scheme has been conceived to tolerate the intrinsic massive parallelism of the ANNs. The SpiNNaker board is programmable thanks to the PyNN interface, PyNN being a Python library for SNN simulation \cite{davison2007pynn, davison2009pynn}, which provides various neuron models (LIF, Izhikevich, etc.) and synaptic plasticity rules such as STDP (Spike-Time-Dependent Plasticity)\cite{bichler_stdp}\cite{kheradpisheh_stdp-based_2018}. In terms of energy usage, a SpiNNaker board has a peak power consumption of 1W.

SpiNNaker is used to implement massively parallel hardware SNNs in the litterature, such as NeuCube in \cite{behrenbeck2018classification}, where a SNN is implemented on SpiNNaker to capture and classify spatio-temporal information from EEG (Electro-EncephaloGram). Notably, this architecture offers the possibility to pause classification process to learn new samples or classes, in an Incremental Learning \cite{carpenter1992fuzzy} \cite{polikar2001learn++} fashion, which is an interesting property.

\subsubsection*{Configurable event-driven convolutional node}
The authors in \cite{camunas-mesa_configurable_2018} proposed a configurable event-driven convolutional node with rate saturation mechanism in order to implement arbitrary CNNs on FPGAs. The designed node consists of a convolutional processing unit formed by a bi-dimensional array of IF neurons and a router allowing to build large 2D arrays dedicated for ConvNets inference. In this structure, each node is directly connected to four other neighboring nodes through ports that carry bidirectional flow of events. Internally, all input and output ports are connected to a router, which dispatches events to its local processing unit or to the appropriate output port. The network described by P\'erez-Carrasco \textit{et al.} \cite{perez-carrasco_mapping_2013} for high-speed poker symbol recognition was  implemented on \textregistered{Xilinx} \textregistered{Spartan} 6 FPGA. With more than 5 K neurons and 500 K synapses, the generated circuit occupied 21,465 slices, 38,451 registers and 202 of block RAMs. The slower versions of the architecture showed recognition rates around 96\% when all the input events were processed by the network, while less than 20\% of the events were processed at real time, obtaining a recognition rate higher than 63\% with a power consumption of 7.7 mW when the stimulus was being processed at real time, and even lower consumptions for slower processing: 5.25 mW when it was 10 times slower, and 0.85 mW for a slow-down factor of 100. 
%BM: Pas d'info sur la conso ?
%NA: j'ai rajouté l'info

\subsubsection*{Conv core}
This paper \cite{yousefzadeh2015fast} proposes a pipe-lined architecture for processing spiking 2D convolutional layers in a fully event-driven system. Indeed, this system takes asynchronous input data from a Dynamic Vision Sensor (DVS) \cite{lichtsteiner2008128} \cite{delbruck2010activity}, a bio-inspired vision sensor which outputs a continuous flow of spikes corresponding to brightness gradient variations in a dynamic image. This architecture benefits from the parallelism offered by FPGAs by implementing a 3-stages-pipeline, thus reaching the great performance of updating 128 pixels of the layer in 12ns; while running on \textregistered{Xilinx} \textregistered{Spartan} 6 FPGA. On the same board, the implementation of a spiking convolution layer with a 128x128 pixel input and a 23x23 convolution kernel occupies 48\% of logic resources and 68\% of block RAM. This architecture uses the LIF neuron model, a bit more complex than our simple IF neuron. This system is adapted to asynchronous spiking input, whereas our system is adapted to conventional CCD (Charge Coupled Device) vision sensors, however we could adapt our architecture to DVS to benefit from the asynchronous input in parallel implementations (FPA architecture, see \ref{FPA}).
%BM: auparavant "our system is also adapted to conventional CCD". Est ce qu'on est vraiment restreint aux imageurs CCD ? On ne peut pas traiter un flux DVS ?
%EL Pas en l'état pour les implémentations multipléxées, mais on purrait se pencher là dessu dans une prochaine étude. J'ai modifié en conséquence

\subsubsection*{HFirst}
HFirst \cite{orchard2015hfirst} is a Spiking CNN architecture. It is based on a frame-free paradigm, as it takes inputs from a DVS. HFirst's particularity is to focus on relative timing of spikes across neurons, benefiting from the continuous flow of input data. Hence, HFirst is dedicated to temporal pattern recognition, whereas our architecture is dedicated to static image recognition, and uses the accessible CCD sensor. Moreover, HFirst uses another IF neuron version which is more complex than ours, emulating physical behavior of an IF Neuron. This model uses several multiplications, which results in a more resource intensive implementation (17 DSP in HFirst versus 0 for ours). HFirst runs on {Xilinx\textregistered}'s {Spartan\textregistered} 6 FPGA, with a 100MHz clock, and consumes between 150mW and 200mW. It performs 97.5\% accuracy on HFirst Cards data-set (4 classes), and 84.9\% on HFirst Characters data-set (36 classes). 
% The MAX-Pooling layer is implemented with a Temporal Winner-Take-All, where the most activated neuron will be the first and only one to spike.

\subsubsection*{Minitaur}
%EDGAR STYLE DEBUT
Minitaur \cite{minitaur} is an event-driven neural network accelerator dedicated to high performance and low power consumption. This is an SNN accelerator on \textregistered{Xilinx} \textregistered{Spartan} 6 FPGA board. The example LIF-based network implemented on the board performs 92\% accuracy on MNIST dataset and 71\% on 20 newsgroups dataset. The Minitaur architecture consists of 32 LIF-based cores dedicated to parallel processing of spikes. The input spikes arrive from a queue where they are stored as packets through USB interface. Those packets are encoded on 6 Bytes : 4 Bytes for timestamp, 1 Byte for layer index and 2 Bytes for the neuron address (Address-Event-Representation). This is a semi-parallel architecture,  where some layer are processed in parallel, and some layer are processed sequentially. Minitaur achieves 19 million neuron update per second on 1.5 W of power and it supports up to 65K neurons per board within fully-connected layers based SNN.  
%EDGAR STYLE FIN
%BM: qu'est ce que c'est les "postsynaptic currents" => spikes ?
%EL : JE crois que c'est le calcul de l'activation d'un neurone par intégration des entrées (membrane potential). J'ai remplacé par "neuron update".
%NA: ok!
\subsubsection*{Loihi}

%EL Inutile : It has been developed by Intel company. 
Loihi \cite{davies2018loihi} is again a fully-digital chip containing 128 cores, each of which are able to simulate up to 1024 different neurons. The memory is also largely distributed, with each core having a local 2MB SRAM memory unit. The chip also contains 2x86 cores and 16MB of SRAM synaptic memory. Accordingly, it is able to support up to 130 000 neurons and 130 million synapses. In contrast with previous systems, the Loihi board is able to perform learning. The chip can be programmed to implement various learning rules, notably STDP. The chip is able to simulate up to 30 billion SOPS, with an average of 10pJ per spike.

\subsubsection*{TrueNorth}

TrueNorth \cite{akopyan2015truenorth} is another fully-digital system, capable of simulating up to 1 million spiking neurons. A TrueNorth board is composed of 4096 Neurosynaptic cores dedicated to LIF neuron emulation. Each core contains a 12.75 KB of local SRAM memory, and is time-multiplexed up to 256 times so that one core can simulate 256 different neurons. Similarly to SpiNNaker, the communication scheme is asynchronous, event-based and able to tolerate a very high level of parallelism. TrueNorth can perform 46 billion synaptic operations per second (SOPS) per Watt, with a power consumption of 100mW when running a 1 million neurons network. The system is programmable thanks to the Corelet programming language \cite{amir2013cognitive}, allowing to tune neuron parameters, synapse connectivity and inter-core connectivity. 
%BM: What does simple synapses mean ?
%EL: OK

\subsubsection*{DYNAPs}
DYNAPs \cite{Dynaps_2017} is a reconfigurable hybrid analog/digital architecture. Its hierarchical routing network allows the configuration of different neural network topologies. This interesting method also tries to solve the compromise between point-to-point communications and request broadcasting in large neural topologies. 
The use of mixed-mode analog/digital circuits allowed to distribute the memory elements across and within the computing modules. As a counterpart, this requires the addition of conversion circuits. The analog parts are operated in subthreshold domain to minimize dynamic power consumption and to implement biophysically realistic behaviors.
The approach is validated by a VLSI design implementing a three-layer CNN network. If the circuit consumption is low (about ten pJ per data movement in the network), the implementation of the 2560 neurons of the targeted spiking CNN requires the use of a PCB composed of 9 circuits. The overall consumption and scalability of the approach therefore remains to be confirmed.

\subsubsection*{BrainScaleS}

BrainScaleS \cite{schemmel2010wafer} is a mixed digital-analog system. The processing units (neuron cores) are analog circuits, whereas the communication units are digital. BrainScaleS implements the adaptive exponential IF neuron model, which can be configured to reproduce many biological firing patterns. BrainScaleS is composed of HiCANN (High-Input Count Analog Neuronal Network) chips, which are able to simulate 224 spiking neurons and 15 000 synapses. Several HiCANN units can be placed on a wafer, so that a single wafer can simulate up to 180 000 neurons and 40 million synapses. The system also integrates general purpose embedded processors, which are able to measure relative spike timings, thus plasticity rules such as STDP can be implemented. Other plasticity rules can also be programmed, and a PyNN interface allows users to program the network in a similar fashion to SpiNNaker. The BrainScaleS platform consumes between 0.1nJ and 10nJ per spike depending on the simulated network model, and reaches a maximum of 2kW of peak power consumption per module.

\subsubsection*{NeuroGrid}

NeuroGrid \cite{benjamin2014neurogrid} is also a mixed digital-analog system, which targets real-time simulation of large SNNs. It employs subthreshold circuits, to model neural elements. The synaptic functions are directly emulated thanks to the physics of the transistors operating in the subthreshold regime. The board is composed of 16 NeuroCore chips, interconnected by an asynchronous multicast tree routing digital communication system. Each core is composed of 256*256 analog neurons, so that NeuroGrid is able to simulate up to 1 million neurons and billions of synaptic connections. Concerning energy, NeuroGrid consumes an average of 941pJ per spike and has a peak power consumption of 3.1W.

\begin{table*}[]
\centering
\caption{Neuromorphic hardware architectures}
\label{BiblioHardArchi}
\resizebox{\textwidth}{!}{%
\begin{tabular}{|l|l|l|l|l|l|l|l|l|}
\hline
Work                                                                 & Electronics                                                 & Technology                                                         & Online Learning & Programmability & Network         & Neuron model                                                   & Input data  & Application domain                                                        \\ \hline
NeuroGrid \cite{benjamin2014neurogrid} - 2014                                                     & \begin{tabular}[c]{@{}l@{}}Analog / \\ Digital\end{tabular} & \begin{tabular}[c]{@{}l@{}}ASIC - CMOS \\ 180nm\end{tabular}       & yes             & NGPyton         & Programmable    & \begin{tabular}[c]{@{}l@{}}Dimensionless \\ model\end{tabular} & Spikes      & NeuroSciences                                                             \\ \hline
BrainScales \cite{schemmel2010wafer} - 2017                                                   & \begin{tabular}[c]{@{}l@{}}Analog / \\ Digital\end{tabular} & \begin{tabular}[c]{@{}l@{}}ASIC - CMOS \\ 180nm\end{tabular}       & yes             & PyNN            & FC              & exp IF                                                         & Frame-based & \begin{tabular}[c]{@{}l@{}}NeuroSciences / \\ Classification\end{tabular} \\ \hline
Loihi \cite{davies2018loihi} - 2018                                                         & Digital                                                     & \begin{tabular}[c]{@{}l@{}}ASIC - CMoS \\ 14 nm\end{tabular}       & yes             & Loihi API       & Conv / FC / RNN & CUBA LIF                                                       & Spikes      & \begin{tabular}[c]{@{}l@{}}LASSO / \\ classification\end{tabular}         \\ \hline
TrueNorth \cite{akopyan2015truenorth} - 2014                                                     & Digital                                                     & \begin{tabular}[c]{@{}l@{}}ASIC - CMoS \\ 28 nm\end{tabular}       & no              & Corelets        & Conv / FC / RNN & LIF                                                            & Frame-based & \begin{tabular}[c]{@{}l@{}}Multi-object\\ detection\end{tabular}          \\ \hline
\begin{tabular}[c]{@{}l@{}}SpiNNaker CMP \\ Chip \cite{furber2014spinnaker}- 2010\end{tabular} & Digital                                                     & \begin{tabular}[c]{@{}l@{}}ASIC - CMoS \\ 130 nm\end{tabular}      & yes             & PyNN            & Programmable    & LIF, IZH, HH                                                   & Spikes      & NeuroSciences                                                             \\ \hline
Minitaur \cite{minitaur} - 2014                                                      & Digital                                                     & \begin{tabular}[c]{@{}l@{}}FPGA - Spartan 6 \\ LX150\end{tabular}  & no              & RTL               & FC              & LIF                                                            & Frame-based & Classification                                                            \\ \hline
ConfConvNode \cite{camunas-mesa_configurable_2018} - 2018                                                  & Digital                                                     & \begin{tabular}[c]{@{}l@{}}FPGA - Xilinx \\ Spartan 6\end{tabular} & no              & RTL               & Conv / Pool     & LIF                                                            & DVS         & \begin{tabular}[c]{@{}l@{}}DVS-based \\ calssification\end{tabular}       \\ \hline
Fast pipeline \cite{yousefzadeh2015fast} - 2015                                                 & Digital                                                     & \begin{tabular}[c]{@{}l@{}}FPGA - Xilinx \\ Spartan 6\end{tabular} & no              & RTL               & Conv / Pool     & LIF                                                            & DVS         & \begin{tabular}[c]{@{}l@{}}DVS-based \\ calssification\end{tabular}       \\ \hline
HFirst \cite{orchard2015hfirst} - 2015                                                        & Digital                                                     & \begin{tabular}[c]{@{}l@{}}FPGA - Xilinx \\ Spartan 6\end{tabular} & no              & RTL               & Conv / Pool     & Complex IF                                                     & DVS         & \begin{tabular}[c]{@{}l@{}}DVS-based \\ object recognition\end{tabular}   \\ \hline
DYNAPS \cite{Dynaps_2017} - 2017                                                        & \begin{tabular}[c]{@{}l@{}}Analog / \\ Digital\end{tabular} & \begin{tabular}[c]{@{}l@{}}FPGA - CMoS \\ 180 nm\end{tabular}      & no              & CHP language    & Conv / Pool     & AdExp-IF                                                       & DVS         & Classification                                                            \\ \hline
This work - 2019                                                     & Digital                                                     & \begin{tabular}[c]{@{}l@{}}FPGA - Cyclone\\ V 28 nm\end{tabular}   & no              & N2D2, TF, Keras & FC              & IF                                                             & Frame-based & \begin{tabular}[c]{@{}l@{}}Embedded-AI \\ Classification\end{tabular}     \\ \hline
\end{tabular}%
}
\end{table*}

\subsection{Contributions}
\begin{figure}
    \centering
   \begin{subfigure}{0.5\linewidth}
	\centerline{\includegraphics[width=0.95\linewidth]{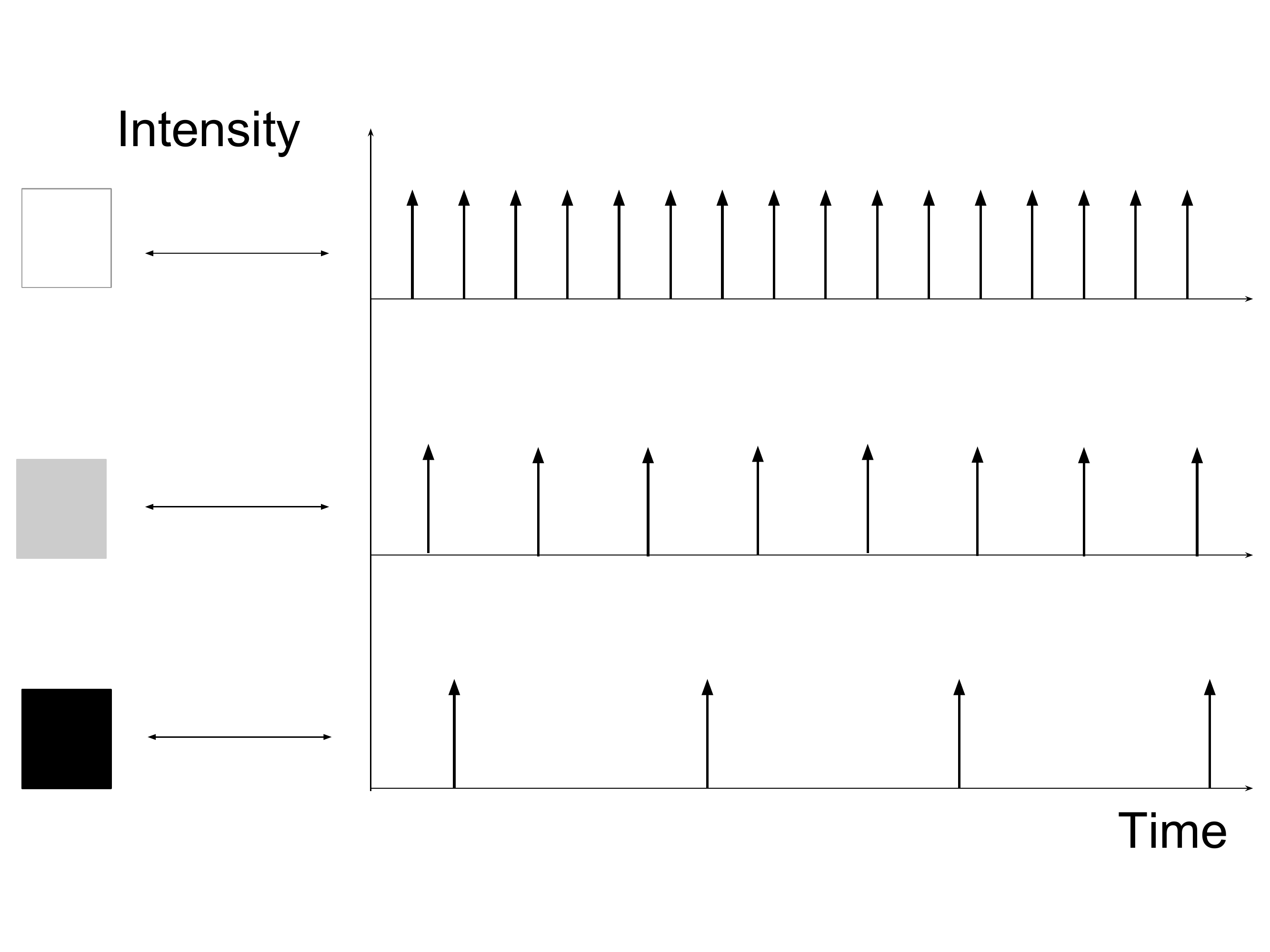}}
	\caption{Rate Coding paradigm}
	\label{rate_coding}
    \end{subfigure}%
    \begin{subfigure}{0.5\linewidth} 
	\centerline{\includegraphics[width=0.95\linewidth]{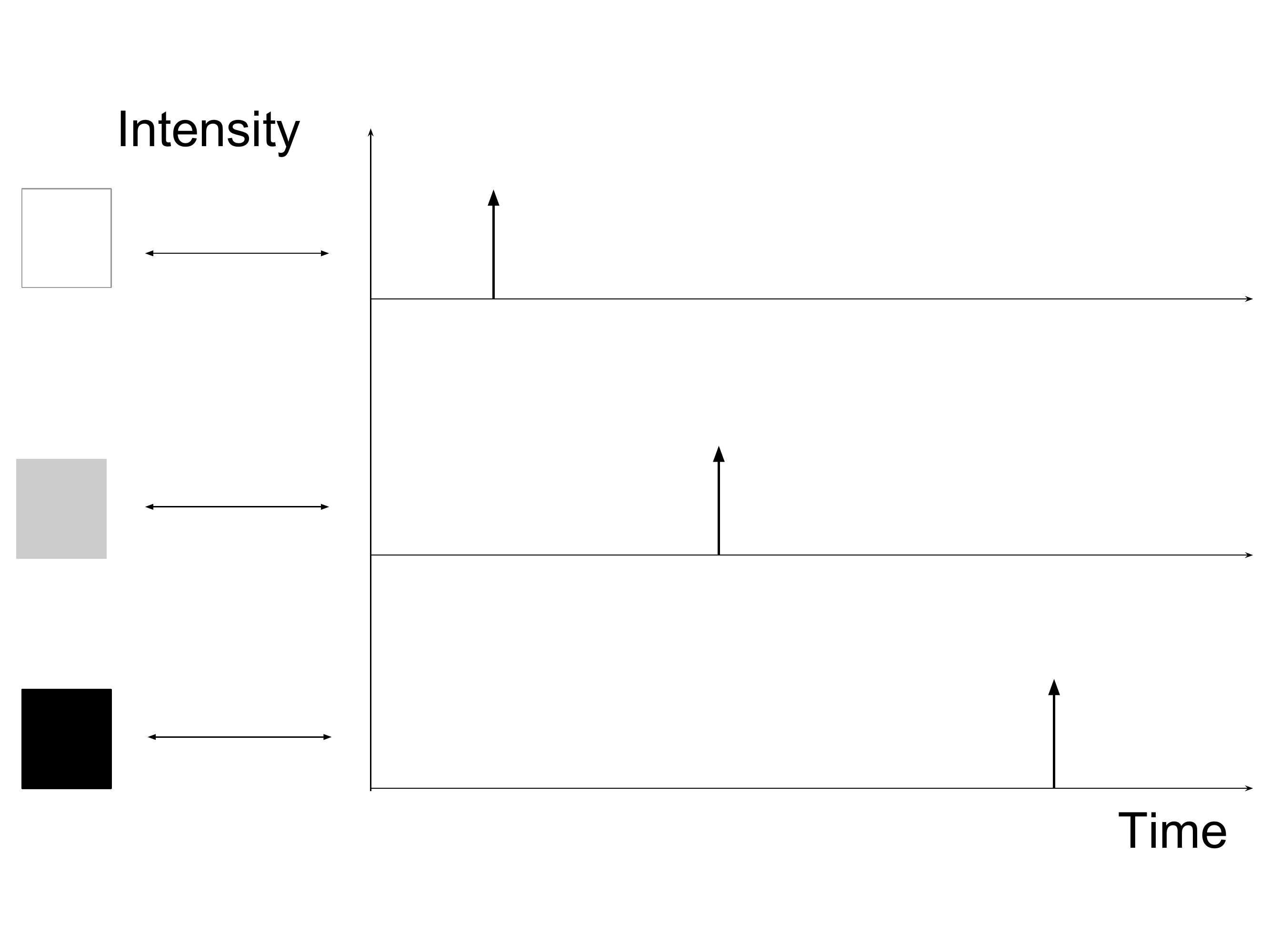}}
	\caption{Time Coding paradigm}
	\label{time_coding}
    \end{subfigure}
    \caption{Information coding methods}
\end{figure}

%AC - does this mean the contributions that your work is making to embedded AI? Perhaps "Context", "Relevance" or "Contribution to the advancement of embedded AI" would better?
%BM No contributions is OK
%El : OK
%BM: après la section précédente qui laisse penser que tout est déjà fait en HW, il faut justifier aussi notre contribution dans ce paragraphe vis à vis de ces travaux.
%NA OK
The hardware accelerators presented so far are largely destined to conduct large-scale simulations of brain-like neural networks, with a bio-mimetic implementation offering several neuron and synapse models (table \ref{BiblioHardArchi}). 
%BM: très contestable comme argument
%Therefore, they are general purpose bio-inspired architectures that are not all suited to embedded systems due to their expensive hardware cost and high energy consumption. 
Therefore, they are either designed for general purpose simulation of bio-inspired neural models, or for processing data coming from event-based cameras.
They are not easily programmable from classical machine learning frameworks. This enables us to combine the efficiency of unsupervised learning and the efficiency of spiking neurons applied to prevalent frame-based sensors. Indeed, in embedded AI applications, the solution has to offer state-of-the-art prediction accuracy. Previous work \cite{SNN_MLP_LEAT} has shown that SNNs cost about 50\% less in terms of hardware, while having approximately the same accuracy compared to MLP (FNNs). In other words, mapping a traditional neural network to a spiking one does not severely impact the recognition rate \cite{diehl_fast-classifying_2015}\cite{perez-carrasco_mapping_2013}, and results in more economical hardware.
%BM: je ne mettrai pas cette phrase à ce stade :
% However, the SNNs are slower than FNNs. 
%Ok
Therefore, in this paper we adopt the same approach, consisting in transcoding FNNs to SNNs to solve classification problems.  Such an approach has been studied in \cite{perez-carrasco_mapping_2013}\cite{diehl_fast-classifying_2015} but few studies have explored the impact of spike coding on both performance and power efficiency \cite{cao_spiking_2015, luo_efficient_2018}. Indeed, our network is first trained in formal domain, and its weights are then exported to be used in an SNN with the same topology, which is then directly ready for inference. Note that there exist learning methods directly in spike domain, such as SpikeProp or STDP \cite{bichler_stdp}\cite{mostafa_supervised_2016}\cite{kheradpisheh_stdp-based_2018}\cite{R_STDP}. Additional information concerning spiking learning methods is available in \cite{kasabov2018time}, which presents a complete survey of Spiking Neural Network training techniques.
%BM : je pense qu'il faut déplacer le paragraphe qui suit dans la partie état de l'art Hw. 1) c'est trop tôt pour être si précis, 2) ce qui permettra aussi de se distinguer.
In this paper, we are dealing with supervised feed-forward networks trained with back-propagation learning, because they are the most dominant deployed models when considering hardware integration\cite{sze_efficient_2017}\cite{DSNN2019}.
%EDGAR STYLE DEBUT
In our study we will also focus on information coding, spike generation and their impact on neuromorphic system efficiency. Our intuition is that using Time Coding rather than Rate Coding, as is widely used in related studies, leads to a system with reduced power consumption. When transcoding an image in spike domain with Time Coding, for example, each pixel will fire at most once (figure \ref{time_coding}). With rate coding, however, spike trains are emitted for each input pixel (\ref{rate_coding}), which results in a greater activity in the network, thus increasing resource and energy intensiveness of the system. Therefore, we developed innovative spike coding method based on Time Coding paradigm, such as Spike-Select and First-Spike.
%BM: Que voulez vous dire par : "Note that researchers are very interested in developing learning methods in spike domain" ? Vérité générale ? Ou spécifique aux auteurs de ce papiers (nous) ?
%EL Modifié, changé par "There exist methods..." plutot que "Researchers are interested in methods...", plus objectif et compréhensible.
In the context of hardware SNN implementation, we have also developed NAXT (Neuromorphic Architecture eXploration Tool), which is a software for high-level neuromorphic system simulation. Developed in SystemC \cite{panda2001systemc}, this software provides coarse energy consumption, latency and chip surface estimations for various built-in architectural configurations. Thus, NAXT acts as the first evaluation tool in our architectural exploration of neuromorphic systems for fast but coarse evaluation of architectural choices. More details will be given in section \ref{exploration}. We have chosen to develop our own high-level hardware SNN simulator, as existing simulators did not allow to perform hardware estimations with such high-level description, which makes our simulator innovative.
High-level exploration with NAXT provides information on the suitable architectural choices, such as parallelism and memory distribution. From those results, architectures are built in VHDL to be tested at Register Transfer Level, thus furnishing precise timing, logic resources and energy measures. Notably, we propose a novel Hybrid Architecture, which combines the advantages of both multiplexed and parallel hardware implementations.
%BM Inutile, le plan est annoncé juste derrière
%More details will be given in section \ref{DHSNNA}.
%EDGAR STYLE FIN

\subsection{Outline of the paper}

%BM: mettre des références aux sections
%NA+EL: OK

%EDGAR STYLE DEBUT
The remainder of this paper is organized across four sections. First, we deal with methods for encoding information into Spikes in section \ref{information_coding}. Novel Information Coding methods are described and evaluated alongside common Rate and Time coding methods. In the four next parts of the paper, our funnel-fashioned Design Space Exploration framework is described in chronological order : In section \ref{secII}, we give a brief description of the framework's philosophy, alongside preliminary steps of our Design Flow. In section \ref{exploration}, we deal with the high-level exploration step, describing the novel NAXT (Neuromorphic Architecture eXploration Tool) software and its simulation results on a typical SNN for MNIST classification. The section \ref{DHSNNA} is dedicated to low-level implementation of SNN hardware architectures based on NAXT results, and their RTL (Register Transfer Level) simulation information. An innovative hybrid architecture with both parallel and multiplexed computation cores will be introduced and evaluated. Lastly, we discuss the work while presenting some perspectives in section \ref{discussion}, and conclude the paper in section \ref{conclusion_sec}.
%EDGAR STYLE FIN

\section{Information Coding}
\label{information_coding}
%EDGAR STYLE DEBUT
In this section, we are going to focus on Information Coding methods : those are the different ways in which information can be encoded into spikes. Existing and novel coding methods will be presented.
%EDGAR STYLE FIN
\subsection{Rate Coding versus Time Coding}

In SNN architectures, information is encoded in spikes. The spikes, also called "action potentials" or "nerve pulses" in biology, are generated by a spiking neuron, in a process called "firing". In a feed-forward SNN with Fully-Connected (FC) layers, these spikes are transmitted to all the neurons of the next layer.
Several information coding methods have been proposed by neuroscientists, including Rate Coding, Time Coding, Phase Coding, Rank Coding, Population Coding, etc. \cite{brette_philosophy_2015}. In this study, we focus on Rate Coding and Time Coding for two reasons: 
\begin{enumerate}
\item Rate Coding: for its maturity. When using this method, SNNs reach State-of-the-Art performance on classification applications \cite{SNN_MLP_LEAT}\cite{cao_spiking_2015};
\item Time Coding: when used, fewer spikes are propagated in the SNN, which reduces computation and resource intensiveness during inference \cite{mostafa_supervised_2016}\cite{YU20143}.
\end{enumerate}
Based on these methods, we propose some modified versions of the standard Rate Coding to make trade-offs with the temporal coding paradigm: maintain high accuracy and reduce the number of spikes flowing in the network. Indeed, the energy consumption of an SNN hardware implementation is directly proportional to the number of spikes it generates. As mentioned in \cite{cao_spiking_2015}, an estimation of the energy consumed by processing an image is calculated using the equation \ref{eq_1}.
\begin{equation}
    E_{total} = N_{spikes/image} * \alpha 
    \qquad (\si{\joule\per image})
    \label{eq_1}
\end{equation} 
Where $E_{total}$ the average energy consumed for the processing of an image, $N_{spikes/image}$ the total number spike emission per input pattern, and $\alpha$ the energy consumption of a spike emission.
Note that the spiking-activity-related energy consumption varies from one accelerator to another and obviously depends on the specific architecture. In this paper, we consider three different amounts: $\alpha\_{FPA}$, $\alpha\_{TMA}$ and $\alpha\_{HA}$ related to the three architectures that will be described in section \ref{DHSNNA}.
%BM: dans le papier de cao_spiking_2015, est-ce que cette estimation concernait à la fois FPA et Multiplexed Archi ?
% Dit autrement que devient alpha dans le cas d'un NPU ?
% En l'absence de réponse précise, Peut-être faut-il mettre une phrase qui indique cette nuance et qu'il y a 2: alpha_FPA et alpha_MA
%NA: j'ai rajouté une phrase, dans le papier ils ont dit: pour ultra-low power spike-based neuromorphic hardware, such as Cruz-Albrecht et al. (2012), Merolla et al. (2011).
\subsubsection{Rate Coding}

Rate Coding is the most widespread method for converting formal data into spike trains. The spike train's period is computed based on the formal data value following equation \ref{period_equation}. In figure \ref{rate_coding}, three pixels of a gray-scale image are transformed into spike trains: each pixel is represented by a spike train whose frequency is proportional to its intensity (image processing). Note that some Rate Coding techniques such as Jittered Periodic apply a random factor to spike emission times, which increases network's prediction accuracy. Among several rate coding techniques presented in a previous work \cite{Nassim_DSD}, we are using Jittered Periodic method as it reaches the highest performances while not increasing the spiking activity compared to the other rate-based techniques.

\subsubsection{Time Coding}

%EDGAR STYLE DEBUT
The Time Coding method encodes information into spike emission date, which allows to use only one spike per input pixel (for image processing example).Unlike rate coding, a gray-scale image is encoded by signals holding only one spike per pixel. This latter is emitted in a time $t$ that is inversely proportional to the pixel's intensity\cite{mostafa_supervised_2016}\cite{YU20143}, as depicted in figure \ref{time_coding}.
In this model, spikes are dependent on each other, because their arrival times can be interpreted only relatively to other spikes.

%EDGAR STYLE FIN
Initially, we were interested in the work proposed by H. Mostafa in 2017\cite{mostafa_supervised_2016}, with a supervised learning algorithm based on temporal coding. The SNN processes input data that are first binarized and then transformed to the so-called Z-domain (change of variable : $exp(t)=z$), where all the computations are held. In this approach, a typical neuron has a function called \textit{Get\_Causal\_Set()}, which returns a set of neurons from its previous layer participating in the process of firing (causal neurons). Indeed, this function uses all the previous layer's spiking times to deduce this group of causal neurons. Thus, if we suppose that it processes the spiking times to compute the emission time $t_{emission}$, then this latter would be greater than all the processed spiking times. Meanwhile, it is mentioned in the paper \cite{mostafa_supervised_2016} that the spike emitted at $t_{emission}$ is fired after that the causal neurons have fired, but not after all the previous layer neurons. Hence, the information exchanged between neurons does not correspond to the real time that we can measure in event-based system. Therefore, from a hardware perspective, this method is not viable as the hardware implementation would not operate in real time. Thus, the methods presented in the following section seems more appropriate for hardware implementation.

\subsection{Exploring novel information coding methods}

In this subsection, we will describe innovative information coding methods we have developed. 

\subsubsection{Spike Select}

%EDGAR STYLE DEBUT
Statistical results from Rate Coding SNNs are shown in figure \ref{spikes_layer}, where most of the spiking activity is located in the input layer. We note that only a few spikes are emitted by the deeper layers, which is sufficient for classification. In other words, most of the spikes generated by the input layer does not impact the winner class selection process. Therefore, we propose a novel kind of Rate Coding : the Spike Select method.
This methods consists in identifying the spikes directly involved in the classification process, by filtering them within the first hidden layer neurons. In doing so, we ensure that only the spikes intended to excite the winning class neuron are emitted. Fewer spikes will propagate to the output layer, but everyone of them will be exciting the winning class neuron. Thus, the Terminate Delta (see Part \ref{tdelta}) procedure is still valid, even if it often takes longer to complete the Terminate Delta condition. In this regard, the latency of the whole process is increased, resulting in a higher number of spikes generated before the filter. However, the number of spikes propagating after the filter remains low. From the hardware perspective, this is a very promising method that allows for efficient hardware usage, especially with the hybrid architecture presented in section \ref{ha_section}. Indeed, in this architecture, the first hidden layer is implemented in parallel, and the deeper layer are time-multiplexed : this architectural configuration fits well with the Spike Select coding method.
%EDGAR STYLE FIN

\begin{figure}
    \centerline{\includegraphics[width=0.72\linewidth]{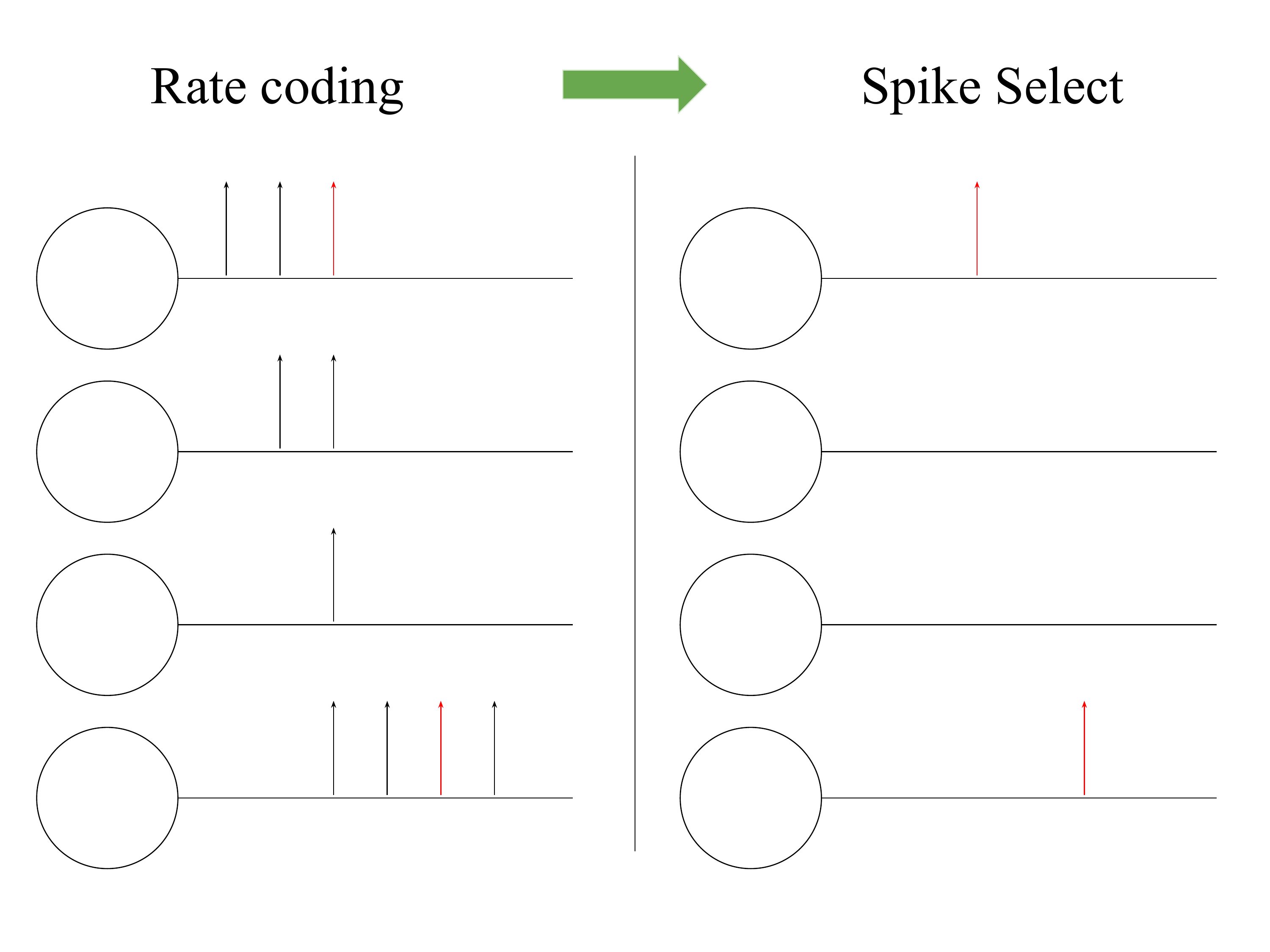}}
    \caption{Spike Select method effect on the first hidden layer neurons. The applied filter consists in raising the threshold value from 1 to 3.}
    \label{spikeselect}
\end{figure}

Figure \ref{spikeselect} shows an example of some first hidden layer neurons to which the Spike Select method has been applied. The filter here consists in raising the threshold from 1 to 3, which reduces the number of emitted spikes from 10 to 2.

The value of the new raised threshold is determined by analyzing the SNN spike flow using standard Rate Coding.

\subsubsection{Single Burst}

With Single Burst coding method the input data stimulus are mapped to temporal domain. An input data value is represented using one spike, which is emitted at a specific time $t$, computed by "$t = \mid 1-v \mid * w_{t}$", with $t$ the emission time, $w_{t}$ the time window dedicated for the generation of the spikes, and $v$ the input value. This method is an adaptation of the existing Single Burst stimulus type in N2D2 \cite{N2D2}.

\subsubsection{First Spike}

Derived from standard rate coding, the novel First Spike method we have developed is an intermediate version between time and rate coding paradigms, having aspects in common with both methods. On one hand, as for time coding, it only uses one spike to represent information. On another hand, similar to rate coding, the compatible neuron model is the IF-neuron.

\begin{figure}
    \centerline{\includegraphics[width=1.6\linewidth]{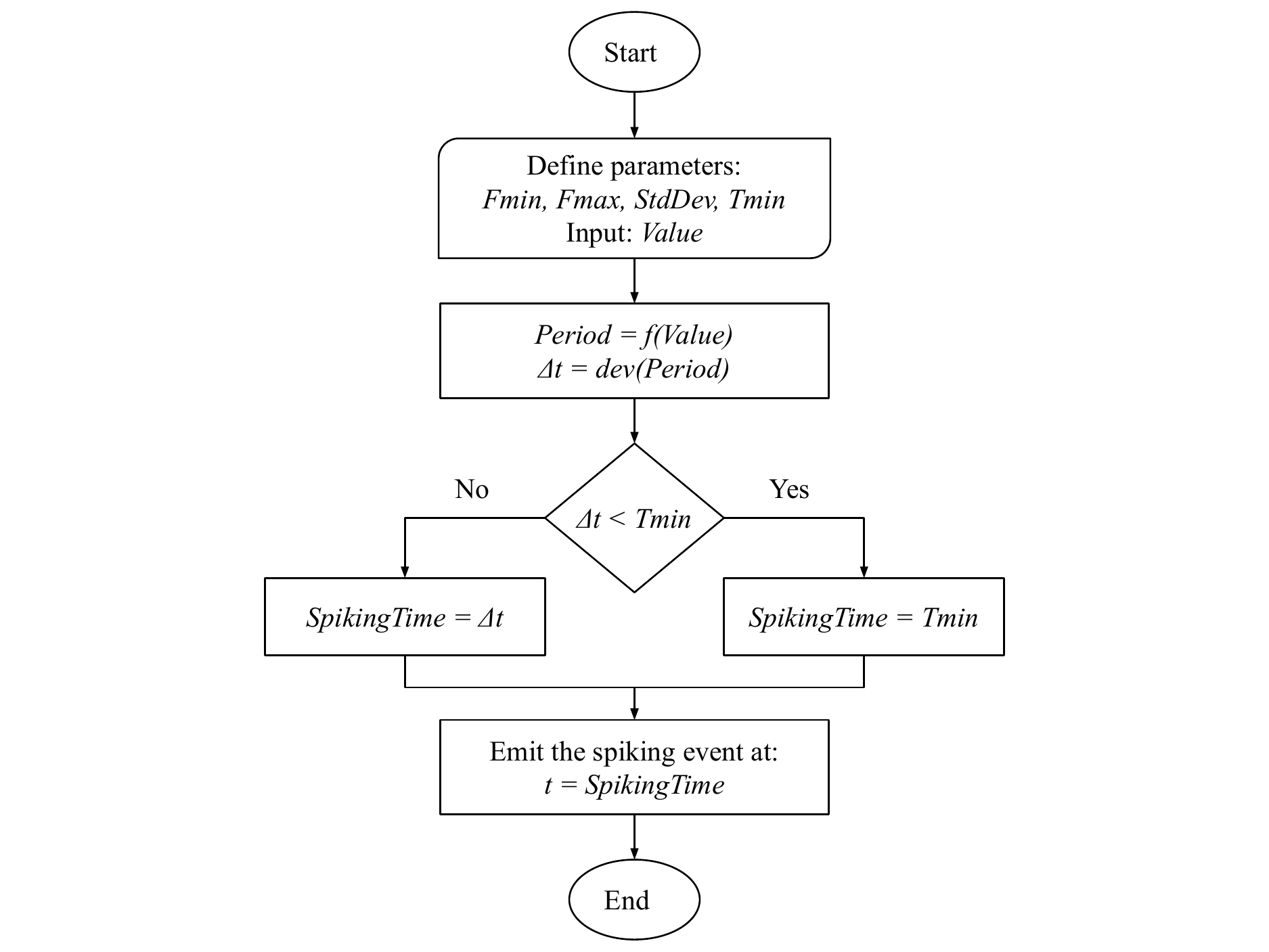}}
    \caption{First Spike method flow-chart. In this method, first, a period corresponding to the input value is computed, which is used to calculate the time step. The time step is the amount by which the actual time is increased to get the spike emission date (only one spike per input value). Fmin: minimum frequency; Fmax: maximum frequency; Tmin: spikes minimum separation time; Value: input value; Period: period equivalent to input value; f(): period conversion function; dev(): deviation function; SpikingTime: spike emission time.}
    \label{single_spike}
\end{figure}

The pseudo-algorithm in figure \ref{single_spike} shows how the information ($v$) is converted to spike domain using the First Spike method. First, we have to define some parameters which will be used in the conversion process. A period is calculated based on value $v$ with $p = f(v)$, using the function $f()$ in equation \ref{period_equation}. 
\begin{equation}
    f(\textbf{v}) = 1/(f_{max}+(1-\mid\ v \mid)*(f_{min}-f_{max}))
    \label{period_equation}
\end{equation}
Where, $f_{min}$ and $f_{max}$ are minimum and maximum frequency parameters. Then, $period$ is used to compute the time step $\Delta t$, which corresponds to the date of the next spike emission, thanks to the function $Deviation()$ presented in equation \ref{diviation_equation}.
\begin{equation}
    Deviation(p)= f_{Udist}(f_{Ndist}(p , s_{dev}))
    \label{diviation_equation}
\end{equation}
With $s_{dev}$ being the standard deviation, $f_{Ndist}()$ the Random normal distribution function and $f_{Udist}()$ the Random uniform distribution function.

Then, this $\Delta t$ value is compared to $Tmin$, which is the minimum spike delay (no spike can be emitted earlier than $Tmin$). If $\Delta t>Tmin$, the spike will be emitted at time $\Delta t$, whereas if $\Delta t < Tmin$, the spike will be emitted at time $Tmin$.

This process is done only once, as this method consists of only one spike emission per input pixel.

%BM: dt n'apparait dans aucune des 2 équations !
%NA: je l'ai fait apparaitre dans le texte maintenant

\subsection{Results}

In this subsection, we present the experimental results of the different information coding methods. We first compare the FNN accuracy results to Rate Coding based SNNs. Then, we analyze performances of the different information coding methods. For training, validating and testing the ANNs, we have used the MNIST data-set, which is a handwritten digits database of 70 000 images (60 000 for learning and validation, 10 000 images for testing) \cite{mnist}.

\subsubsection*{Spiking versus formal neural networks}

We test the robustness of the mapping method through several ANN topologies. For this purpose, we are using N2D2 framework following the steps presented in section \ref{n2d2}.\\
%, with the same hyper parameters used in %\cite{ijcnn_2019}.
%BM: vérifier puis indiquer qu'il s'agit de valeurs moyennes sur X runs
%NA: To be done, after
\begin{table}
\begin{center}
\caption{FNNs versus SNNs accuracy results on MNIST data-set. The information coding method used the SNNs is Jittered Periodic. The results correspond to an average of 10 simulations.}
\label{FormalSpiking}
\begin{tabular}{lcc}
\hline
\multirow{2}{*}{\textbf{ANN topology}} & \multicolumn{2}{c}{\textbf{Accuracy (\%)}} \\  
                                  & \textbf{Formal}          & \textbf{Spiking}       \\ \hline
784-100-10                        &    96.42           &   96.30                \\
784-200-10                        &    97.44        	&    97.29      	        \\
784-300-10                        &    97.85            &    97.74                \\
784-300-300-10              &    98.08            &    98.00               \\ 
784-300-300-300-10          &    98.35           &   98.24                \\ \hline
\end{tabular}
\end{center}
\end{table}

The ANN topologies are typical MNIST recognition topologies (784 inputs, 10 outputs) with variable hidden layer sizes. The table \ref{FormalSpiking} shows the accuracy results for each topology in both domains. These results are nearly the same in both ANN domains, with a small loss in the spiking domain. Thus, the mapping from formal to spiking domain does not significantly degrade the accuracy, which justifies, in part, the adoption of SNNs instead of FNNs. In table \ref{accu_state_art}, the accuracy results obtained in this paper are compared to the different SNNs that we found in literature. Indeed, we obtained slightly higher accuracy compared to those in \cite{du_neuromorphic_2015, mostafa_supervised_2016}. However, with 1500 fewer neurons than in \cite{diehl_fast-classifying_2015}, we have an accuracy loss of 0.36\%.

\begin{table}
\centering
\caption{Classification accuracy results of different SNNs}
\begin{tabular}{lc}
\hline
\textbf{SNN topology}       & \textbf{Accuracy (\%)} \\ \hline
In \cite{diehl_fast-classifying_2015}: \textit{784-1200-1200-10}    & 98.60 \\
In \cite{SNN_MLP_LEAT}: \textit{784-300-10}          & 95.37 \\ 
In \cite{du_neuromorphic_2015}:\textit{ 784-300-10}          & 95.40 \\
In \cite{mostafa_supervised_2016}: \textit{784-800-10}          & 97,55 \\ 
In this paper: \textit{784-300-10}          & 97.74 \\ 
In this paper: \textit{784-300-300-300-10}  & 98.24 \\ \hline
\end{tabular}
\label{accu_state_art}
\end{table}

\subsubsection*{Information coding methods}
\label{results_snn_inf_cod}

\begin{figure}
\centering
  \centerline{\includegraphics[width=\linewidth]{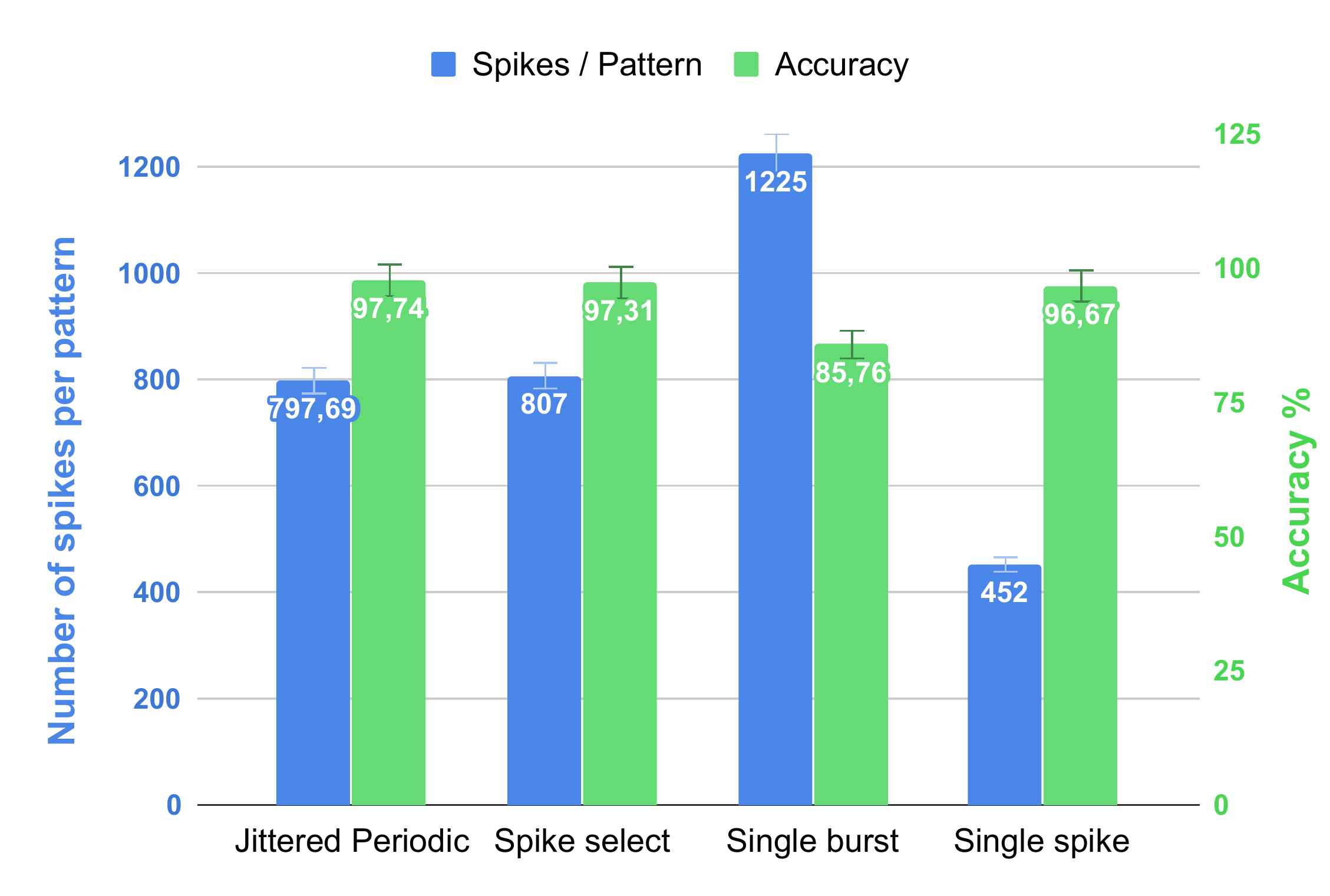}}
\caption{Information coding method impacts the accuracy and the number of spikes processed by the SNN in average for a pattern. The histogram represents results obtained for the 784-300-10 SNN topology using the MNIST test data.}
\label{inf-codi_histo}
\end{figure}

We present, in figure \ref{inf-codi_histo}, an illustration of the results obtained with the modified Rate Coding methods. The results are presented in histogram format showing the evolution of the number of propagating spikes in the network with respect to the information coding method.

The First Spike method, due to the fact that it uses only one spike per input value, mitigates the spike throughput when compared to other methods. Meanwhile, the accuracy is kept approximately the same as with Jittered Periodic method for one hidden layer SNNs, but for the "784-300-300-300-10" deeper SNN it has a loss of 11.32 \%, as shown in table \ref{SpikesInfoCod}

\begin{figure}
\centering
  \centerline{\includegraphics[width=\linewidth]{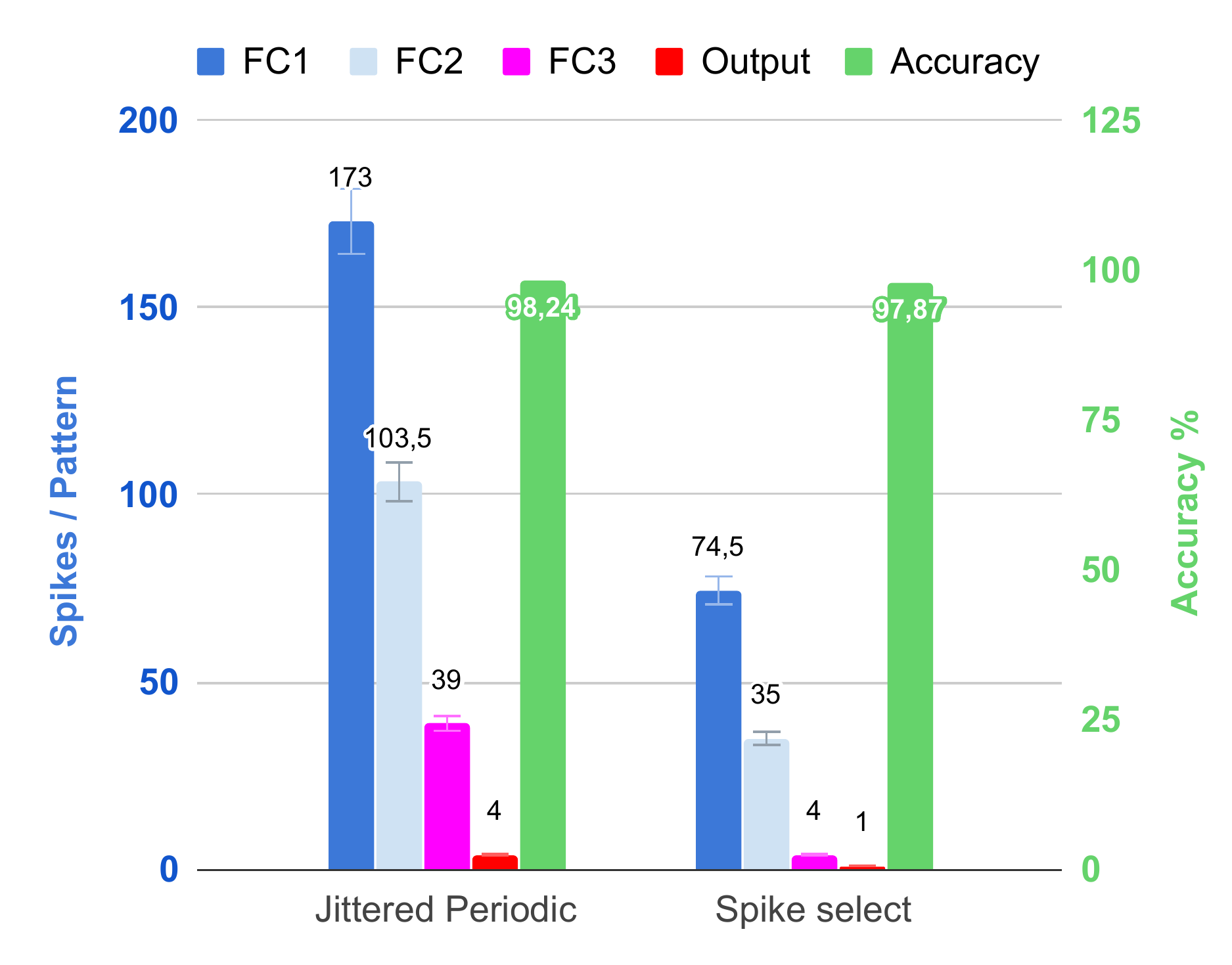}}
\caption{Spike Select versus Jittered Periodic in terms of accuracy and number of spikes generated for a pattern in SNN deeper layers. The histogram represents results obtained for the 784-300-300-300-10 SNN topology using the test data of MNIST (cf table \ref{SpikesInfoCod}).}
\label{Spike Select-layers}
\end{figure}

On an other hand, the {\it Spike Select} method is a well-tuned method for SNN hardware implementation despite the fact that it generates more spikes than Jittered Periodic and First Spike methods. Indeed, when looking at the distribution of these spikes over the SNN layers shown in table \ref{SpikesInfoCod}, we observe that they are condensed in the first hidden layer and, compared to rate coding (Jittered Periodic), refer to figure \ref{Spike Select-layers}, fewer spikes propagate in the deeper layers. Using this method, with only one spike in the output layer, we reach 97.87\% accuracy which is very close to the Jittered Periodic equivalent (98.24\%). Leaving only few spikes flowing in the remaining layers of the network, only 35\% of the spikes flow in the hidden layers compared to rate coding, Spike Select is therefore well-tailored for deep SNN hardware implementations, (cf table \ref{SpikesInfoCod}). From this perspective, as mentioned before, the use of specific architecture with a massively parallel computation for the first hidden layer, combined with multiplexed hardware for the remaining SNN layers would be an optimized solution for the Spike Select method. Such hardware architecture will be presented in section \ref{ha_section}.

%NA: nombre de spikes de Spike Select comparé à l'état de l'art
The authors in \cite{kheradpisheh_stdp-based_2018}, proposed a Spiking Deep Neural Network (SDNN) which consists in an STDP-based CNN combined with an SVM\footnote{Support Vector Machine} classifier. For an MNIST image, about 600 spikes are propagated over the SDNN that correspond to inhibitory events which occurred over the network. Since these events occur only in convolutional layer neurons, the input spikes (generated by DoG\footnote{Difference of Gaussians} cells) and the ones propagated in the classifier are not included in this amount of spikes. Therefore, despite the fact that SDNN spends fewer time steps compared to ours, the proposed SNN based on the Spike Select information coding method is more efficient in terms of hardware processing, because in average only 113.5 spikes are propagated over the network (refer to figure \ref{Spike Select-layers} and table \ref{SpikesInfoCod}). Moreover, in \cite{diehl_fast-classifying_2015} the Rate Coding based SDNN generates from $10^{3}$ to $10^{6}$ spikes in the different layers for a single MNIST image.

\begin{table}
\centering
\caption{Average number of spikes generated for processing one image by the "784-300-300-300-10" SNN with the different information coding methods. Where, JP stands for Jittered Periodic, SS for Spike Select, SB for Single Burst and FS for First Spike.}
\label{SpikesInfoCod}
\begin{tabular}{lllll}
\hline
\multirow{2}{*}{Layer} & \multicolumn{4}{c}{Spikes per pattern}                        \\  
                               & JP & SS & SB & FS \\ \hline
Input                          & 724               & 1547         & 62,5         & 170          \\ 
FC1                            & 173               & 74,5         & 363,5        & 14           \\ 
FC2                            & 103,5             & 35           & 1055         & 61           \\ 
FC3                            & 39                & 4            & 1597,5       & 87           \\ 
Output                         & 4                 & 1            & 181,5        & 4            \\ 
Total                          & 1043,5            & 1661,5       & 3260         & 336          \\ 
Accuracy \%                   & 98.24             & 97.87       & 76.80        & 86.92 \\ \hline
\end{tabular}
\end{table}

%In figure \ref{conf_matrix}, we present the confusion matrices of Spike Select and Jittered Periodic methods. Indeed, both methods behave in the same way when performing the classification task on MNIST test dataset. 

\section{Methodology for Design Space Exploration}\label{secII}

\subsection{Description of the design flow}

\begin{figure}
	\centerline{\includegraphics[width=1.6\linewidth]{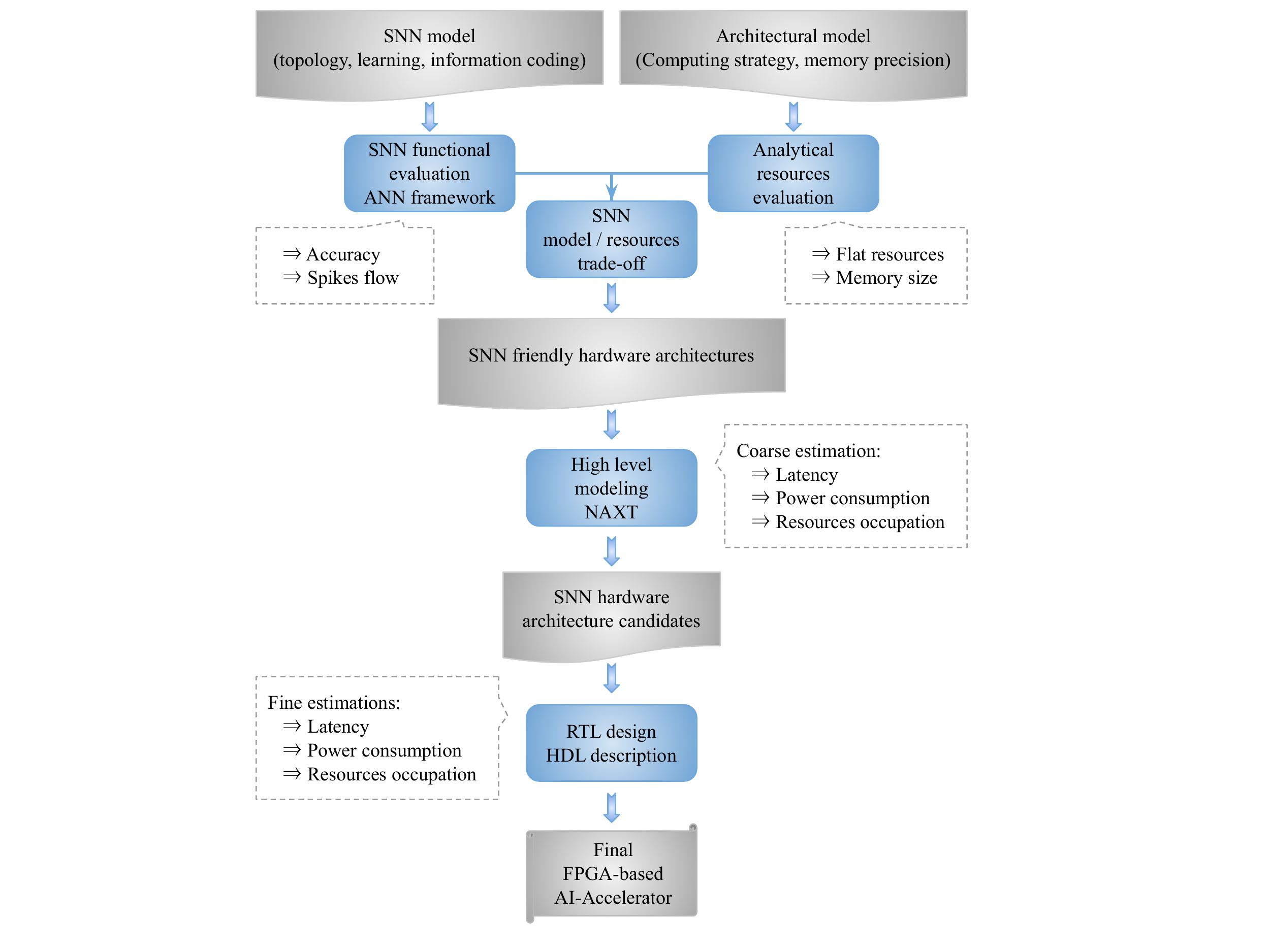}}
	\caption{Design Flow Diagram. First stage: Perform a functional evaluation of different SNN models using an ANN framework (N2D2). Simultaneously, using an analytic model evaluate the cost (memory size and flat resources) of some architectural models. Then, make a model / resources trade-off to select some SNN friendly hardware architectures; Second stage: High-level simulation of the chosen topology with different architectural choices, to select the architectural paradigm for the last stage; Third stage: Develop RTL designs of the SNN hardware architecture candidates to have finer cost estimations. Then, based on these results select one architecture as the final FPGA-based AI-Accelerator.}
	\label{work_plan}
\end{figure}

In this section, the adopted design flow methodology will be described. This design flow is synthesized in figure \ref{work_plan}. It follows a funnel philosophy: knowing the application context, we start from a wide variety of possible hardware implementations and incrementally refine the scope to find the most suitable at the end. In our case, the example application context will be image classification.

First, a behavioral software simulation using the N2D2 framework \cite{N2D2}  (available online at: https://github.com/CEA-LIST/N2D2) is carried out to perform learning, test and validation for several SNN topologies with different information coding methods. The most suitable model in terms of prediction accuracy and spiking activity (the amount of spikes processed by the SNN to perform classification inference) is selected for the following steps, and the learned parameters are extracted. A preliminary analytic study is carried to get the first estimations of flat hardware resources and memory intensiveness corresponding to the chosen SNN model: these first results will serve as a frame for the next steps of our design flow, giving indications for the most suitable architectural choices and hardware target.

%EDGAR STYLE DEBUT
Second, we perform a high-level architectural exploration aiming to confirm or invalidate the assumptions resulting from the preliminary analytic study. The NAXT simulator is configured with the model and parameters extracted from N2D2. The software will generate systemC architectures corresponding  to different high-level architectural choices, such as memory distribution, memory technology and processing parallelism. It then performs high-level simulation of their operation on the specific user-defined application task. For each simulation, we obtain coarse estimations for power consumption, surface and latency: those results allow to discriminate suitable architectural choices which will be used in following steps.

Third, a precise hardware description of the architecture is made, according to NAXT results, and using the parameters extracted from N2D2. The architecture is described in VHDL \cite{navabi1997vhdl}, and a physical synthesis is performed. Thus, this last step leads to a fine-grained evaluation of a suitable architecture on FPGA (Field-Programmable Gate Arrays) or on ASIC (Application Specific Integrated Circuit).
%EDGAR STYLE FIN

%BM: FPGA synthesis simulation n'existe pas. On distingue logical synthesis => logic gates et physical synthesis => place & route
%=> physical synthesis
%EL : ok

\subsection{Hardware targets of the DSE}

The present work aims to deliver an architecture for Spiking Neural Network hardware implementation. To this end, two digital hardware targets are considered: Field-Programmable Gate Arrays (FPGA) and Application Specific Integrated Circuit (ASIC).

\subsubsection{FPGA}

In previous studies, FPGAs have been frequently employed for the design of neuromorphic computing circuits \cite{pani2017fpga} \cite{rotermund2018massively}. This technology can be used either for prototyping and delivering a sub-part of a greater system, or directly as the final chip design implementation. The main advantages of this technology are its high programmablity, high reconfigurability, and moderate cost. As our objective is Design Space Exploration, we are interested in a reconfigurable platform : indeed, the chip must be reconfigured for each architecture. Thus, an FPGA device is an adequate technology for our purpose.

Some devices, namely SOCs (System On Chips), include one or several CPUs of the alongside with the Programmable Logic array, which offers possibilities for both software and hardware programming. As we aim to develop a general-purpose neuromophic IP capable of executing any feed-forward SNN configuration, this type of device would suit the programmability requirement. In the present work, only Programmable Logic part has been used. However, we intend to use both in future studies, with a CPU (ARM-based) acting as a master responsible for FPGA reconfiguration and computation scheduling, and the Programmable Logic acting as a slave dedicated to inference processing.

\subsubsection{ASIC}

ASIC chips have also been widely employed for neuromorphic digital hardware implementations (see \ref{SOTA}). In contrast with conventional processor architectures, which are designed to handle a wide variety of tasks, ASICs are fully customized to run a particular type of application. Some of those chips, such as TrueNorth \cite{merolla2014million} \cite{akopyan2015truenorth}, are very highly specific: they are designed for one particular neuron model with very low programmability, whereas others such as SpiNNaker \cite{furber2014spinnaker} \cite{painkras2013spinnaker}, are designed with a much higher capacity for flexibility. Usually, these chip architectures are designed to support the high level of parallelism and distribution found in neural algorithms. Thus, most of the time they are based on a massively parallel computation paradigm, with great care given to the communication between computing units. However, these ASICs focus not only on pure computation acceleration, but also on the constraints of their application domain.

For integration in embedded systems for example, particular attention has to be paid to the chip surface and energy consumption limitations. These application-related constraints are also of major concern for ASIC design, and can be found in the differences between TrueNorth and SpiNNaker: the first is focused on energy savings, whereas the second is focused on flexibility. Even if the task is very similar, the implementation design is dramatically different, and so are performances: TrueNorth \cite{merolla2014million} \cite{akopyan2015truenorth} shows an energy consumption of 12pJ per connection, in contrast to 20nJ for SpiNNaker \cite{furber2014spinnaker} \cite{painkras2013spinnaker}. In this paper, design space exploration requires a high programmability and reconfigurability, and we have thus targeted FPGA design instead of ASIC. By this way, we favour the automatic generation architectures on a reconfigurable substrate rather than the definition of a programmable architecture on a fixed one. 

\subsection{N2D2 framework description}
\label{n2d2}

The ANN models used in this paper are learned, validated and tested using the open source Neural Network Deployment and Design (N2D2) framework \cite{N2D2}. This software is an event-based simulator for DNNs. A wide variety of deep-learning frameworks for design and deployment of ANNs have been described in the literature \cite{DNN_Learning}\cite{SpykeTorch}. However, we selected N2D2 essentially for two reasons : First, it is an open source solution that gives the ability to develop new methods without designing a whole simulator. Second, N2D2 offers the possibility to transcode and test ANNs into spiking domain, which is essential for our purpose.
In order to perform simulations of SNN with N2D2, we followed these steps:

\begin{enumerate}
\item Define FNN topology;
\item Learn, validate and test the defined FNN;
\item Define a transcoding method to generate the SNN;
\item Test the SNN defined in 3.
\end{enumerate}

Note that our configuration parameters are listed in table \ref{tableParam}. We have chosen Xavier Filler as a Weight Initialization method as it is a popular method which offers state-of-the-art performance \cite{glorot2010understanding}. Moreover, we have chosen to implement Rectifier activation function \cite{krizhevsky2012imagenet} \cite{nair2010rectified} in our hidden layers as this model offers state-of-the-art classification performance according to literature \cite{krizhevsky2012imagenet}. Usually, in ANNs, a Softmax layer is used at the output for classification purpose. However, Softmax layers are difficult, if not impossible to implement in spike domain \cite{rueckauer2016theory}. For this reason, we replace this classification method by a Linear activation function in the output layer\cite{diehl_fast-classifying_2015}, completed by a Terminate Delta procedure to determine the winning class (see Part \ref{tdelta}).

%EDGAR STYLE DEBUT
Once the simulation is complete, if the network prediction accuracy is satisfying, the network parameters are ready to be extracted for use in the following steps of the Design Space Exploration.
%EDGAR STYLE FIN

\begin{table}
\centering
\caption{ANN learning hyperparameters used in this work. *LR = Learning Rate}
\label{tableParam}
\begin{tabular}{lll}
\hline
Hyperparameter               & Value                                       \\ \hline
Weight Initialization   & Xavier Filler                               \\ 
Activation Function     & Linear (last layer), Rectifier (others)     \\ 
Learning Rate           & 0.01                                        \\ 
Momentum                & 0.9                                         \\ 
Decay                   & 0.0005                                      \\ 
LR* Policy              & Step Decay                                  \\ 
LR* Step Size           & 1                                           \\ 
LR* Decay               & 0.993                                       \\ \hline
\end{tabular}
\end{table}

\subsection{Analytic preliminary work} 
\label{Analytics} 

In this subsection, we will present some preliminary results that will drive our further investigations. These results deal with on-chip memory capacity and resource restrictions, which have to be taken in account upstream. Indeed, both of these restrictions will have strong influence on our future choices in terms of architectural model and hardware target.

\subsubsection{Memory capacity: a limiting factor}
\label{mem_capacity_sec}

On-chip memory capacity will always be limited, no matter which target hardware we choose. Indeed, even if the most recent FPGA devices such as {Xilinx\textregistered} {Virtex\textregistered} {Ultrascale\texttrademark} + and {Intel\textregistered} {Stratix\textregistered} 10 reach huge on-chip memory capacity, it remains inadequate to deal with most of the neural network models. For information, state-of-the-art classifiers such as VGG16 require a total of 230 MB for weight storage \cite{simonyan2014very}.
%BM: je pense qu'avec le démarrage de la section sur les mémoires on-chip de 64Mb pour xilinx et 90Mb pour Altera, il serait nécessaire de faire des histogramme pour de beaucoup plus gros réseaux (1024 neurones par couche) pour arriver à ces montants. Sinon on est en décalage.
%EL OK

\begin{table}[]
\caption{Memory footprint of synaptic weight storage with respect to coding precision estimated with our analytic model}
\label{tableMemory}
\begin{tabular}{ll}
\hline
Weight coding precision                                                                 & Memory footprint \\ \hline
Binary (1 bit)                                                                          & 238 kb           \\ 
Ternary (2 bits)                                                                        & 476 kb           \\ 
\begin{tabular}[c]{@{}l@{}}8 bits (TF Lite minimum precision / \\ Our work)\end{tabular} & 1.9 Mb           \\ 
16 bits                                                                                 & 3.8 Mb           \\ 
32 bits (Half Precision Floating Point)                                                 & 7.6 Mb           \\ 
64 bits (Floating Point)                                                                & 15.2 Mb          \\ \hline
\end{tabular}
\end{table}

\begin{figure}
	\centerline{\includegraphics[width=0.9\linewidth]{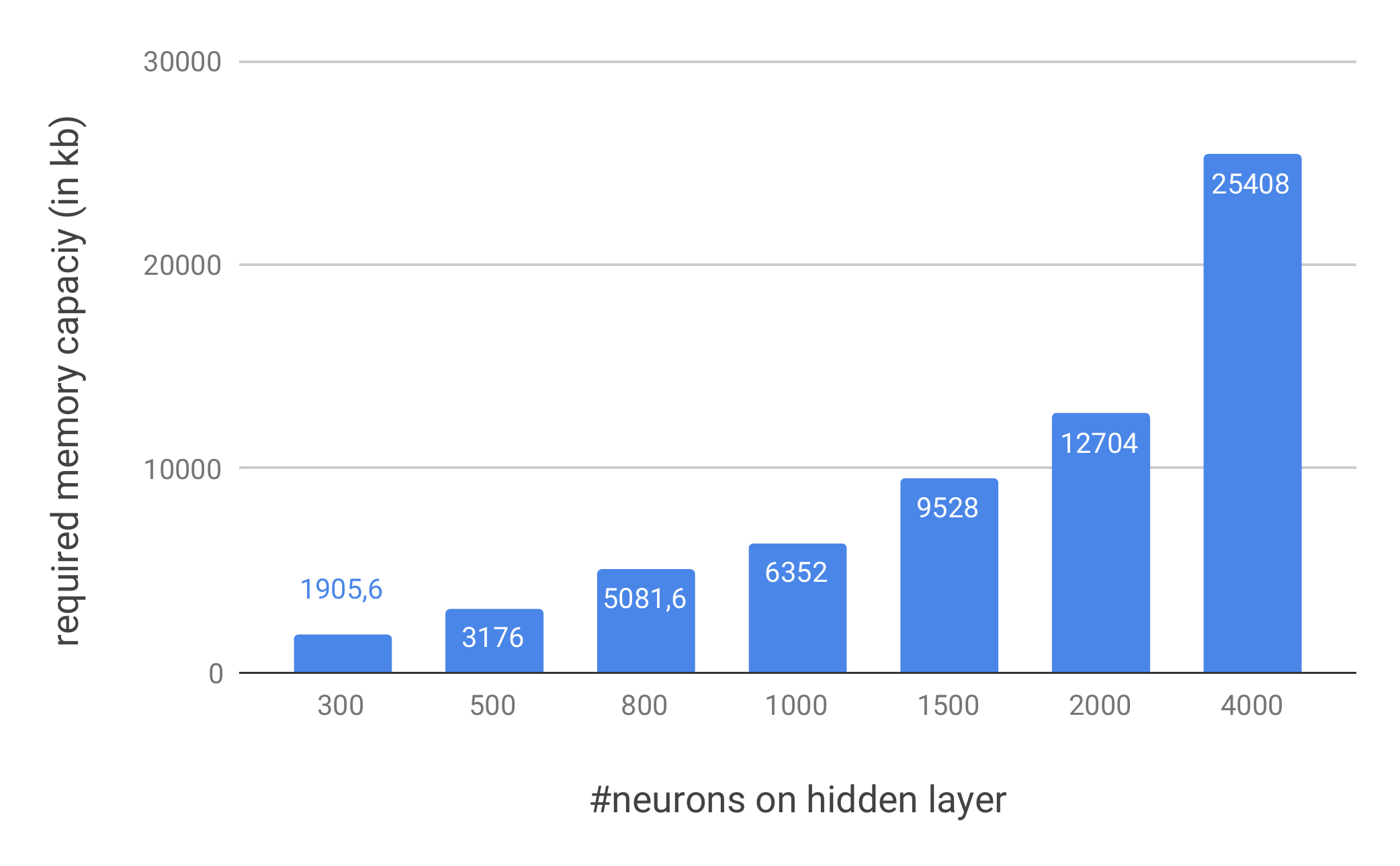}}
	\caption{Required memory capacity for a 3 layers classifier with respect to the number of neurons in the hidden layer. Synaptic weights are coded on 1 Byte each. The network has 784 input neurons, and 10 output neurons (typical ANN for testing on MNIST database).}
	\label{memoryneurons}
\end{figure}

\begin{figure}
	\centerline{\includegraphics[width=0.9\linewidth]{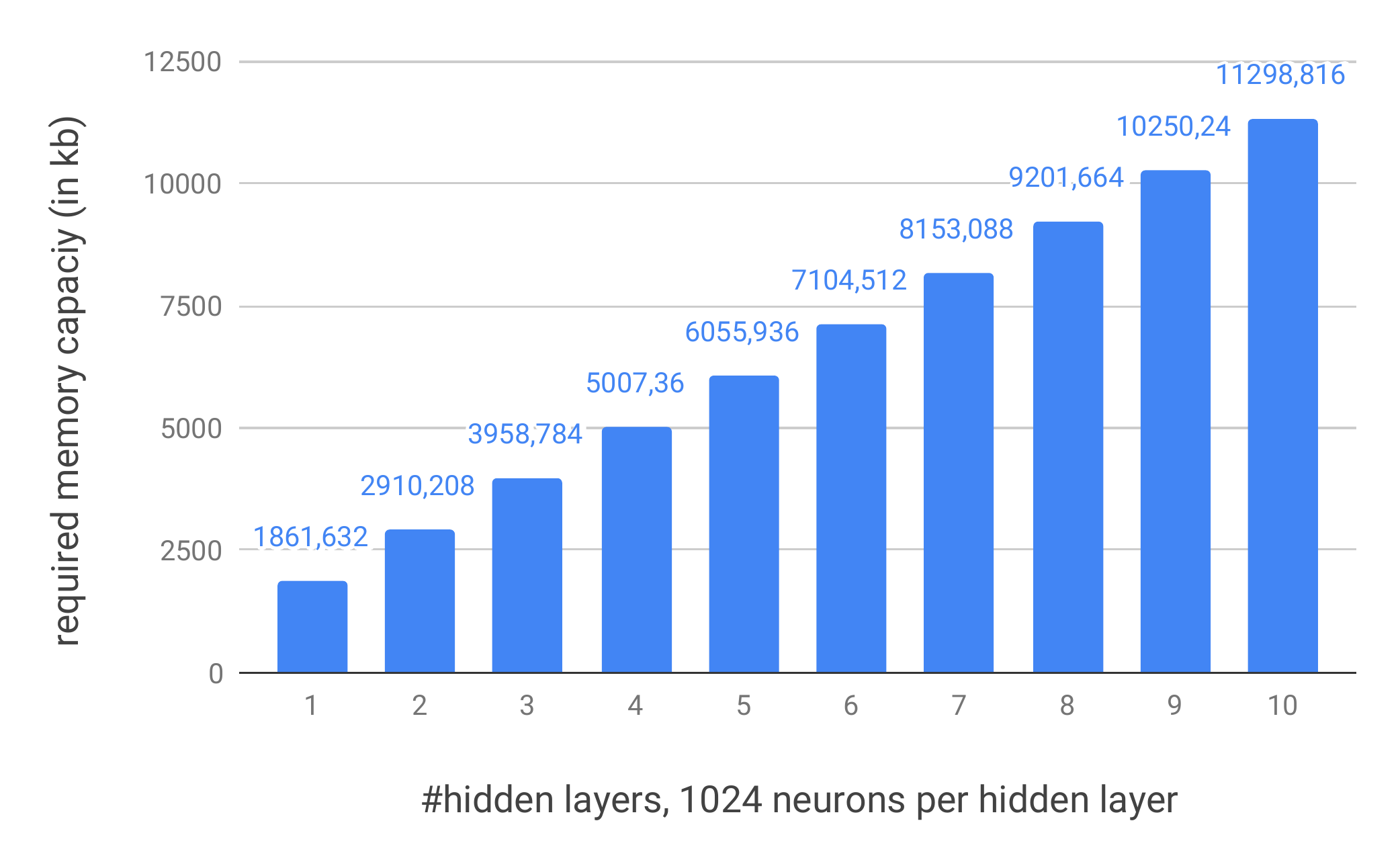}}
	\caption{Required memory capacity for a n-layer-classifier, 1024 neurons per hidden layer, with respect to the number of hidden layers. Note that, 1 Byte is used to encode 1 weight and there are 784 input and 10 output neurons.}
	\label{memorylayers}
\end{figure}

%BM: il faut ajouter une phrase dans ce paragraphe pour préciser que la mémoire de poids nécessaire pour un SNN est la même que pour un ANN en rappelant en quoi consiste le transcodage de l'étape 3 précédente
%EL OK
Consequently, we investigated the evolution of the required memory capacity with respect to the number of implemented synaptic connections. Our analytical model for memory capacity is based on the total memory footprint of network parameters, in our case: synaptic weight storage. Hence, our analytical approach is related to the parameters coding precision: in our case, we have chosen an 8 bits precision, as it offers a good trade-off between memory footprint and prediction accuracy; but our results can be generalized to other parameters-coding precision with a simple cross product calculation. The analytical results are represented in figures \ref{memoryneurons} and \ref{memorylayers}. Figure \ref{memoryneurons} depicts the evolution of required memory  capacity  for a 3-layer-spiking-classifier, with respect to the number of neurons in the hidden layer. On the other hand, figure \ref{memorylayers} depicts the evolution of required memory capacity for an n-hidden-layer-classifier, with respect to the number of hidden layers (each hidden layer is 1024 neurons wide).
%EDGAR STYLE DEBUT
Note that the memory required to store synaptic weights is the same for FNNs and SNNs: the transcoding method presented in section \ref{n2d2} does not affect the synaptic weights.
%EDGAR STYLE FIN

%EDGAR STYLE DEBUT
According to the results illustrated in figure \ref{memoryneurons} and \ref{memorylayers}, the required memory capacity increases drastically with respect to the number of neurons, reaching several MBs for large-scale networks. The difference with our theoretical results and VGG16 memory requirement is due to the difference in number of neurons, weight coding precision, and the absence of convolutional layers in our estimations. Consequently, on-chip memory capacity is a major limiting factor for hardware SNN implementation, and has to be taken in account quite early in the design flow. Indeed, the hardware target must offer sufficient on-chip memory capacity to store model parameters. Those results also induce that intelligent memory management policy might become necessary (for example, cache hierarchies), when implementing very large models such as VGG16. Such implementation would mitigate memory footprint, even though this might result in slowing down the system and increasing logic resources intensiveness.
%EDGAR STYLE FIN
Moreover, our model allows to evaluate the influence of weight coding precision on the memory footprint. Results are presented for a classic 784-300-10 MNIST classifier in Table \ref{tableMemory} for various coding precision, from Binary coding to full-precision floating point (64 bits). Those results are interesting to choose a coding precision satisfying hardware target requirements, or \textit{vice versa}.

 %By our estimations, we only reached 9.5 MB }because we considered a maximum of 10 layers of 1024 neurons each, which corresponds to a total of 10.5 million synaptic weights to be stored}. For VGG16, the classifier stage alone implements roughly 16.7 billion} synaptic connections, explaining the difference between our estimations and VGG16 memory requirements. 
\subsubsection{FPGA occupation: towards multiplexing}
\label{FPGA_occup}
%BM: la figure fpga_resources n'est pas dans le PDF ?!
%ok!
Logic resources occupation is the second limiting factor we encounter when implementing hardware SNN on FPGA devices. The FPGA occupation statistics can be obtained by FPGA synthesis simulation tools. However, this synthesis requires a long processing time, especially when synthesizing a large-scale network. In order to bypass this long processing time, we have built an analytical model capable of estimating the number of logic cells occupied on an FPGA according to the network topology and size.

\begin{figure}
	\centerline{\includegraphics[width=0.85\linewidth]{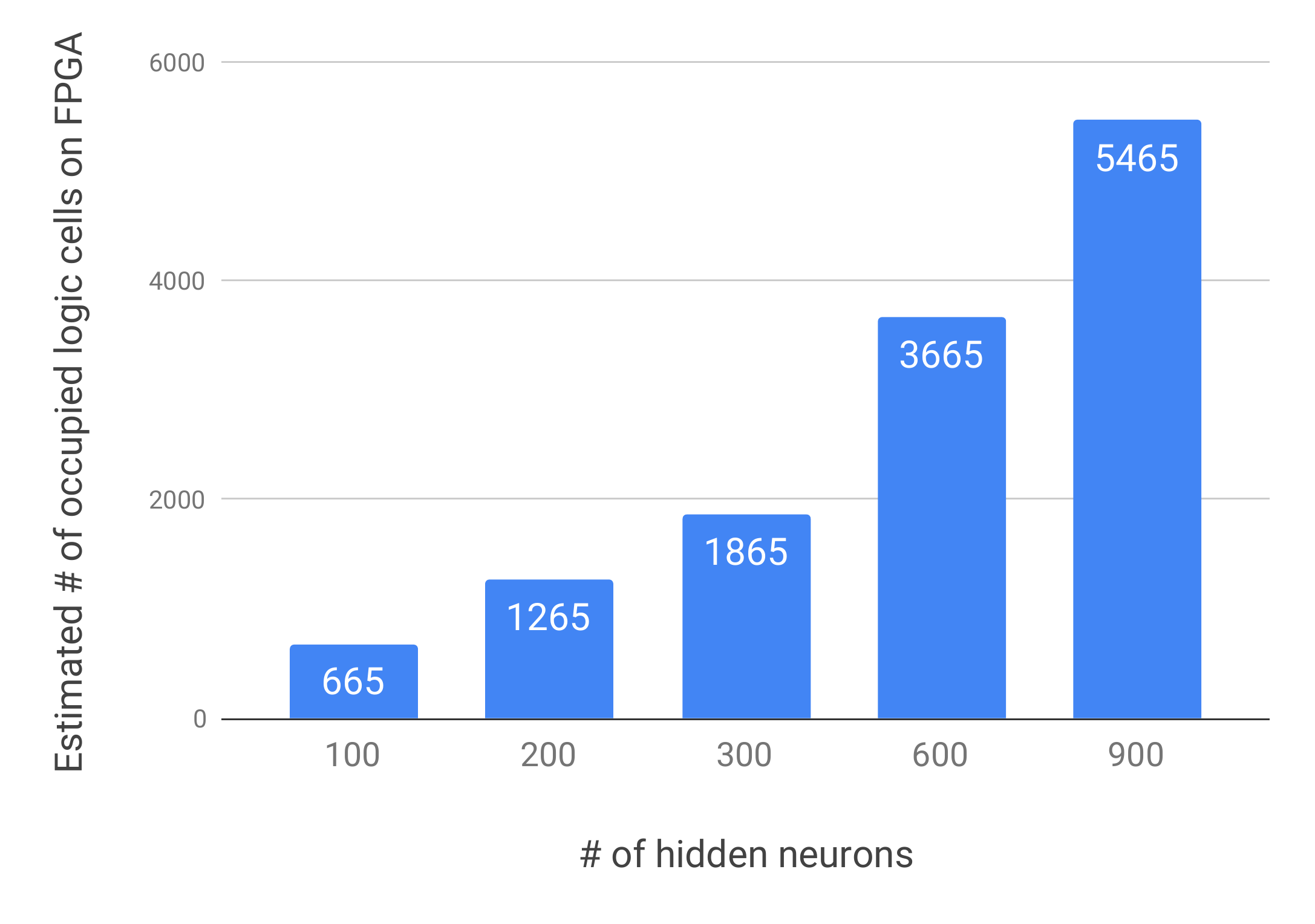}}
	\caption{Theoretical values for FPGA occupation against hidden layer size. 784 input and 10 output neurons.}
	\label{FPGA_Logic_operators}
\end{figure}

To build our analytical model, we have separated a generic SNN hardware architecture in elementary modules (neurons, spike generation cell, counter, etc.). For each elementary element, we have measured corresponding hardware implementation cost in terms of logic cells, using Quartus Prime 18.1.0 Lite edition. Quite straightforwardly, every SNN topology is then expressed as a combination of those elementary modules, and hence can be related to an estimation of its flat hardware implementation cost (note that some part of the system can be multiplexed in the final design, but this model only outputs the flat hardware resources as an indicator). The results of our analytic model for a fully-parallel implementation of an SNN with 784 inputs, 10 outputs and a variable number of 100-neuron-hidden-layers, are shown in figure \ref{FPGA_Logic_operators}. As depicted in the figure, FPGA occupation grows drastically with respect to network size. Note that this model does not reproduce organization optimization performed by the synthesizer (for example, a single logic unit can be used to perform two different functions in some cases), as it can be seen in figure \ref{theory_experiment} which compares experimental and theoretical results. However, this model is sufficient to give a proper estimation of FPGA occupation against network size in the early stages of our design space exploration.

The analytic model shows that, consistent with our expectations, such fully-parallel implementations face FPGA capacity limits: according to the model, a fully-parallel implementation of 900 IF-neurons would cover 5465 logic cells on FPGA. %, which is already reaching the limits of most FPGA devices. 
%BM: Les stratix 10 contiennent plus de 1 Million de Logic Elements : https://www.intel.com/content/dam/www/programmable/us/en/pdfs/literature/pt/stratix-10-mx-product-table.pdf
% J'ai donc couper la fin de la phrase 
%EL: Ok
Compared to real convolutional networks, which present several tens of thousands of neurons (65,000 for AlexNet \cite{alexnet}), it is quite obvious that the fully-parallel implementation paradigm is not viable when using FPGA devices. Moreover, when using ASIC technology instead of FPGAs, chip size would drastically grow with network size, as would production costs. Therefore, we assume that time-multiplexed architectures are way more viable when dealing with the hardware implementation of deep SNN.

On the other hand, time-multiplexing consists in implementing fewer neurons in hardware than there is in the model : each hardware neuron will thus operate successively for several neurons of the model. This method results in slowing down computation (as shown in section \ref{exploration}), but allows one to implement large-scale networks with fewer resources, notably for FPGA implementation or cost reduction purposes. Those assumptions will be evaluated in further steps of our design flow (see sections \ref{exploration} and \ref{DHSNNA}).

\section{High-level architecture exploration}  \label{exploration}

In this section, we are going to introduce a tool developed by our research team, namely "NAXT" for Neuromorphic Architecture eXploration Tool, which aims to simulate SNN hardware implementations with various architectural choices, such as processing parallelism, memory distribution and memory technology. The goal is to match application specific constraints (power, consumption, logic resources) with high-level architectural choices. The simulator is configured with the SNN parameters extracted from N2D2 (topology and learned synaptic weights) and with user-defined architectural choices (i.e., level of multiplexing, level of memory distribution and memory technology). It subsequently generates a SystemC code corresponding to those parameters, and performs inference on a test data-set. The simulator estimates the chip surface, average latency and energy-consumption per inference. Hence, the role of NAXT simulator in our funnel-like architectural exploration workflow is to easily and quickly provide coarse estimations for different architectural paradigms. Although RTL modeling gives much finer results, it requires a long design and development time. Therefore, NAXT simulator is used to clear the path, as its results will guide further and finer architectural exploration in following steps of the workflow. Hence, the NAXT simulator is quite innovative as it brings hardware estimation at a very early stage of a design flow, basing on a functional description of the network. 
%BM2: cette phrase est un "claim" : analogous to GA. Y a t il des chiffres qui le prouvent ?
%EL : C'est une analogie qualitative, dans NAXT on travaille avec la surface des élèments de base (surface d'un NPU, surface d'une mémoire...), mais on s'intéresse ici aux FPGA. On aurait pu simplement changer dans NAXT l'unité et considérer un "nombre de logic cells" en lieu et place d'une surface, les résultats auraient été qualtitativement similaire. Enfait, le resultat cest "hardware resources" qui se traduit en surface dans NAXT (cible ASIC) mais qu'on veut ici exprimer en terme de FPGA occupation. J'ai modifié pour que ce soit plus clair.
% OK
Note that the chip surface estimations are relative to an ASIC target, and are analogous to Gate Array occupation for an FPGA target. Indeed, those two metrics are relative to the same "hardware resource" evaluation: a hardware resource can be seen as a piece of circuit from the ASIC point-of-view, or as a group of logic cells from an FPGA point-of-view. Accordingly, chip surface estimations can be taken for FPGA occupation qualitative estimations.

\subsection{SystemC modeling}

To develop our simulator, we used SystemC \cite{panda2001systemc}, a behavioral-level hardware description library for C++. This language is often chosen for architectural exploration purposes at a high-description-level, as it enables simple functional system description, overcoming the usual finer-description-level constraints (transaction modeling, etc.). SystemC enables users to develop functional modules that run concurrent processes and communicate with each other via signals. Thus, we developed three different modules as "elementary bricks" of our hardware architecture models: a \textit{Neural Processing Unit module}, a \textit{Memory Unit module}, and an \textit{Input module}. Before we start a more precise description of each module, it is important to note that our simulator was developed according to a synchronous paradigm: every process is executed at a clock rising edge. The \textit{clock signal} is generated by the \textit{Input module}. Despite our enthusiasm for asynchronous processing, we have chosen a synchronous simulation paradigm for the purposes of coding ease. We plan to enable asynchronous processing simulation in future development of NAXT simulator.
 
\subsubsection{Neural Processing Unit module}

The \textit{Neural Processing Unit module} (NPU) is basically a digital implementation of a spiking neuron. Thus, it is fully dedicated to the Integrate and Fire task. At every \textit{clock rising edge}, it integrates synaptic weights corresponding to spikes received during the last cycle. The integration is done in a simple accumulator. After integration, the accumulator value is compared to the membrane's threshold value: if the threshold is exceeded, a spike is emitted at the neuron's output, and the accumulator is reinitialized. If not, it waits until next \textit{clock rising edge} to start a new integration, and so on.
 
\subsubsection{Memory Unit module}

Synaptic weights are stored in \textit{Memory Unit modules}. Thus, \textit{NPUs} must access \textit{Memory Units} whenever a spike is integrated. As our architectures work on a synchronous paradigm, integration processes are run simultaneously by all \textit{NPUs}. Consequently, a \textit{Memory Unit} can receive several access requests at the same time, but real \textit{Memory Units} can only answer one request at a time. Thus, our \textit{Memory Unit} model focuses on this aspect: this module must store incoming requests in the right order, and answer those requests one by one in that same order.
 
\subsubsection{Input module}

The \textit{Input module} is dedicated to input image transcoding. Indeed, we have to translate input data from the formal domain to the spiking domain. Various spike coding techniques exist, see section \ref{information_coding}.
Each input image pixel is associated with an input neuron. Thus, the \textit{Input module} is responsible for input data transcoding and spike train injection into input neurons. Ultimately, this module should disappear as we aim to simulate and evaluate a fully spiking implementation, with true spiking data coming from an asynchronous camera, for example. 

\subsection{Parallelism and distribution}
% NA: it was "Parallelism / distribution vs. multiplexed executions"

As previously described, there are two main exploration rungs available in the NAXT simulator. The first is processing parallelism. Indeed, as SNNs are intrinsically parallel algorithms, computation parallelization should result in great acceleration of processing. On the other hand, a high level of parallelization requires a large number of \textit{NPUs} (ideally, one per logical neuron), resulting in the drastic increase of chip surface (i.e., FPGA occupation). This first level of exploration thus allowed us to evaluate the trade-off between chip surface savings and processing acceleration.
In the NAXT simulator, this exploration level is modeled by two different architectural paradigms: \textit{Fully-Parallel Architectures}, and \textit{Layer-Multiplexed Architectures}.
 
\subsubsection{Fully-Parallel Architecture}

\textit{Fully-Parallel Architecture} (\textit{FPA}) in NAXT stands for the extreme case where every logical neuron in the algorithm is implemented by an \textit{NPU} on the chip. This architectural choice should result in fast processing but a large area.

\subsubsection{Time-Multiplexed Architecture}

\textit{Time-Multiplexed Architecture} (\textit{TMA}) in NAXT simulator stands for the case where each layer is composed of only one \textit{NPU}. This is quite an arbitrary choice, as we could have chosen one single \textit{NPU} for the whole network as an extreme case, but this would be the equivalent to conventional \textit{Central Processing Unit (CPU)} architectures, which is not interesting as we want to explore innovative neuromorphic architectures. In future work, we plan to let the user choose the number of \textit{NPUs} per layer, for flexibility and finer exploration purposes. Multiplexed architectures should result in slower processing, but will be interesting in terms of chip area savings.
 
\subsection{Memory organization}

\label{memory}
The second rung of architectural exploration in our simulator is memory  distribution. Thus, three levels of memory distributions have been developed: a \textit{Centralized Memory} architecture (one \textit{Memory Unit} for the whole network), a \textit{Layer-Shared Memory } architecture (one \textit{Memory Unit} per layer), and \textit{Fully-Distributed Memory} architecture (one \textit{Memory Unit} per \textit{NPU}). Note that in the case of \textit{TMA}, in the current version of NAXT simulator, layer-shared and fully-distributed memory organizations are the same (1 \textit{NPU} per layer = 1 \textit{Memory Unit} per layer in both cases).

\begin{figure} 
	\centerline{\includegraphics[width=\linewidth]{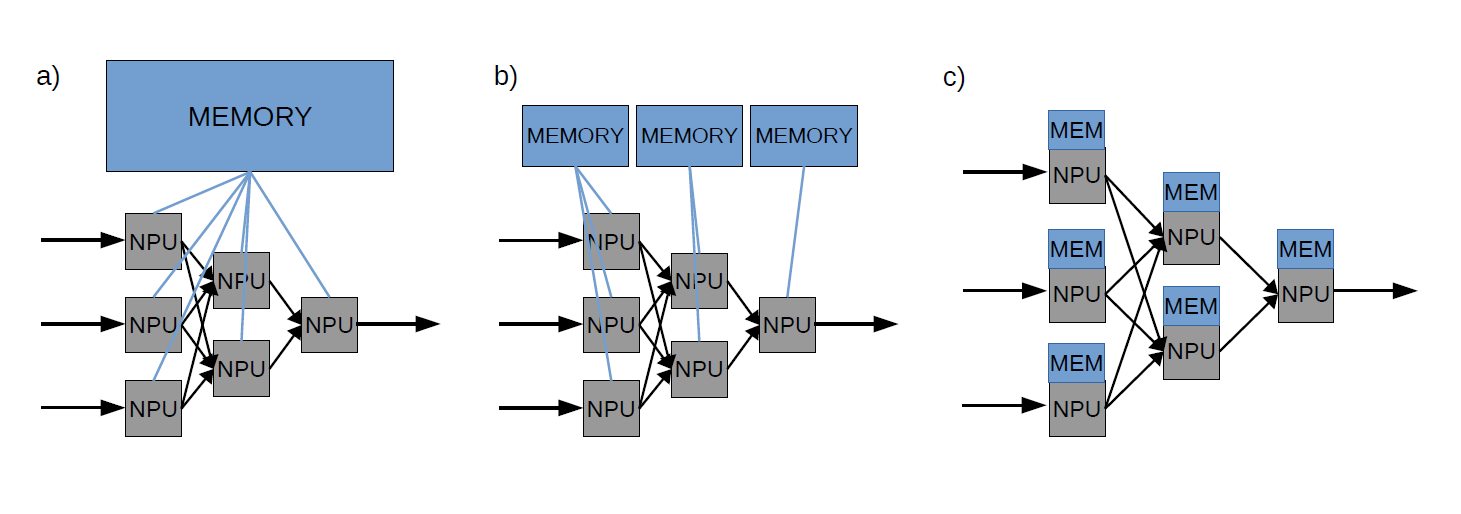}}
	\caption{Representation of all memory organizations for fully-parallel architectures: a) centralized memory unit, b) layer-shared memory units, c) fully-distributed memory units}
	\label{MemoryOrganization1}
\end{figure}

\begin{figure}
	\centerline{\includegraphics[width=\linewidth]{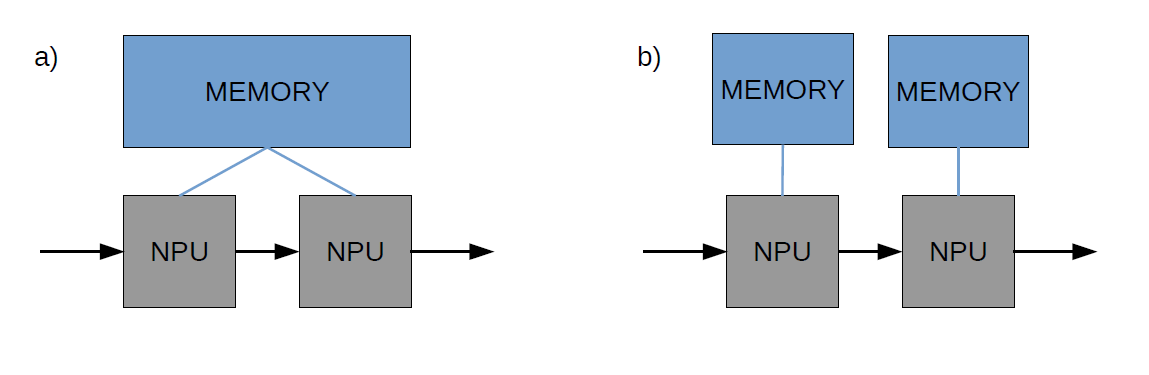}}
	\caption{Representation of all memory organizations for TMA: a) centralized memory unit, b) layer-shared/fully-distributed memory units}
	\label{MemoryOrganization2}
\end{figure}

These three different memory architectures allow users to evaluate, once again, the trade-off between processing latency, energy consumption and chip surface (i.e., FPGA occupation). For example, a \textit{Centralized Memory} architecture will be more compact than a multitude of \textit{Distributed Memories}, but will slow down processing as it can only answer one single \textit{NPU} request at a time. \textit{Layer-Shared Memory} architecture is an intermediate between both architectures. 
Figure \ref{MemoryOrganization1} depicts all memory distribution levels for fully parallel architectures, and figure \ref{MemoryOrganization2} shows all memory distributions for multiplexed architectures.

\subsection{Latency, Power and Surface estimations}
\label{estimations}
The aim of NAXT Simulator is to give an estimation of power, latency and logic resources for a user defined SNN topology considering different architectural paradigms. To do so, estimations are performed \textit{a posteriori}, using traces generated during inference simulation. More precisely, during inference, all events are recorded: spike emission, read memory access, write memory access, etc.. These records, alongside with the number of clock cycles spent for processing, constitute the trace used for estimations.

\subsubsection{Latency}
Latency is calculated as the product of the number of clock cycles spent for processing and the clock period. The number of clock cycles being recorded in the trace file, we only have to estimate the clock period. To do so, we have chosen to constrain the clock period to the maximum memory access latency, as it is often the limiting factor in a non-pipelined architecture like ours (worst-case critical path). This latency is estimated using NVSim \cite{dong2012nvsim}, an open-source software aiming to simulate memory behavior for different memory technologies and technology nodes, which returns various estimations, including memory access latency.

\subsubsection{Hardware Resources}
The Hardware resources estimation is calculated in a similar fashion than in \ref{FPGA_occup}: we separate our architecture in elementary modules, for each of which we measure the hardware implementation cost in terms of resources. Each architecture is expressed as a combination of elementary modules, and hence can be related to a global hardware resources cost estimation. 

Note that this estimation method does not take in account placement and routing optimizations performed by FPGA design softwares (\textregistered{Quartus}, \textregistered{Vivado}, etc.).

\subsubsection{Power}
The power estimation is calculated as a sum of energy consumption of the two main subparts of the system: memory and processing. Concerning memory, static and dynamic energy consumption of Memory Units are extracted from NVSim offline simulations. Static energy consumption of Memory Unis are then multiplied by the total inference latency, and summed together. Dynamic energy consumption are multiplied by the number of memory accesses (read and write), and summed together. Both of those results are summed together to give the total power estimation for memory units. Concerning Neural Processing Units, we have taken from literature (\cite{mayr2015biological}) the average energy consumption per spike of a state-of-the-art hardware digital spiking neuron. This average energy consumption per spike is multiplied by the number of spike emitted during inference to obtain a gobal power estimation of Neural Processing Units. Although this power estimation method is not directly related to our developed Neural Processing Unit architecture, it is a relevant approximation, as it concerns state-of-the-art hardware digital spiking neurons.
Finally, both power estimations (for Memory Units and for Neural Processing Units) are summed together to give a global power estimation for the whole system.
This power consumption evaluation method is quite approximate and thus gives coarse estimations, hence it should be improved in future works.

\subsection{Results}

Here, we show some data obtained with the NAXT simulator. Note that these estimations are made \textit{a posteriori} thanks to the network activity traces (the number of spikes processed by each NPU, the number of memory accesses per memory unit, etc.). Simulations have been run for a relatively small network "784-10-10". Our simulator achieves $62\%$ accuracy, which roughly corresponds to the equivalent N2D2 recognition accuracy for the same network. NAXT performs latency, surface and power estimations based on traces generated during processing: during each inference, we record the spiking activity of each \textit{Neural Processing Unit}, alongside all memory accesses for each \textit{Memory Unit}. Memory-related estimations have been computed using SRAM technology.

\begin{figure}
	\centerline{\includegraphics[width=\linewidth]{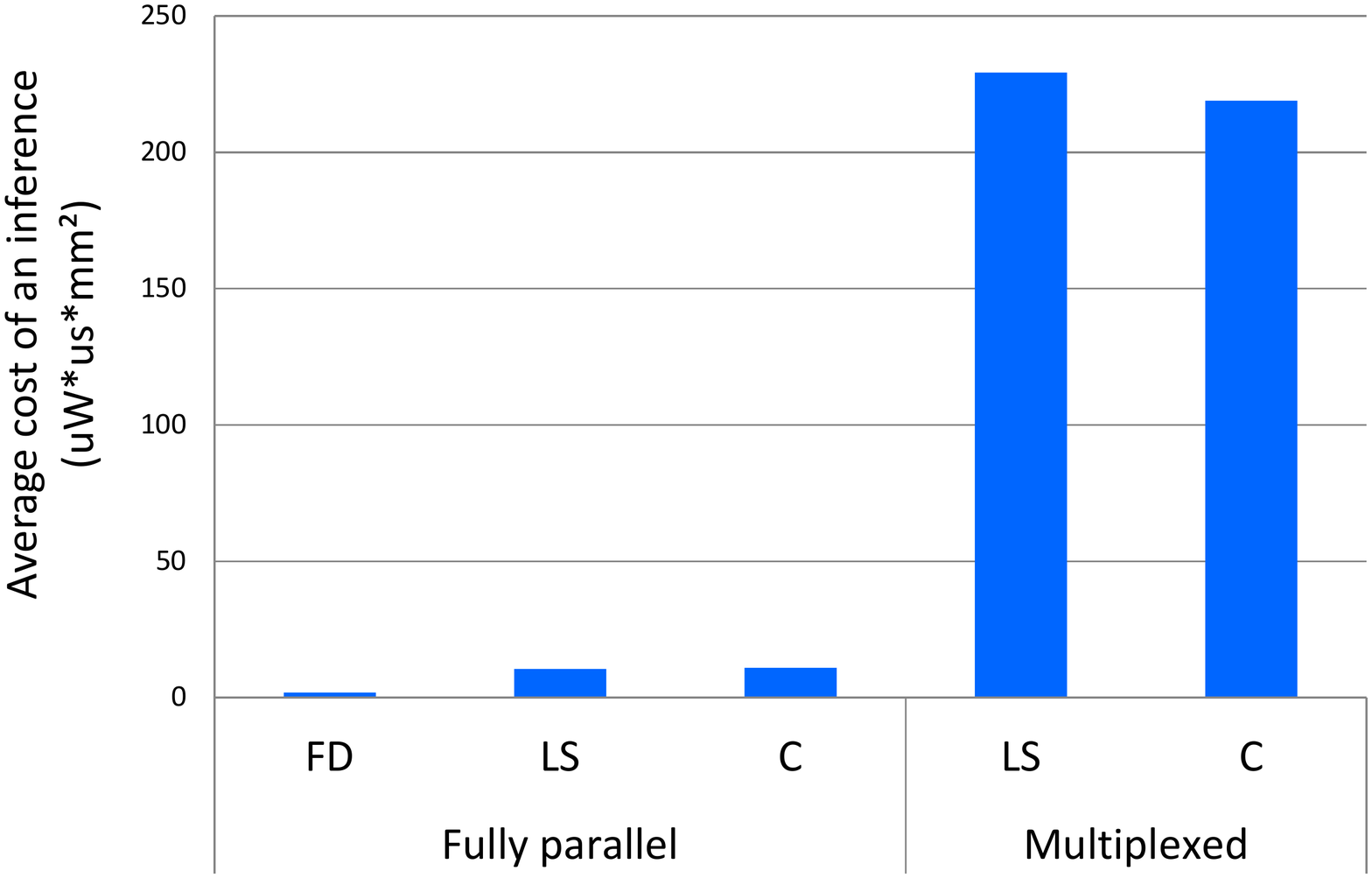}}
	\caption{Qualitative cost function for a 804 neurons hardware SNN for the different architectures available in NAXT}
	\label{Resultatfigure}
\end{figure}
Results are presented in table \ref{Resultattable}. For a better understanding of these results, they are also depicted in figure \ref{Resultatfigure} by virtue of a qualitative cost function. This cost function is calculated as the product of three parameters (latency, energy and chip surface), as we seek to minimize those parameters at the same time. Note that this representation is purely qualitative, but gives a good indication of which architectures are the most suitable for embedded implementation.

%BM2: Puisqu'on donne les surfaces, il faut indiquer quelle est la techno visée. Et puisque la techno peut changée, cela serait donc un troisième rung de configuration. Donc ajoute-t-on une section 4.4 Technology ? Qui précise comment se calculent surfaces et conso ? % Ou plutôt, pourquoi cette section a t elle été enlevée ?
%EL On avait convenu lors de la réunion qu'on faisait sauter toutes les réfèrences aux MRAMS dans cette partie pour se concentrer sur le reste, comme l'aspect "techno mémoire" n'était pas dans le scope du sujet.
%BM2: OK mais indiquer la techno dans le caption
\begin{table}
\begin{center}
\caption{Simulation results for a 784-10-10 SNN hardware for the available architectures in NAXT, with SRAM on-chip memories}
\label{Resultattable}
\begin{tabular}{|c|c|c|c|c|c|}
\hline
Architecture & \multicolumn{3}{c|}{Fully parallel} & \multicolumn{2}{c|}{Multiplexed} \\ \hline
\begin{tabular}[c]{@{}c@{}}Memory \\ organization \footnotemark\\ \end{tabular} & \begin{tabular}[c]{@{}c@{}} FD \end{tabular} & \begin{tabular}[c]{@{}c@{}} LS \end{tabular} & C & \begin{tabular}[c]{@{}c@{}} LS \end{tabular} & C \\ \hline
\begin{tabular}[c]{@{}c@{}}Chip area \\ ($mm^{2}$)\end{tabular} & 13 & 13 & 13 & 1.3 & 1.3 \\ \hline
\begin{tabular}[c]{@{}c@{}}Energy \\ consumption per\\ inference (uJ)\end{tabular} & 3.34 & 3.37 & 3.35 & 27.9 & 27.2 \\ \hline
\begin{tabular}[c]{@{}c@{}}Latency per \\ inference (us)\end{tabular} & 0.042 & 0.24 & 0.25 & 6.32 & 6.19 \\ \hline
\end{tabular}
\end{center}
\end{table}
\footnotetext{LD: Layer Distributed; LS: Layer Shared; C: Centralized}
The obtained results are consistent with our expectations: the trade-off between chip surface on one side, and energy consumption and latency on the other side, is clearly visible in these estimations. These results show that \textit{fully-parallel architectures} globally decrease latency and energy cost at the expense of chip surface, while \textit{time-multiplexed architectures} have the opposite effect. This acknowledgement is quite straightforward, as \textit{TMA} is based on an opposite design paradigm compared to 
%AC - an extension of/derived from parallel architectures?
%BM - sentence has been modified
parallel architectures: they are more compact, but processing serialization results in higher latency, increasing energy consumption (notably because of leakage power). Moreover, we confirmed that the more memory is distributed among processing units, the faster processing will be. Indeed, when memory is centralized, parallel access to stored data is impossible and must be serialized as explained in subsection \ref{Analytics}. This involves a severe increase in latency when memory is centralized. On the other hand, memory architecture does not significantly influence energy consumption and chip surface.

%EDGAR STYLE DEBUT
Therefore, we found that both multiplexed and parallel architectures have their own advantages and drawbacks, that is, the trade-off between processing latency, energy consumption and chip surface (i.e., FPGA occupation). In light of these findings, we will develop three architectures: Fully-Parallel, Time-Multiplexed, and the novel Hybrid Architecture, which uses both paradigms to optimally fit the spiking activity in the network.
%EDGAR STYLE FIN

\begin{figure}
    \begin{center}
        \centerline{\includegraphics[width=0.85\linewidth]{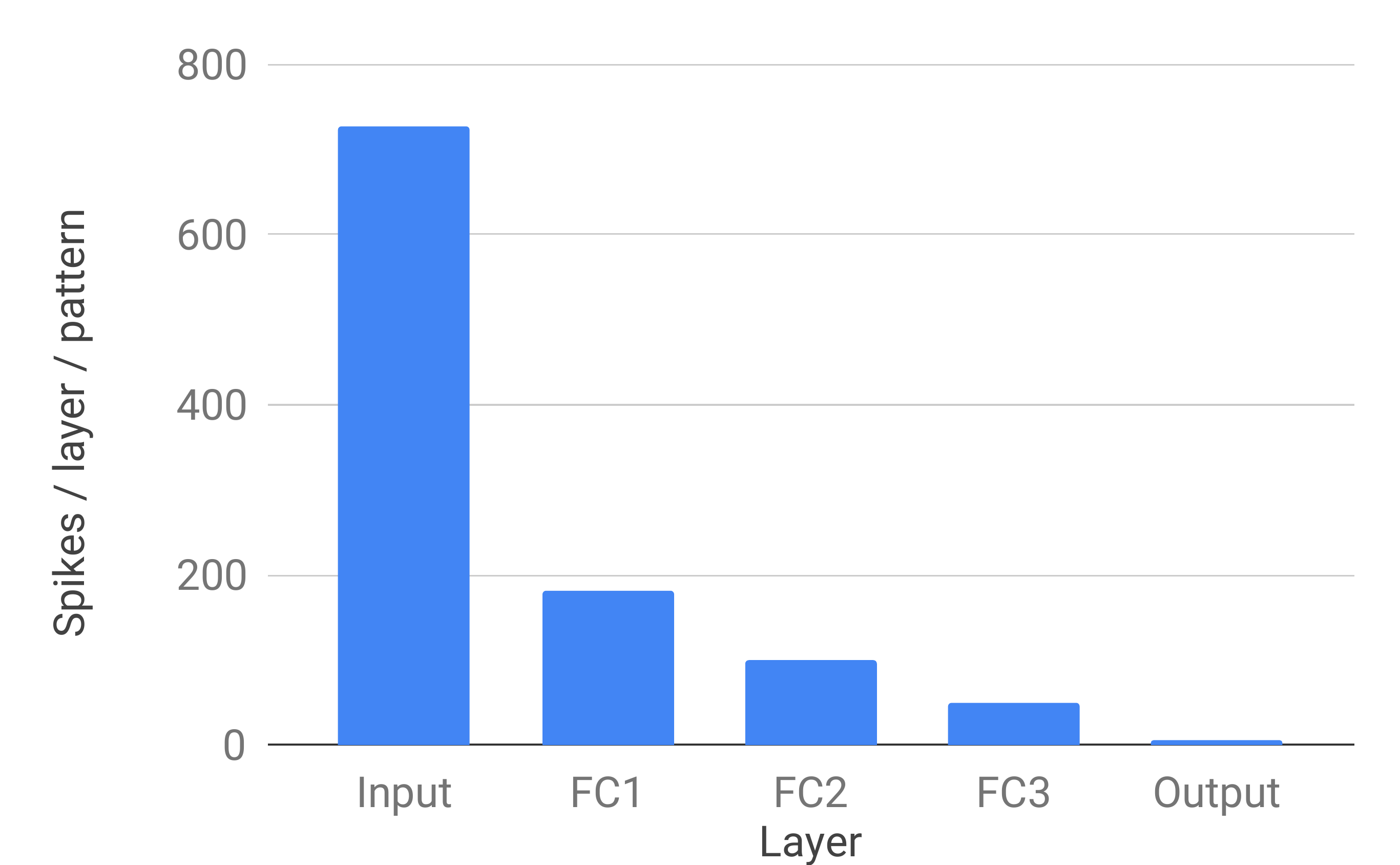}}
        \caption{Average number of spikes generated for one MNIST pattern for each layer in the "784-300-300-300-10" SNN}
        \label{spikes_layer}
    \end{center}
\end{figure}

Indeed, as shown in figure \ref{spikes_layer}, in a feed-forward SNN, the number of input spikes per layer decreases drastically as we go deeper in the network: the first layers are much more solicited than deeper layers during inference. This effect is even more prominent when using our novel Spike Select information coding method (see section \ref{information_coding}). Consequently, we assume that the first layers must be implemented in a fully-parallel fashion to prevent spike bottlenecks, whereas deeper layers can be implemented in a multiplexed fashion. From this assumption, a hybrid architecture has been developed in VHDL and simulated at the Register Transfer Level (RTL), which provides finer estimations than the NAXT simulator. This architecture will be presented in section \ref{DHSNNA}, alongside with \textit{Fully-Parallel} and \textit{Time-Multiplexed architectures}, which have also been developed and simulated at the RTL level.
 
\section{SNN hardware architecture design} \label{DHSNNA}

In this section, we describe the hardware design implementation of the SNN architectures studied in section \ref{exploration}. Indeed, among the different models we implement: \textit{Fully-Parallel Architecture} (\textit{FPA}), \textit{Time-Multiplexed Architecture} (\textit{TMA}), and \textit{Hybrid Architecture} (\textit{HA}) which is the major contribution of the present work. We have selected those three different architectural paradigms according to NAXT simulation results, which enlightens how those three architectures are well suited to evaluate the trade-off between resource intensiveness, power consumption and latency.  To do so, we first present the different modules used to build the different designs, then we describe the complete systems. As mentioned in section \ref{n2d2}, we use N2D2 to extract the different parameters of SNNs to move to the hardware implementation of the neuromorphic architectures. This phase is realized with the Intel\textregistered{} Quartus\textregistered{} Prime 18.1.0 Lite edition for FPGA prototyping, and ModelSim\textregistered{} for the validation with simulation of the design behavior. 

\subsection{Hardware modules}

\subsubsection{Integrate-and-Fire neuron module}
\label{neuron_module}

\begin{figure} 
	\centerline{\includegraphics[width=0.9\linewidth]{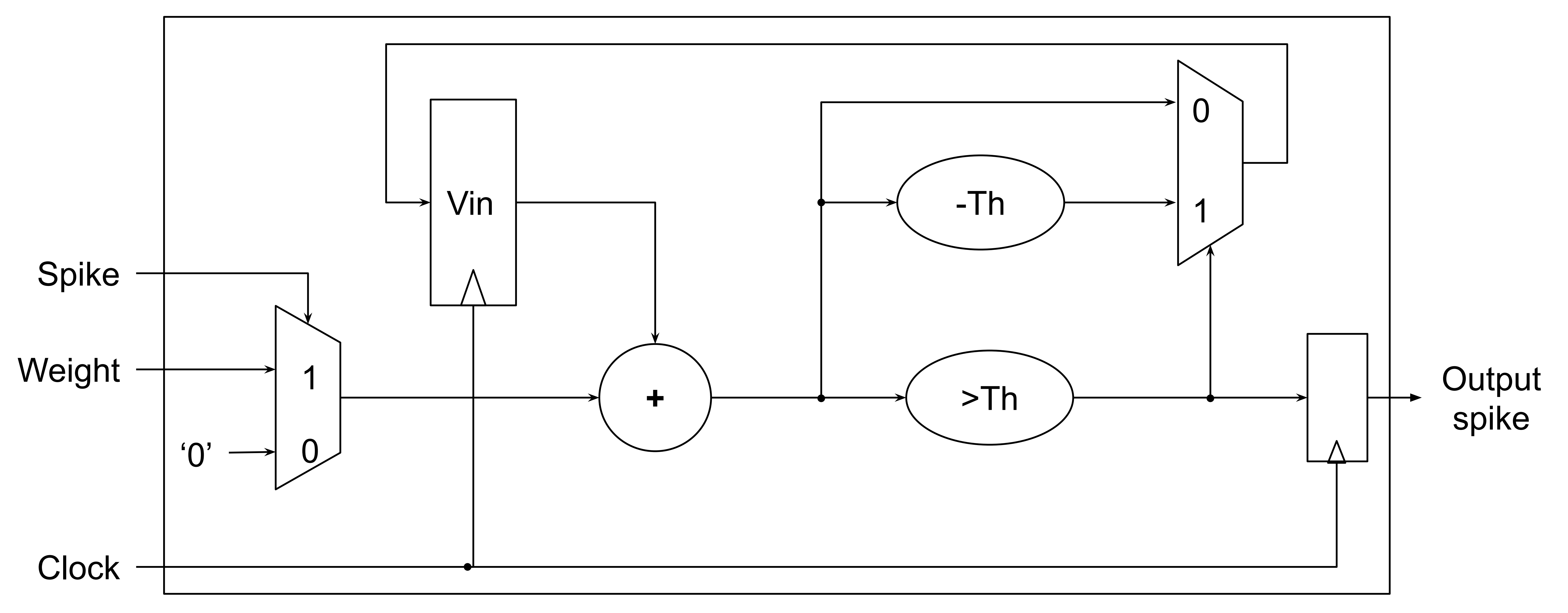}}
	\caption{IF neuron module internal structure}
	\label{hard_if}
\end{figure}

%EDGAR STYLE DEBUT
The \textit{IF-neuron} hardware structure is illustrated with the simplified schematic diagram in figure \ref{hard_if}. In contrast to the perceptron, it does not have a multiplier and thus results in cheaper hardware with only elementary components. The module has two inputs: the input spike and its corresponding synaptic weight;  and one output for output events. For clarity purpose, only positive spikes are considered.

When the neuron receives a spike, it accumulates the corresponding weight with the previous internal potential stored in a register. Afterwards, it compares this accumulated potential with the membrane potential threshold and fires whenever it is exceeded. In the case of a firing, the internal potential is decreased by the threshold amount, otherwise it remains as it is.
%EDGAR STYLE FIN

\subsubsection{Counter module}

The \textit{counter modules} are used for synchronization between layers and neurons. On one hand, they order the beginning and ending of computations for neuron modules and indicate the synapse addresses corresponding to input spikes (in Time-Multiplexed Architectures). On the other hand, they are linked to each other in series to ensure the coherent flow of spikes in the network and thus synchronize the different layers, referring to its usage with FPA (figure \ref{FPA_fig}).

\subsubsection{SNN class selection module}
\label{tdelta}
%EDGAR STYLE DEBUT
Before starting the description, let us give a quick reminder concerning class selection procedures. First of all, note that each output neuron corresponds to a data class. During inference, the winning class is selected as the most spiking output neuron. In Terminate Delta procedure, the class prediction is enacted when the most spiking neuron has spiked \textit{delta} times more than the second most spiking neuron. On other hand, in Max Terminate, the classification process is completed whenever an output neuron (the most spiking neuron) reaches \textit{max-value} spikes. \textit{Delta-value} and \textit{max-value} are user-defined parameters, usually set at 4.% NASSIM DESCRIPTION MAX TERMINATE !!

\begin{figure}
	\centering{\includegraphics[width=0.9\linewidth]{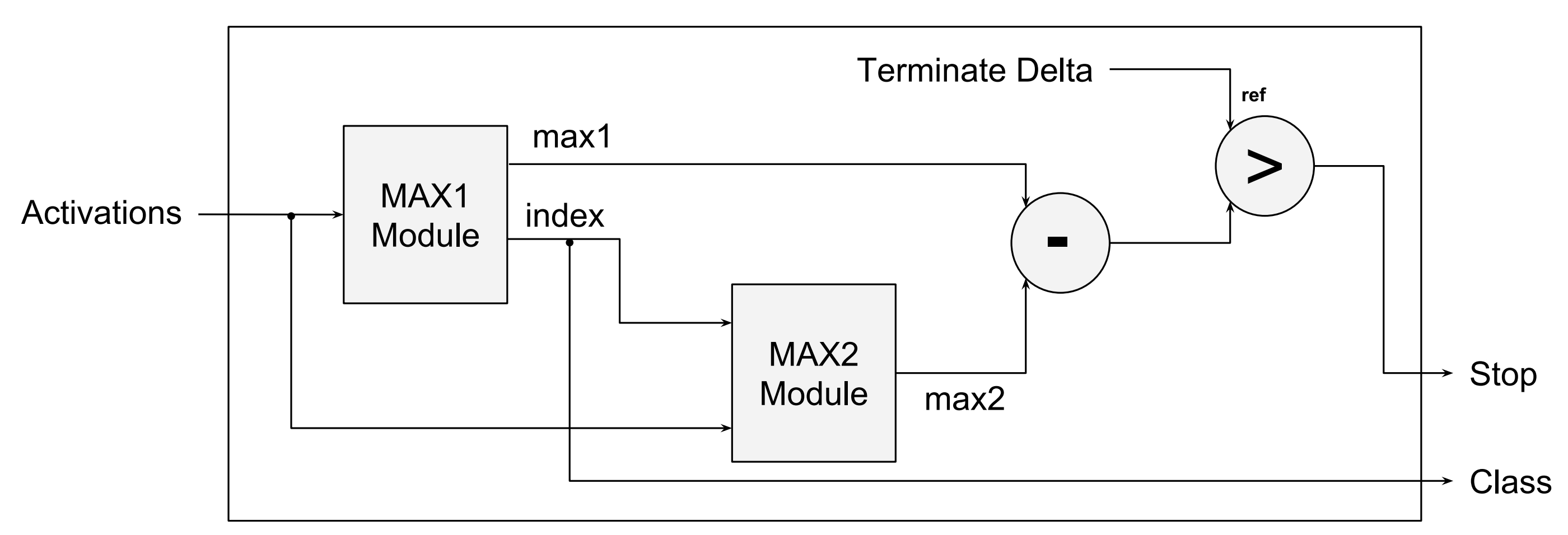}}
	\caption{Schematic diagram of the Terminate Delta module}
	\label{terminate_delta_block}
\end{figure}

\begin{figure}
	\centering{\includegraphics[width=0.8\linewidth]{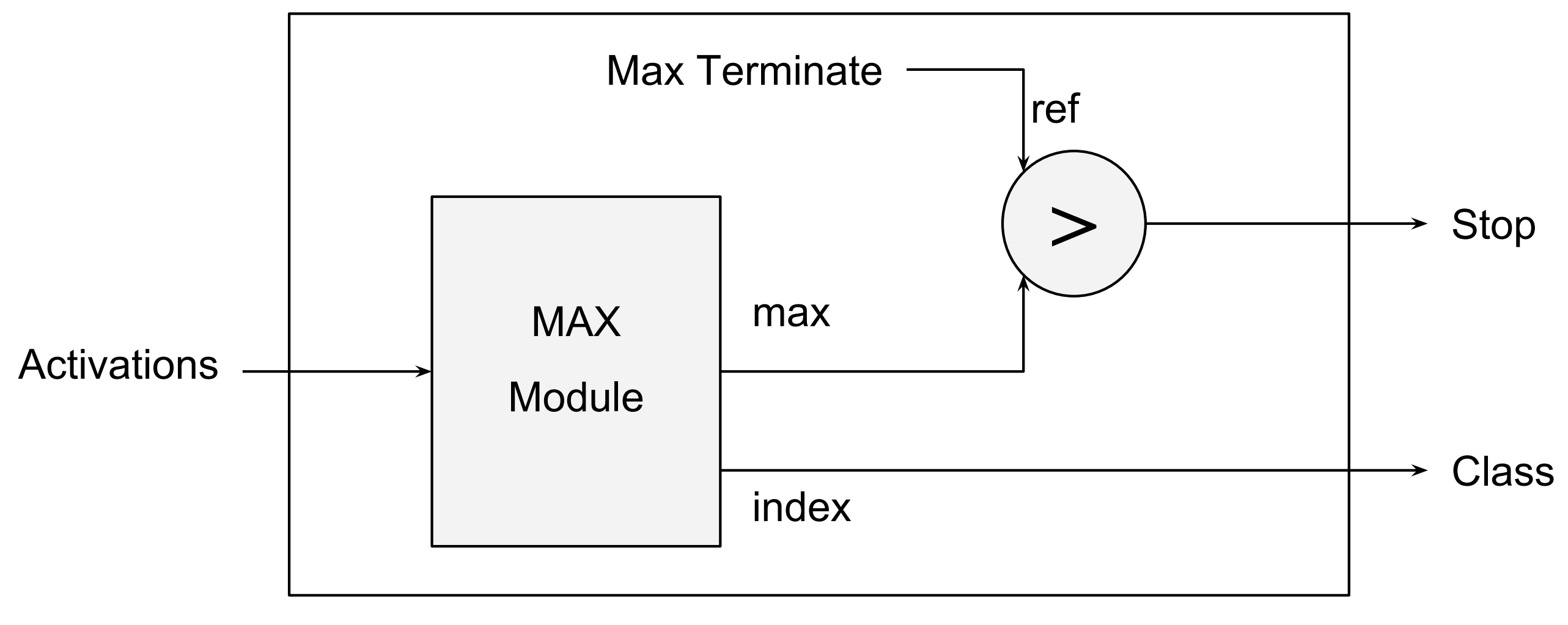}}
	\caption{Schematic diagram of the Max Terminate module}
	\label{terminate_max_block}
\end{figure}

For the design of our architectures, to select the output winner class we chose either \textit{Terminate Delta} or \textit{Max Terminate}, for which the initial software versions are defined in N2D2 framework \ref{n2d2}. We have chosen those methods because they offer State-of-the-art accuracy and fast class selection. The figures \ref{terminate_delta_block} and \ref{terminate_max_block} show the internal structures of these modules. The input of the module is a vector (\textit{Activations}) containing the output activity of the SNN (number of spikes emitted by each output neuron so far). 

On one hand, in the \textit{Terminate Delta} module two {\it maximum sub-modules} are designed to detect the maximum value of an array, which are then used to determine the winning class and to terminate the processing. The first {\it maximum sub-module}, namely Max1, detects the maximum value of the output activation vector, and the second, namely Max2, detects the second maximum value of this same vector. The difference between the outputs of \textit{Max1 module} and \textit{Max2 module} is then computed. Finally, if the difference is greater than a threshold (\textit{delta-value}), the class corresponding to \textit{Max1 Module} is enacted as the winner.

On the other hand, the \textit{Max Terminate} module integrates only one {\it maximum} block that returns the index of the output neuron with the highest spiking activity and its activity. Then this activity is compared to a user-defined threshold \textit{max-value}. If the maximum spiking activity is greater than \textit{max-value}, the corresponding output neuron is enacted as the winner class, and the processing is stopped.
%EDGAR STYLE FIN


\subsubsection{Memory modules}

\subsubsection*{First-in First-out (FiFo) module}

The \textit{FiFo modules} are used in the \textit{Neural Core (NC)}, \textit{Network Controller} and \textit{NPU} modules that are described later. They serve as buffers, where the output spikes of neurons are interpreted as events and are stored in a sorted way, i.e., in an ascending order according to their times of arrival. Figure \ref{fifo} illustrates the schematic block of the designed module, showing its \textit{I/O ports}. Indeed, the input and output data correspond to the \textit{neuron address} (origin of the received spike). They are stored in this format to facilitate the search of related weights in the next layer, due to the huge number of weights. The other signals are for \textit{read/write enable}, \textit{clock/reset} and \textit{FiFo memory empty/full}.

\subsubsection*{ROM module}

%BM2: Les RAM contiennent juste les poids. Ce sont donc des ROM ?!
%NA: Bien vu!! j'allais le modifier hier
%OK!

\begin{figure}
      \begin{subfigure}{.5\linewidth}
        \centerline{\includegraphics[width=\linewidth]{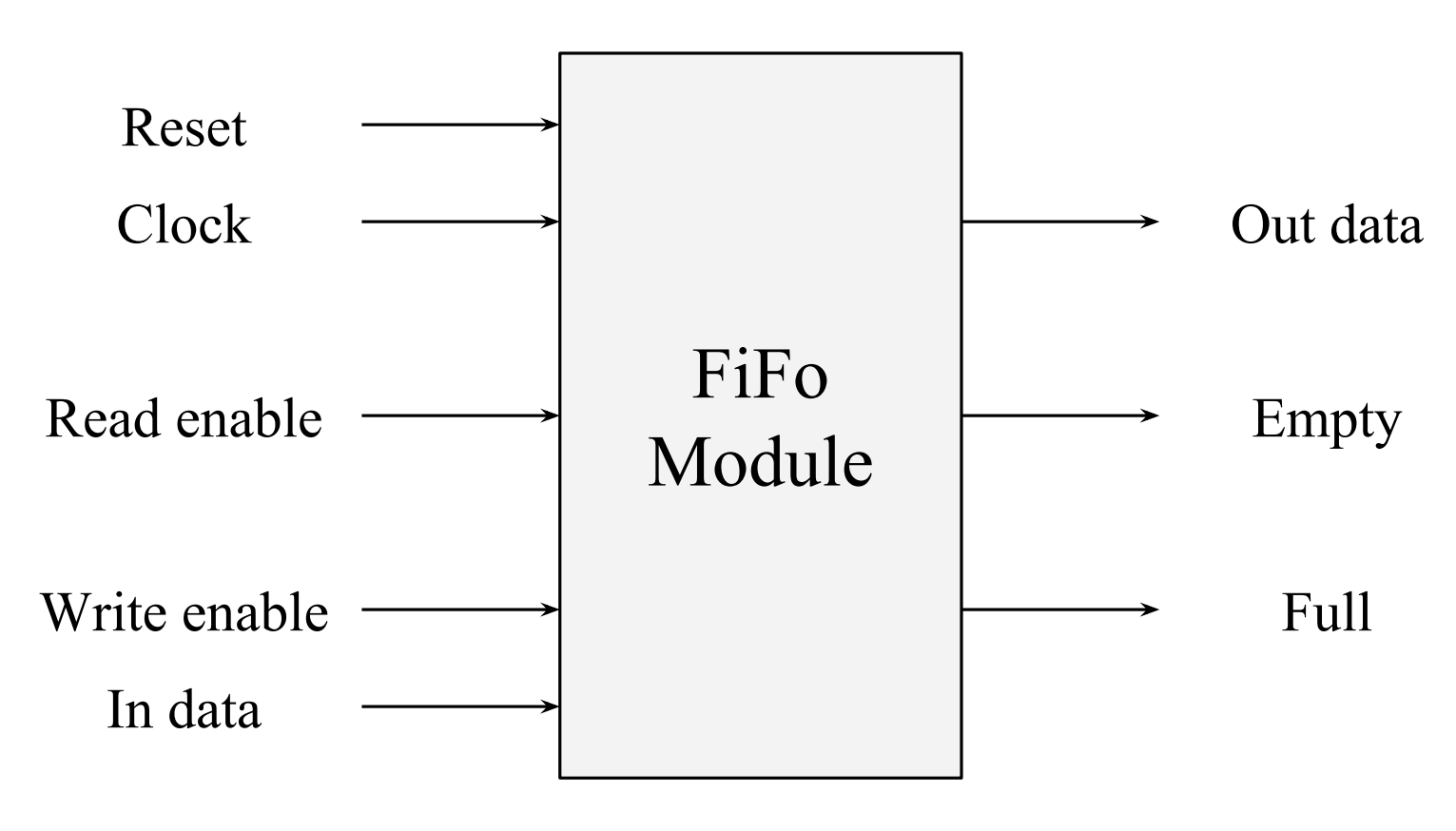}}
            \caption{FiFo I/O ports}
            \label{fifo}
    \end{subfigure}%
    \begin{subfigure}{.5\linewidth}
        \centerline{\includegraphics[width=\linewidth]{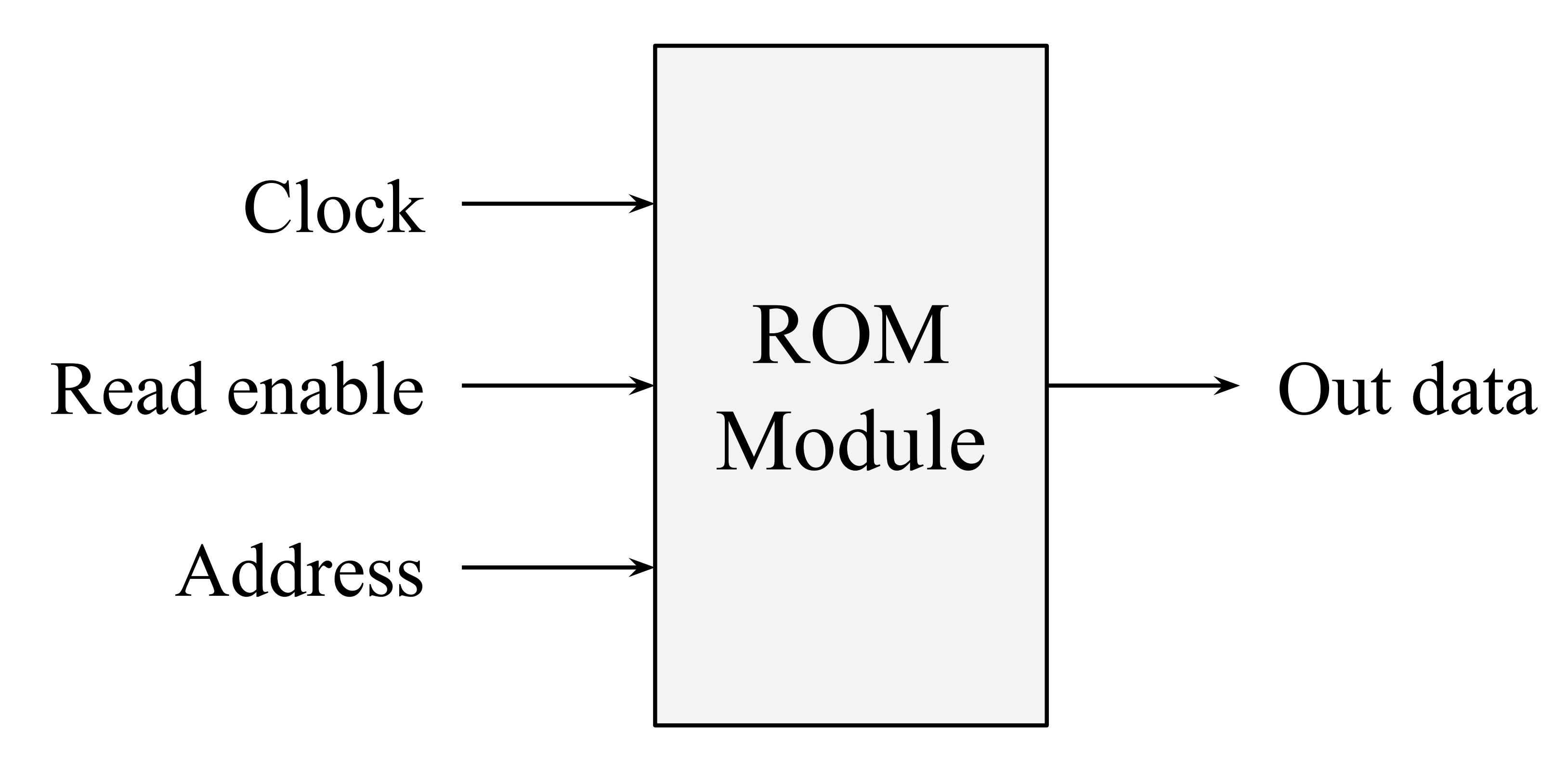}}
            \caption{Memory I/O ports}
            \label{memo}
    \end{subfigure}
    \caption{FiFo and memory I/O port blocks}
    \label{fifo-memo}
\end{figure}

%EDGAR STYLE DEBUT
\textit{Memory blocks} are required for the proper operation of the neuromorphic system. In FPGA technology, they can be of different types: RAM, ROM, registers or latches. The \textit{ROM modules} are used in the design of \textit{TMA} (Time-Multiplexed Architecture), and they store the weights of the \textit{NPU}'s logical neurons in an SNN layer. Therefore, the \textit{ROM}s are of different sizes depending on the number of emulated synaptic connections. The \textit{I/O ports} of a \textit{ROM block} are shown in fig. \ref{memo}.
%EDGAR STYLE FIN

\subsubsection*{SDRAM}

Memory usage is the common limitation for SNN architectures, as mentioned in section \ref{mem_capacity_sec}, which is due to all the parameters and activities of the neurons that must be stored. From that perspective, in order to deal with deeper networks that require a significant memory size, the FPGA on-chip memory will not be sufficient. Therefore, external memory must be used to overcome this problem. 

In this paper, we use \textit{SDRAM} to reinforce the memory capabilities of the FPGA fabric. To do so, we designed a \textit{Network Controller module} that connects the other modules to this external memory. 
%BM2: le Neural Core n'apparait pas dans les figures à part la 20. Bizarre ?!
%NA: Yes!! tu as raison, je l'ajoute de suite 
%OK!

\subsubsection{Neural Core module}
\label{NCModule}
%BM2: ref{ha} n'est pas la bonne ref de section
%NA:OK!

%EDGAR STYLE DEBUT
 The \textit{Neural Core module} is the computation unit which emulates the two first layers (input and first hidden) of the Hybrid Architecture (HA) presented in section \ref{ha_section}. This module includes an \textit{Input Neuron Module} which forwards input spikes to downstream neurons; \textit{IF Neuron Modules} which integrates incoming events from the \textit{Input Neuron Module} and generate spikes according to Integrate and Fire rule. The weights are stored in registers, so that each \textit{IF Neuron module} has its weights in a dedicated register. There are as many \textit{IF Neuron Module} as logical neurons in the layer. Their outputs are stored in a \textit{FiFo buffer} as events, with a \textit{Counter Module} indicating the corresponding neuron address to be stored.
 
 %A mon avis, trop lourd et technique, ce n'est pas nécessaire à la compréhension.. :
 
 %To do so, the {\it write enable} signal of the \textit{FiFo module} is connected simultaneously to the output spike signal of each IF-neuron, and stores the counter output when enabled (output spike = 1), refer to figure \ref{neural_core}.

%EDGAR STYLE FIN

\begin{figure}
    \centering
    \includegraphics[width=\linewidth]{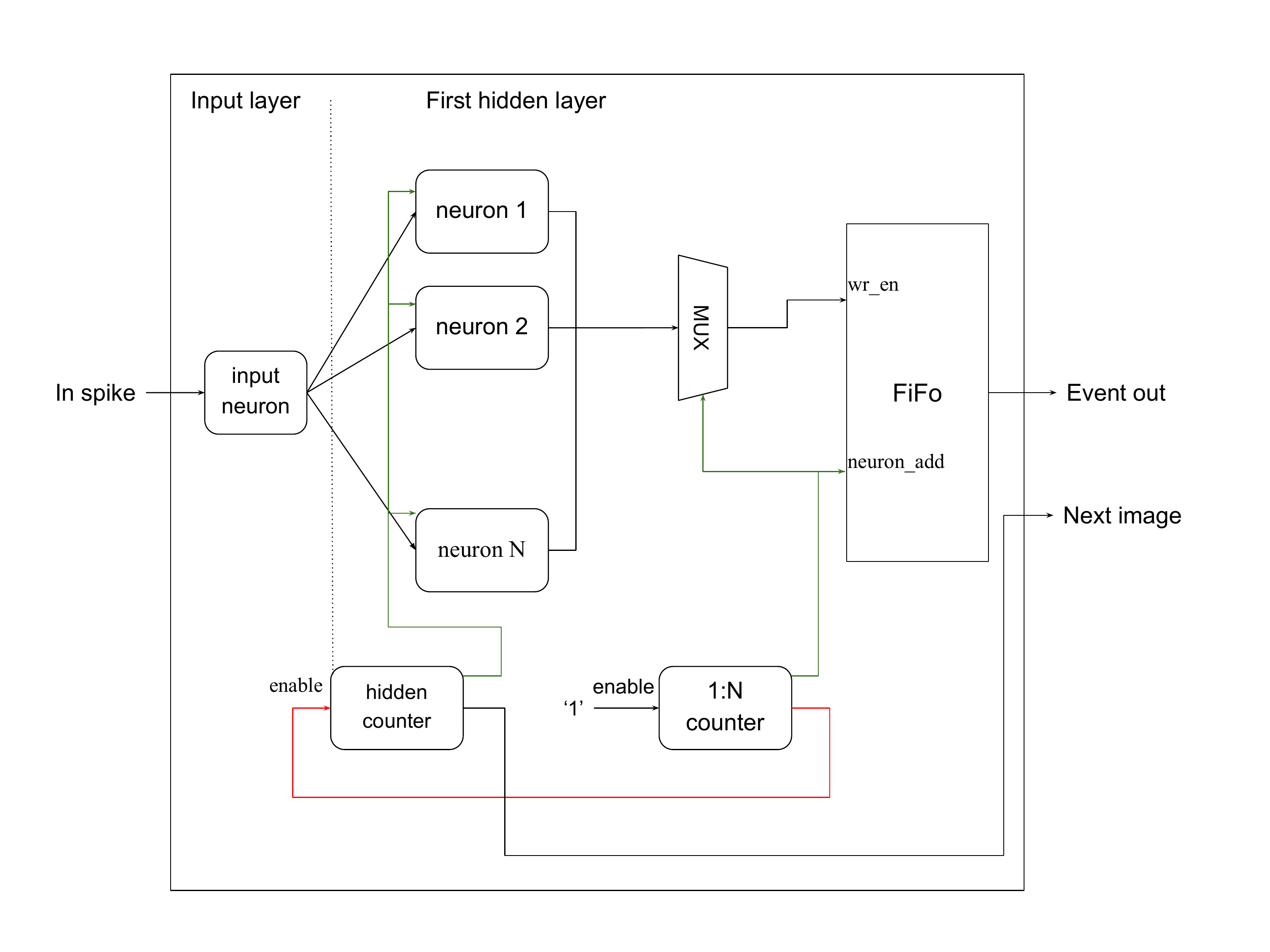}
    \caption{Neural Core simplified schematic diagram. The input neuron forwards input spikes, spike by spike, to the $1^{st}$ hidden layer neurons. The $1^{st}$ hidden counter is indicating to that hidden neurons the address of the input spike in order to retrieve their appropriate weights. Another counter (1:N) is controlling a MUX component to store the $1^{st}$ hidden layer's spikes in a FiFo memory.}
    \label{neural_core}
\end{figure}

\subsubsection{Neural Processing Unit module}
\label{npumodule}
\begin{figure}
	\centering
	\includegraphics[width=0.9\linewidth]{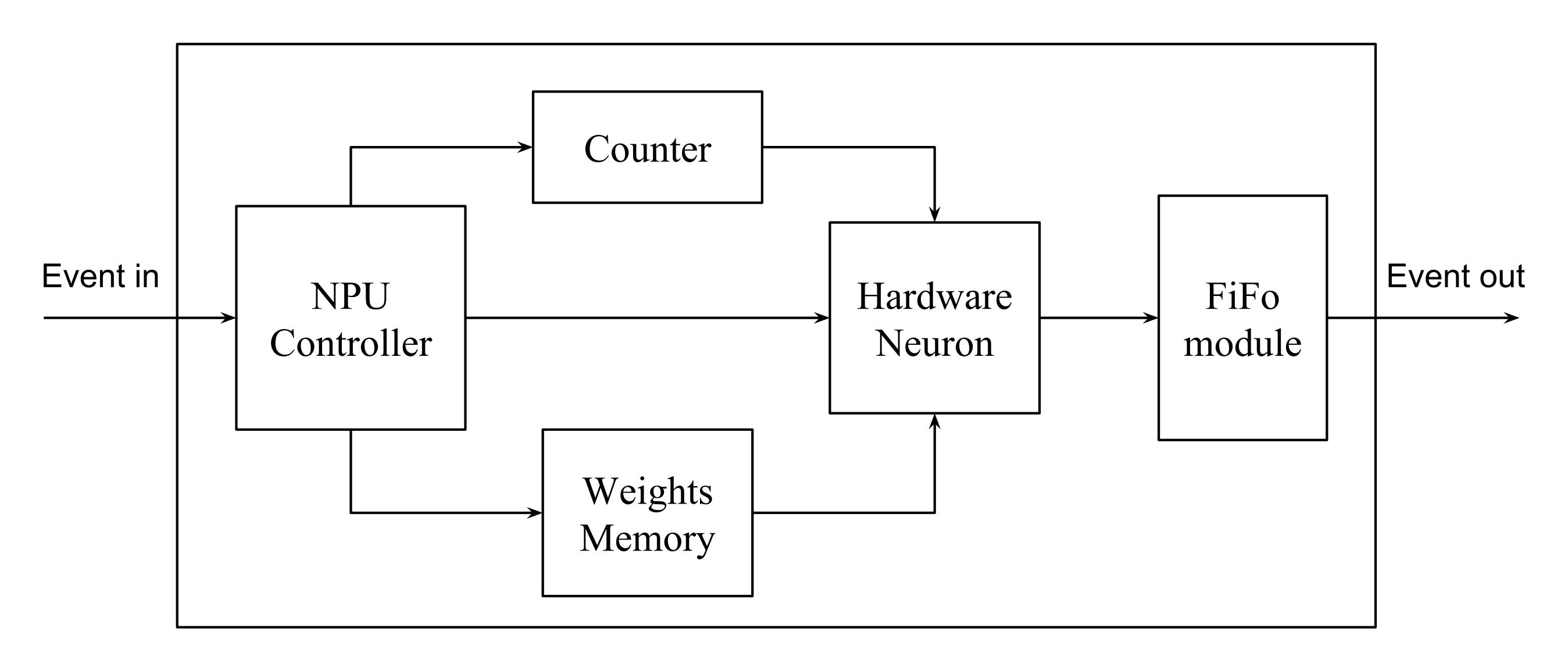}
	\caption{Neural Processing Unit simplified block diagram. When there is an input event to process by the NPU: first, the hardware neuron is enabled by the NPU controller to retrieve the address of the logical neuron it represents from the counter and the corresponding weights from the memory block. Second, do its computation, and whenever it fires, the output spikes is stored in a FiFo as an event.}
	\label{NPU}
\end{figure}

%EDGAR STYLE DEBUT
The \textit{Neural Processing Unit Module} (NPU) is used to emulate time-multiplexed layers. A single \textit{IF Neuron module} will operate successively for all neurons in the layer. Moreover, the NPU includes a \textit{FiFo Memory module},a  \textit{Counter module} and an \textit{NPU controller}. These modules are connected as shown in figure \ref{NPU} to form a \textit{NPU} which processes spiking events in a coherent way. However, besides \textit{NPU controller}, all the other modules were presented before, and they are used by the NPU to accomplish their dedicated tasks. Consequently, only NPU controller will be described in details.
%Whereas, the controller's role is to drive these modules to get the IF-neuron processes an input event.
%AC - this is confusing, perhaps "...to get the IF-neuron to process an input event"?
%BM- OK sentence has been changed
The goal of the \textit{NPU controller} is to manage the different \textit{NPU modules} to trigger logical neurons in a coherent way, allowing the hardware neuron to be fed with valid weights and activities.

%Le role du controleur est de coordonner les differents modules du NPU pour que le neurone physique traite les neurones logiques de manière cohérente avec les évenements d'entrée, et les parametres de poids et d'activité associés.
In addition, \textit{NPU controllers} of different NPUs are connected together in order to ensure synchronization at the network level. This synchronisation is required as output classification process (Terminate Delta) depends on the arriving order of the spikes. Each \textit{NPU module} can represent several logical neurons thanks to time-multiplexing. Note that the used weights memory type is ROM with \textit{TMA}, and SDRAM with \textit{HA}.

%EDGAR STYLE FIN

\subsubsection{Network controller}
\label{netcontroller}

The \textit{network controller module}, used in the HA architecture, is a combination of a \textit{FiFo module} and a \textit{demultiplexer} (\textit{DEMUX}); which is organized as shown in figure \ref{ha}. The \textit{FiFo module} accesses the SDRAM according to the \textit{NPU} requests with a first-come-first-served policy, i.e., when an \textit{NPU} requests a weight, this request is put in the \textit{FiFo} queue. Then, whenever the weight is ready, it is sent via the \textit{DEMUX block} by selecting the right \textit{NPU}.

\subsection{Fully-Parallel Architecture}
\label{FPA}
\begin{figure}
	\centering{\includegraphics[width=\linewidth]{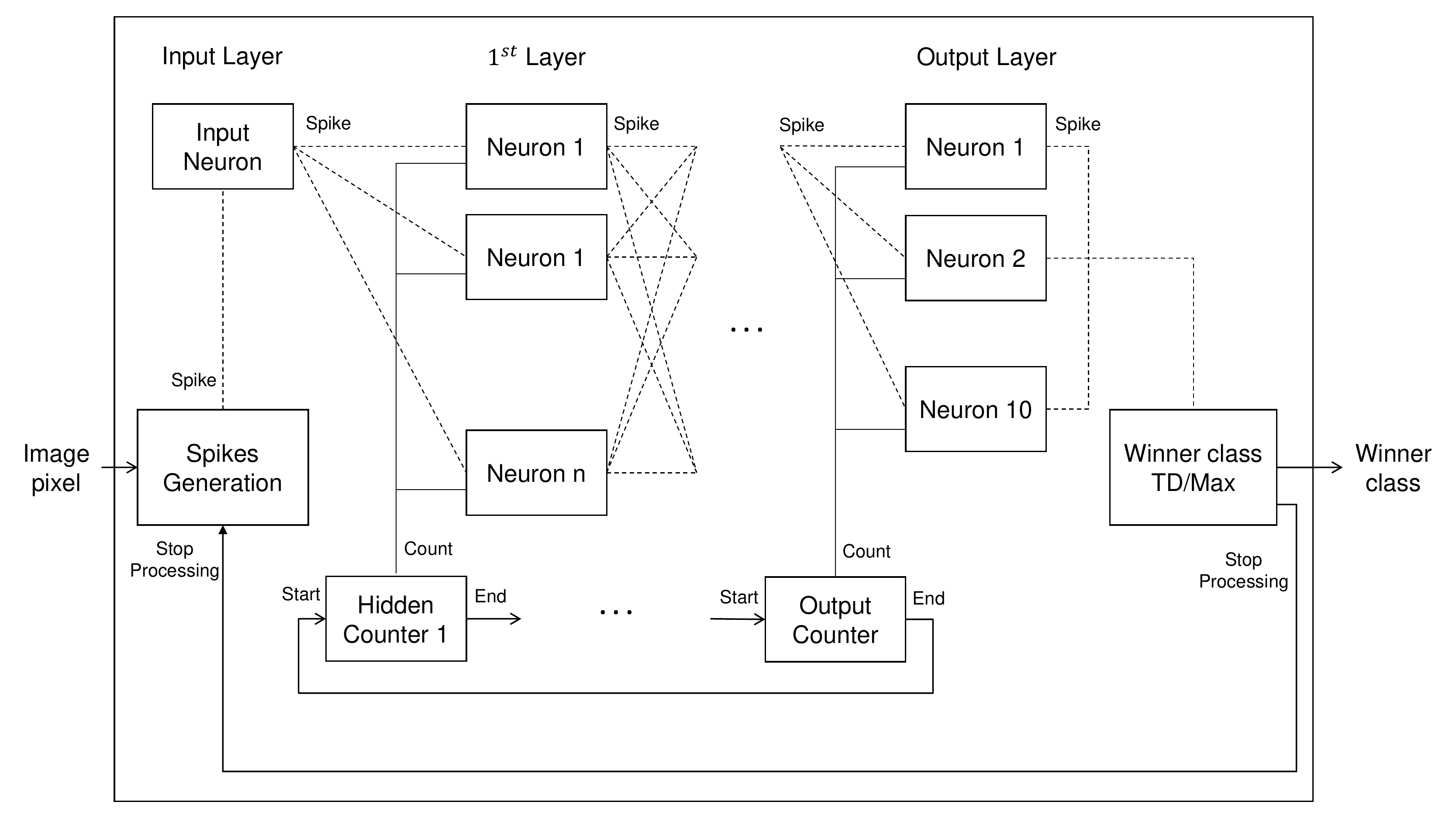}}
	\caption{FPA simplified schematic diagram. In this architecture, from the $1^{st}$ layer to the output layer all neurons are implemented in hardware. Counters connected in series are coordinating the processing of the different network spikes in a coherent way.}
	\label{FPA_fig}
\end{figure}

%EDGAR STYLE DEBUT
This subsection describes the Fully-Parallel Architecture (FPA) we have developped for present work. This architecture has been conceived alongside Time-Multiplexed Architecture (TMA) to evaluate the trade-off between latency and resource intensiveness at a much finer level than NAXT Simulator.
In the \textit{Fully-Parallel Architecture},  all the logical neurons of the SNN are implemented in hardware. In other words, the \textit{IF-neuron module} is instantiated as many times as the number of logical neurons. Figure \ref{FPA_fig} shows the connectivity of the different components of the architecture. There is one \textit{Counter module} for each layer, used to synchronize the neuron computations in the network. Indeed, in this architecture, each layer waits for the previous one to finish all its processing before starting : all spikes are processed layer by layer. The \textit{Input Neuron} module forwards the data, spike by spike, to the first hidden layer where a {\it Hidden Counter} is counting them. At each clock cycle, the hidden layer \textit{IF Neuron modules} integrate the incoming spike and store their consequent output spikes in a buffer. Then, when all the input spikes are processed, the {\it Hidden Counter} sends an {\it End Signal} to the next layer \textit{Counter}. All the hidden layers accomplish the same process on their own incoming spikes, layer after layer. The last hidden counter enacts the end of the process to the output layer {\it Counter}. At this level, it is up to this counter (output layer counter) to trigger the output neurons to process the last hidden layer output spikes. Finally, the outgoing spikes are processed by the {\it Winner Class Selection module}, which decides whether to end the computations or to repeat the process.
%EDGAR STYLE FIN
This fully-parallel architectural choice should result in fast processing but high logic resource intensiveness. In the following subsection, we present the second developed architecture which takes the opposite architectural choice : Time-Multiplexed Architecture (TMA).

\subsection{Time-Multiplexed Architecture}
\begin{figure}
    \centerline{\includegraphics[width=\linewidth]{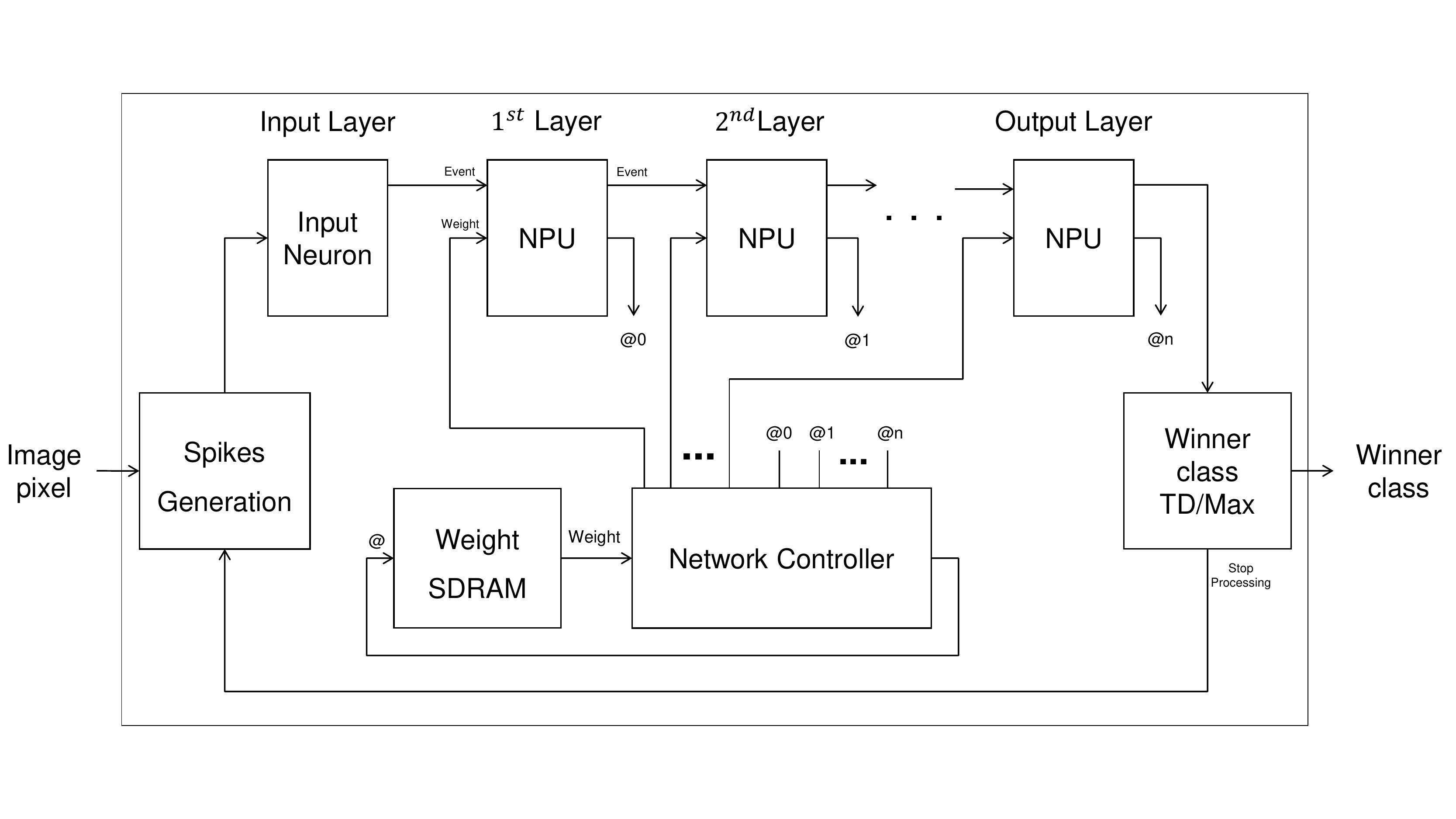}}
    \caption{TMA simplified schematic diagram. Here, the input neuron module forwards input spikes to NPUs connected in series, each NPU represents a distinct layer, that execute the SNN neurons to finally using the winner class selection module output the SNN's class.}
    \label{tma_fig}
\end{figure}

%EDGAR STYLE DEBUT
The {\it TMA} architecture is designed to save hardware resources, in contrast with \textit{FPA} architecture. In this implementation, the main computation unit is the \textit{NPU module} described in section \ref{npumodule}. Contrary to \textit{FPA}, the number of hardware neurons is smaller than the number of logical neurons: each layer is represented by one single \textit{NPU}, instead of one \textit{NPU} per neuron. The complete hardware architecture consists of \textit{NPU modules}, interconnected with each other as shown in figure \ref{tma_fig}. As in \textit{FPA}, the input layer is represented by a dedicated {\it Input Neuron module}, which forwards input spikes to the first hidden layer. Each one of the other layers are represented by one single \textit{NPU}, which successively compute the layer's logical neurons in a time-multiplexed manner. These \textit{NPUs} have their own \textit{ROM memory} containing their parameters. This architecture should drastically diminish the hardware occupation, but increase the system latency as a counterpart. In other words, \textit{TMA} and \textit{FPA} represent the two extremes of the latency versus hardware intensiveness trade-off. In the next subsection, we will describe a middle ground between those two extremes, taking advantages from both to fit the reality of spiking activity in an SNN : the novel Hybrid Architecture (HA).
%EDGAR STYLE FIN

\subsection{Hybrid Architecture}
\label{ha_section}

\begin{figure}
    \centering
    \centerline{\includegraphics[width=\linewidth]{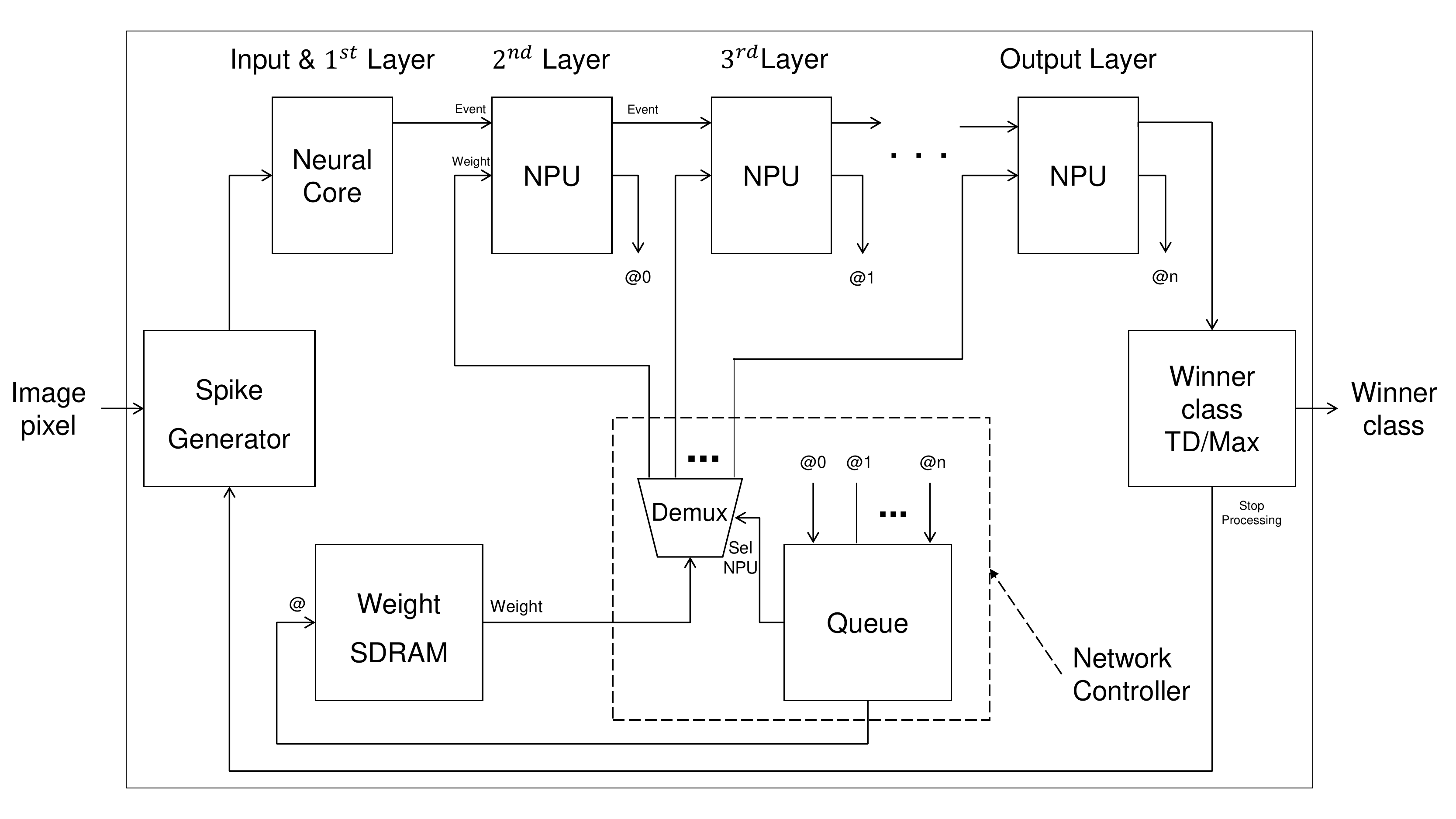}}
    \caption{Hybrid architecture simplified schematic diagram. The input and the $1^{st}$ layers are implemented using a Neural Core and NPUs are used for the remaining layers (one NPU per layer). The output layer spikes are fed to the winner class selection module for classification. A Network Controller is used to manage and connect the NPUs to an SDRAM holding their logical weights.}
    \label{ha}
\end{figure}

%EDGAR STYLE DEBUT
In section \ref{exploration}, it was mentioned that most of the spiking activity in the network is located in the first layer. Therefore, the first hidden layer is the most solicited layer during processing. To take advantage of this aspect, the Hybrid Architecture (\textit{HA}) is designed, mixing both \textit{TMA} and \textit{FPA}. Moreover, this novel hybrid architecture is appropriated for the use of the novel Spike Select method described in \ref{information_coding}, in which spiking activity is concentrated in the first layer. This implementation is the main novelty of the present work, and it derives from the findings and observations we made thanks to our funnel-like Design Space Exploration framework. It is a mixture of \textit{FPA} and \textit{TMA}, where: first, the initial two layers are implemented using a \textit{Neural Core} module as in \textit{FPA}; second, the remaining layers are time-multiplexed using one \textit{NPU} per layer, as in \textit{TMA} . The time-multiplexed part is driven by a network controller, to retrieve the weights from the external SDRAM memory and forward them to the corresponding \textit{NPUs}. The complete hardware schematic diagram is illustrated in figure \ref{ha}, showing its modules and their connectivity.
%EDGR STYLE FIN

% First, the network controller uses a FiFo module for storing and sorting the weight requests of the different NPUs with their identifiers\footnote{NPU identifier: the layer number it represents}. Second, it uses these identifiers to select the NPU for which a weight is retrieved from the SDRAM.
%BM2: Je pense qu'il faut trouver un autre terme que identities : address, ID... et remplacer dans tout le document. Identities = identité et non identifiant
%NA: maintenant, on utilise identifier pour NPU et address pour neuron
%OK!

\subsection{Results: hardware resources occupation}

In the design of AI-embedded architectures, it is important to consider resources occupation due to the lack of silicon area. Therefore, we quantify and compare the hardware cost estimations of the architectures presented in section \ref{DHSNNA}. They are described by three generic VHDL codes, which are compatible with any fully-connected multi-layer SNN topology. These VHDL codes use parameters extracted from N2D2. Their hardware costs, latency and computation performance on the \textit{"5CGXFC7C7F23C8" Cyclone\textregistered V FPGA} board were measured through a synthesis in {\it {Intel\textregistered} {Quartus\textregistered} Prime Lite 18.10 edition}. Therefore, several SNN topologies of different size are implemented with the three hardware architectures.

\begin{figure}
\centering
\includegraphics[width=\linewidth]{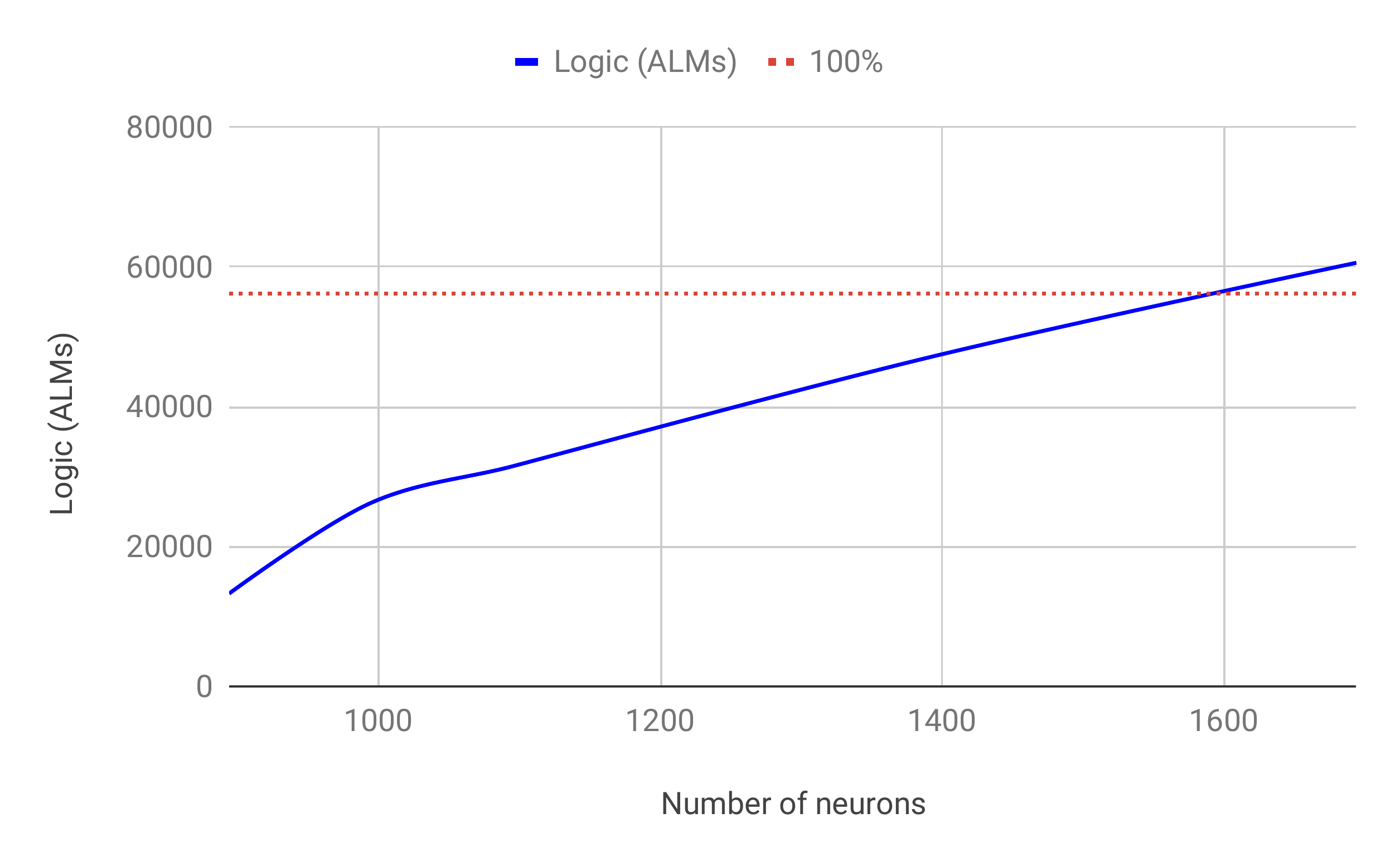}
\caption{FPA architecture: FPGA logic (ALM) utilization versus the SNN number of neurons; Different SNN topologies are used, see table \ref{fpa_table}.}
\label{fpa_logic}
\end{figure}
\begin{figure}
\centering
\includegraphics[width=\linewidth]{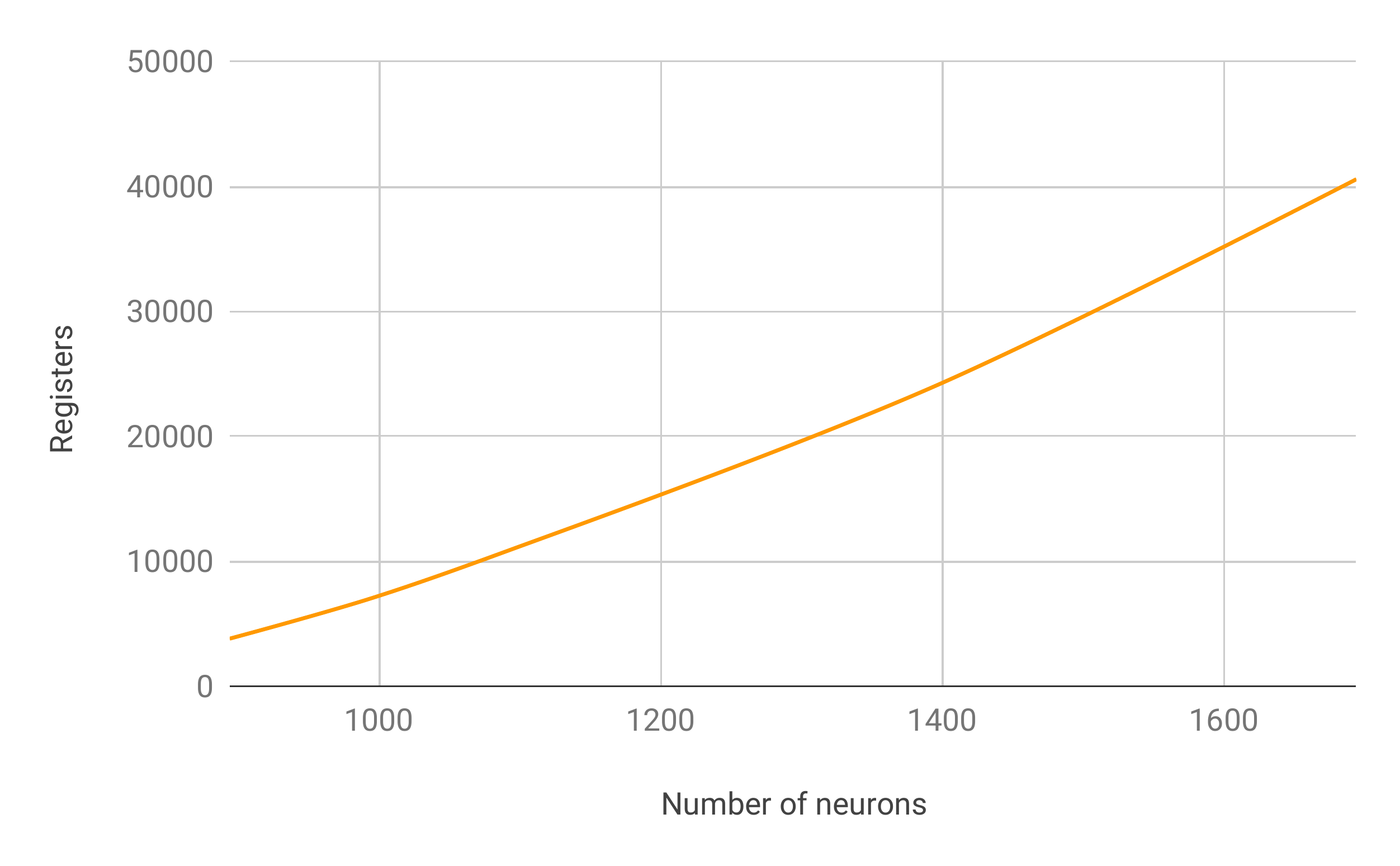}
\caption{FPA architecture: FPGA registers occupation versus the SNN number of neurons; Different SNN topologies are used, see table \ref{fpa_table}.}
\label{fpa_registers}
\end{figure}

The synthesis results using \textit{FPA} are summarized in table \ref{fpa_table}, giving the logic (ALM) and registers occupation related to the number of neurons. Then, those results are plotted in two graphs showing the evolution of resource intensiveness against amount of neurons (figures \ref{fpa_logic} and \ref{fpa_registers}). From these results, we observe that the \textit{FPA} logic occupation is directly proportional to the SNN depth/size, i.e., increases linearly with the amount of neurons. Nevertheless, the generated circuits are supported by the FPGA fabric when the networks are smaller than the {\it 784-300-300-10} topology, but not for bigger ones. Therefore, our first intuition regarding the limited scalability of \textit{FPA} when used for deep SNNs is confirmed. But, we are yet to confirm if the \textit{TMA} or \textit{HA} architectures occupy less resources, and are thus more viable.

\begin{table}
\begin{center}
\caption{FPGA occupation (Logic ALMs and registers) of different network topologies in the FPA architecture}
\label{fpa_table}
\begin{tabular}{lccc}
\hline
SNN: Topology         & Logic       & Registers    \\ \hline
784-100-10            & 13317       & 3836         \\ 
784-200-10            & 26225       & 7048         \\ 
784-300-10            & 31461       & 10974        \\ 
784-300-300-10        & 47257       & 24008        \\ 
784-300-300-300-10    & 60628       & 40600        \\ \hline
\end{tabular}
\end{center}
\end{table}
\begin{figure}
    \centering
    \centerline{\includegraphics[width=\linewidth]{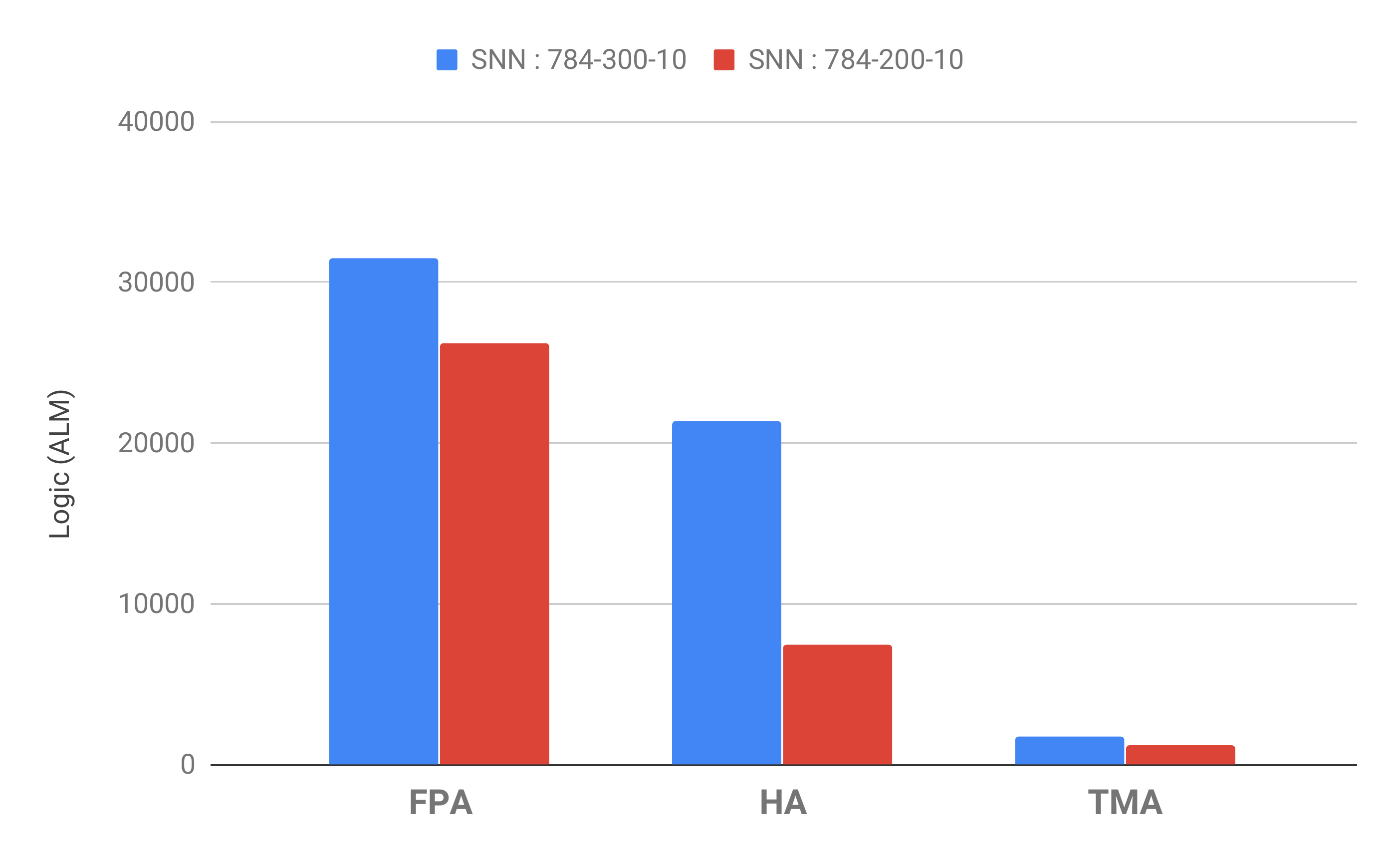}}
    \caption{Logic occupation of two SNN topologies comparing the three architectures (FPA, TMA and HA)}
    \label{logic_archi}
\end{figure}
%EDGAR STYLE DEBUT
In this context, we have synthesized the same SNN topologies using these two architectures (\textit{TMA} and \textit{HA}), the results are shown in tables \ref{tma_table} and \ref{ha_table}. Different memory types and organizations are used, but the memory footprint should be the same since the same SNN topologies are implemented, i.e., equal amount of parameters and activities to store in memories. Therefore, we focus on the occupation of FPGA logic cells, where the major difference between the three architectures should be found. For improved clarity, we have plotted the histogram shown in figure \ref{logic_archi} representing the logic occupation of the three architectures. As expected, \textit{FPA} occupies much more logic resources than the other architectures.
%EDGAR STYLE FIN
\begin{table}
\begin{center}
\caption{FPGA cyclone V resources occupation of different SNN topologies with the TMA architecture}  
\label{tma_table}
\begin{tabular}{lcccc}
\hline
SNN topology            & Logic & Registers & BRAM (KB) \\ \hline
784-100-10              & 690   & 1255      & 64    \\
784-200-10              & 1192  & 2168      & 128   \\ 
784-300-10              & 1714  & 3082      & 230   \\
784-300-300-10          & 3235  & 5937      & 241   \\ 
784-300-300-300-10      & 4736  & 8799      & 249   \\ \hline
\end{tabular}
\end{center}
\end{table}

\begin{table}
\begin{center}
\caption{FPGA cyclone V resources occupation of different SNN topologies with the \textit{HA} architecture}
\label{ha_table}
\begin{tabular}{lccc}
\hline
SNN topology & Logic & Registers \\ \hline
784-100-10 & 2440 & 1383 \\
784-200-10 & 7478 & 2434 \\
784-300-10 & 21406 & 3455 \\
784-300-300-10 & 22638 & 6318 \\
784-300-300-300-10 & 22859 & 9336 \\ \hline
\end{tabular}
\end{center}
\end{table}

On the other hand, using average spikes generated for a pattern with the same SNN topologies, we have estimated the latency of each architecture, as shown in table \ref{latency_table}. We observe that the processing latency is increased as hardware resources are decreased. Indeed, time-multiplexing allows to reduce the quantity of hardware resources, but relies on the sequentialization of a parallel task, thus resulting in a higher processing latency. This is why three different architectures have been designed: to evaluate the trade-off between hardware resources and processing latency. In this context, the \textit{HA} is an intermediary solution with a significant reduction in the amount of hardware resources, while maintaining reasonable latency. On average, it has a gain of \textbf{56.19\%} in terms of latency compared to \textit{TMA} and \textbf{57.05\%} in terms of logic occupation compared to \textit{FPA}\footnote{Referring to: latency table \ref{latency_table} and logic occupation tables \ref{fpa_table}, \ref{tma_table} and \ref{ha_table}}.

\begin{table}
\centering
\caption{Latency represented as the number of cycles spent in average to process an input image by the different architectures. The results correspond to the 784-300-300-300-10 SNN using the different information coding methods.}
\label{latency_table}
\begin{tabular}{cccc}
\hline
\multirow{2}{*}{Coding method } & \multicolumn{3}{c}{Latency (cycles)}                                                     \\ 
                    & FPA  & HA     & TMA    \\ \hline
Jittered Periodic                   & 1039,5                    & 84064                            & 300540                      \\
Spike Select                        & 1660,5                    & 34437                            & 496990                      \\
First Spike                        & 332                       & 23540                            & 74370                       \\ 
Single Burst        & 3077          & 441432                    & 459970 \\ \hline
\end{tabular}
\end{table}

Finally, in order to analyze the computation performance of our architectures, a measurement of SOPS (Synaptic Operation per Second) was performed on all three architectures for the 784-300-300-10 SNN topology on the same FPGA board. The \textit{FPA} achieves the best computation performance with 51.02 billion SOPS, whereas the \textit{TMA} only achieves 283.80 million SOPS. The \textit{HA} is just below \textit{FPA}, as it achieves 23.12 billion SOPS with the same topology and FPGA fabric. Their respective measured maximum computation frequencies are 83.51 MHz for \textit{FPA}, 76.3 MHz for \textit{TMA} and 70.95 MHz for \textit{HA}. An ongoing work concerns the power consumption analysis of the different SNN hardware architectures.

\section{Discussion} \label{discussion}

\subsection*{Review of our design flow}

In the present study, we have presented and explained a thoughtful Design Space Exploration framework for neuromorphic hardware. This framework is based on a funnel fashion: We start with high-level modeling leading to coarse architectural choices, which will drive lower-level modeling providing finer architectural choices. Here, we will validate our design flow by showing the coherence between high-level and low-level results, and the relevance of this funnel-like design flow for neuromorphic architecture design for specific embedded applications.

\begin{figure}
    \centering
    \includegraphics[width=\linewidth]{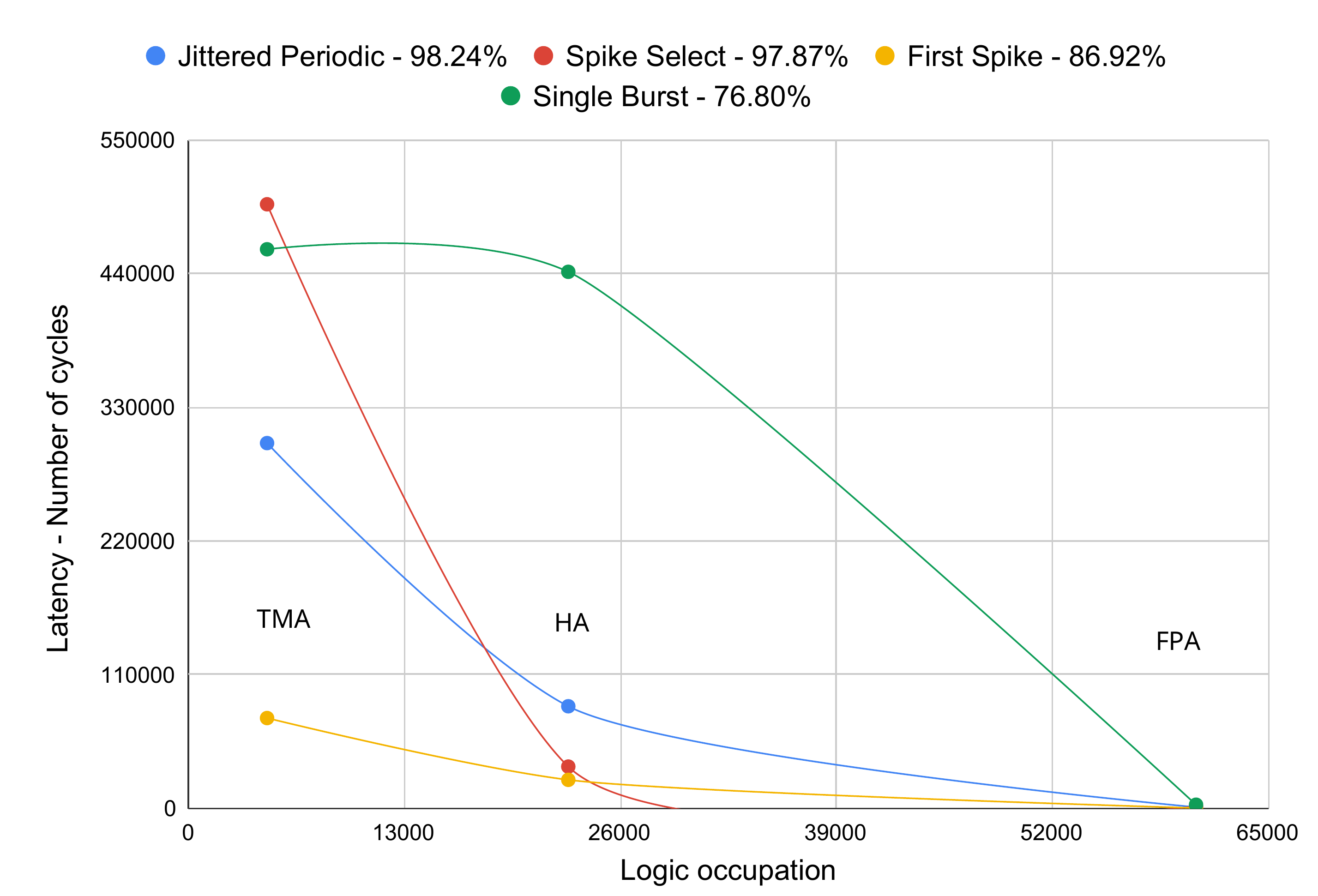}
    \caption{Tradeoff between logic occupation and latency of the hardware architectures according to the different information coding methods. The graph corresponds to FPGA logic occupation in tables \ref{fpa_table}, \ref{tma_table} and \ref{ha_table}) and latency estimation in table \ref{latency_table}, which are the data recorded for the SNN of "784-300-300-300-10" topology. The obtained recognition rates on MNIST test dataset with the different methods are indicated in the legend of the figure. }
    %BM2: pour quelle taille/topologie de réseau ?
    %NA: OK! le tableaux de latences et l'occupation de ressources FPGA (tables 4,5,6,7) 
    \label{logic_latence}
\end{figure}

% 1) MONTRER LA COHERENCE DES RESULTATS
As a reminder, the high-level results obtained with the NAXT simulator (in section \ref{exploration}) are summarized in table \ref{Resultattable} and figure \ref{Resultatfigure}. As already explained, surface estimations provided by NAXT correspond to an ASIC target, and can be seen as qualitatively equivalent to logic occupation for an FPGA target. In these results, the trade-off between latency and FPGA occupation (i.e., chip surface in the figure and table) was clearly visible: \textit{FPA} had low latency but high FPGA occupation, and the opposite was true for \textit{TMA}. Fine-grained results provided by RTL synthesis and latency estimations for the "784-300-300-300-10" SNN are summarized in figure \ref{logic_latence}.
%BM2: j'ai ajouté la fin de phrase précédente. Il s'agit bien de ce réseau ?
%NA2: ok!
This figure shows Pareto curves representing FPGA logic occupation versus latency (number of cycles) for the three hardware architectures according to the information coding methods described in section \ref{information_coding}. These results are consistent with the high-level estimations, where the same trade-off can be seen between latency and logic occupation: \textit{FPA} has a low latency but high logic occupation, whereas \textit{TMA} has high latency but low logic occupation. If we consider the information coding method without looking at the recognition rate, the First Spike method combined with TMA architecture has the best "latency / chip surface" trade-off. However, this method has a loss of around 10\% in terms of accuracy compared to Spike Select and Jittered Periodic methods. Therefore, taking into account the accuracy criterion, the Spike Select method combined with HA architecture has the best latency / logic occupation trade-off. The method, while performing 97.87\% accuracy, allows only few spikes propagating in the deeper layers of the SNN, which fits well the HA architecture making this combination one of the best choices for hardware implementation of deep SNNs.

\begin{figure}
    \centering
    \includegraphics[width=\linewidth]{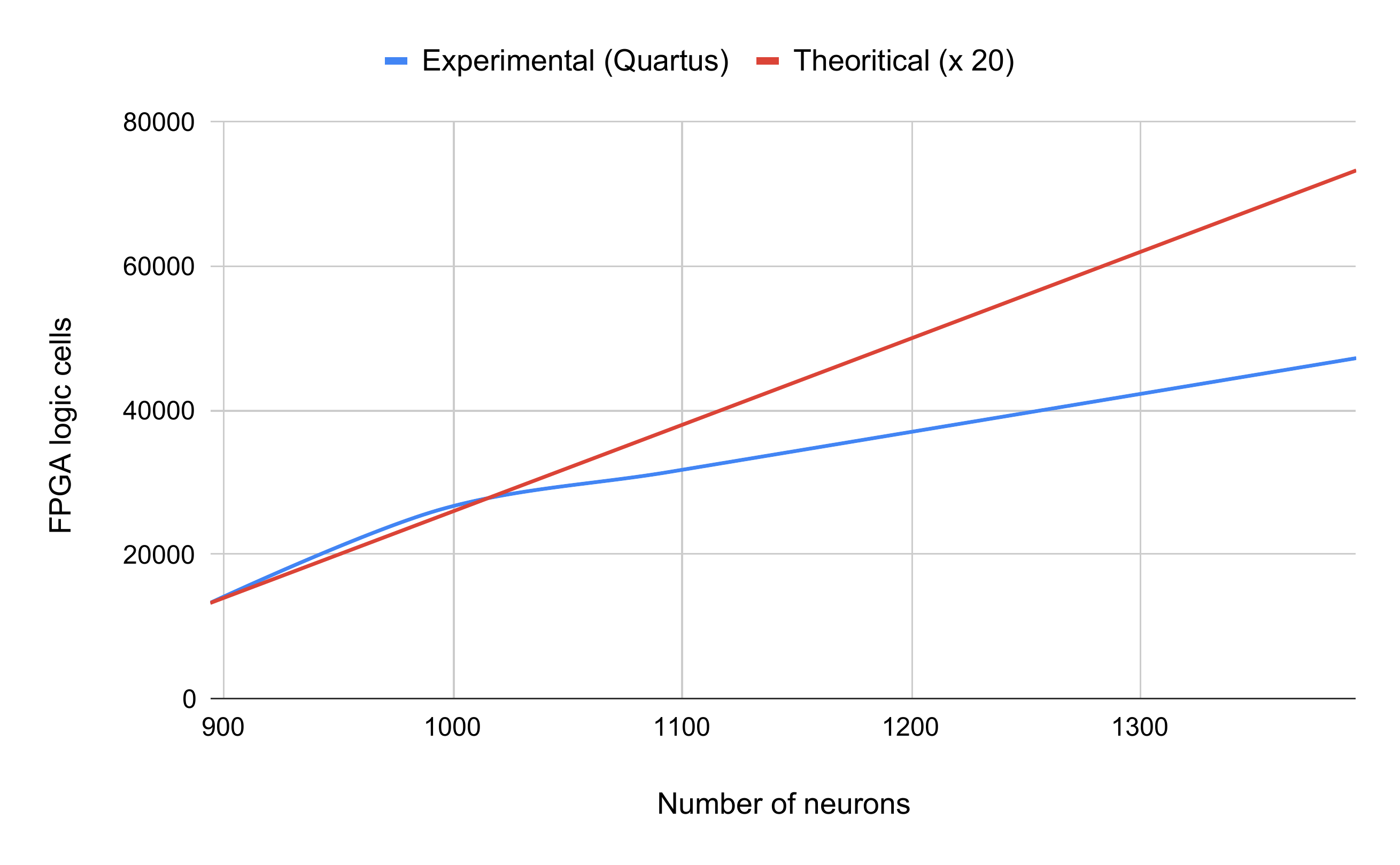}
    \caption{FPA theoretical versus experimental FPGA occupation results with respect to the number of neurons.}
    \label{theory_experiment}
\end{figure}

The coherence of these results is shown in figure \ref{theory_experiment}, which depicts the evolution of FPGA occupation (in terms of logic cells) against the number of neurons, for both theoretical estimations and {Quartus\textregistered} experimental results. The considered network has a fully-parallel architecture in both cases. Both curves are very similar for a low number of neurons, which confirms coherency between estimations and experimental results. The divergence observed for higher numbers of neurons is due to synthesis optimizations performed by {Quartus\textregistered}, which are not taken in account in our estimations. However, the two curves remain qualitatively coherent, as they follow similar linear growths. Hence, the results are coherent between the high-level part and the low-level part of our design-flow.

% 2) TEMPS DE SIMULATION FAIBLE AVEC NAXT MAIS ELEVE AVEC QUARTUS: On commence par "débroussailler" avec NAXT puis on affine notre choix avec RTL/Quartus.
The main interest of our Design Space Exploration framework lies in its funnel-like organization. The high-level simulations performed by the NAXT simulator are quite fast (a few seconds to a few minutes, depending on network size and architectural choices), and provide sufficiently precise results to disqualify unsuitable architectural choices. In doing so, we select a restrained number of potentially suitable architectural choices. The RTL synthesis are longer (they can take many hours for deep SNN topologies with \textit{FPA}) but more precise, and let us determine the best choices among the pre-selected options. Indeed, operating low-level architectural exploration with RTL synthesis among the whole design space would take too long, since the range of possibilities is very wide. Thus, our presented funnel-like framework allows for efficient and reliable Design Space Exploration of Neuromorphic hardware.

\subsection*{Future architectures (CNN)}

At this point, our work focused on fully-connected networks, the so-called classifiers \cite{zhang2000neural}. However, this type of neural networks is restrained to simple classification tasks: they are not able to perform classification on complex data (face recognition, for example), and are not resilient to image rotation, scaling or translation. Thus, modern ANNs for complex data recognition and classification involve convolution and pooling layers: these are the Convolutional Neural Networks (CNN) \cite{lecun1995convolutional} \cite{alexnet}. The Convolution and Pooling layers enable feature extraction and combination, resulting in a Feature Map that can be fed to a simple classifier afterwards.

In order to simulate state-of-the-art ANN hardware implementations, we aim to develop hardware architectures for spiking Convolution and Pooling layers in future work.

\subsection*{Asynchronous sensor: towards frame-free SNNs for video recognition}

In this work, we have focused on static image recognition. Thus, we based our approach on a transcoding method in which input data (pixel values) are translated into spikes (see section \ref{information_coding}). This transcoding step is one of the main drawbacks of our approach to SNNs utilization, as it may counterbalance the energy, latency and surface savings we achieved thanks to spike-based processing. When it comes to video recognition, however, this issue can be tackled by using event-based cameras.

Indeed, in contrast to static images, videos can be directly recorded in an event-based fashion, with so-called asynchronous cameras \cite{delbruck2010activity}. In contrast to classical cameras, which output a succession of discrete frames, an event-based camera emits a continuous flow of events: each pixel outputs a spike whenever an edge crosses its receptive field. In other words, an asynchronous vision sensor outputs a flow of spikes representing the movement happening in its field of view. We expect that SNNs could benefit from the use of such innovative sensor, as the processing would eliminate the time-and-energy-consuming transcoding step.

Moreover, Farabet \textit{et al.} \cite{farabet_comparison_2012} have proven that such a fully event-based frame-free processing flow would bring input-to-output pseudo-simultaneity, that is, real-time processing ability. Thus, we expect that SNNs combined with asynchronous sensors would be very well suited to embedded artificial intelligence for real-time video recognition and classification. In light of these expectations, we aim to adapt our current architectures to video processing and to develop an asynchronous sensor interface for that purpose.

\section{Conclusion}
\label{conclusion_sec}

The present work describes an efficient novel workflow for Design Space Exploration of Neuromorphic hardware. This design flow follows a funnel-like structure:  First, an analytical preliminary study determines if an architecture is feasible in terms of hardware resources (FPGA occupation) and memory footprint, which helps matching our application with a hardware target; second, a high-level exploration performed with the novel NAXT software indicates suitable high-level architectural choices; third, a low-level RTL simulation let us determine the best implementation and provides a fine-grained evaluation of this architecture for the FPGA target. The results obtained from all these steps are consistent with the tested network models, indicating that this workflow is suitable for Neuromorphic System Design Space Exploration.
In this paper, we chose the typical application case of handwritten digits recognition (MNIST dataset) to illustrate our workflow, which led to the realization of three different SNN implementations: \textit{Fully Parallel Architecture} and \textit{Time-Multiplexed Architecture} were developed to emphasize the two extremes of the latency versus hardware resources trade-off; and a novel and innovative \textit{Hybrid Architecture} was created as a middle ground, deriving from the findings and observations of our Design Space Exploration work.

Moreover, the present work addresses the information coding influence on accuracy and spiking activity. This study shows that the most suitable information coding paradigm was the novel Spike Select coding, as it ensures high prediction accuracy and sparse spiking activity in the network. Spike sparsity implies a lower number of spikes per pattern, resulting in a shorter processing and a lower energy consumption, which is suitable for embedded system applications. Moreover, this novel spike coding method is tightly suited to our innovative Hybrid Architecture.

For deep SNNs, and according to our design flow, the most suitable architecture is our novel Hybrid Architecture, as it takes advantage of the increasing spiking activity sparsity as we go deeper into the network. This novel architecture has been developed in our lab, and to the best of our knowledge, is completely original. Combined with Spike Select Coding, it appears to be one of the most suitable approaches for future Deep SNN implementation into embedded systems. 
%BM2: Joli travail !
%BM2: Le seul regret à la fin est de ne pas disposer d'estimations de conso de nos archis. J'ai peur que cela soit un possible reproche. On verra si on a le temps d'y réfléchir... et surtout de faire les simus avec Qua

\section*{Acknowledgement}
We would like to thank Olivier Bichler and CEA LIST for providing us the N2D2 framework licence. This work is funded by "Universit\'e C\^ote d'Azur", "CNRS" and "R\'egion Sud Provence-Alpes-C\^ote d'Azur, France".

\bibliographystyle{ieeetr}
\bibliography{main.bib}
\makeatletter

%\def\pct{\expandafter\@gobble\string\%}
%\immediate\write\@auxout{\pct\space This is a test line.\pct }

\end{document}